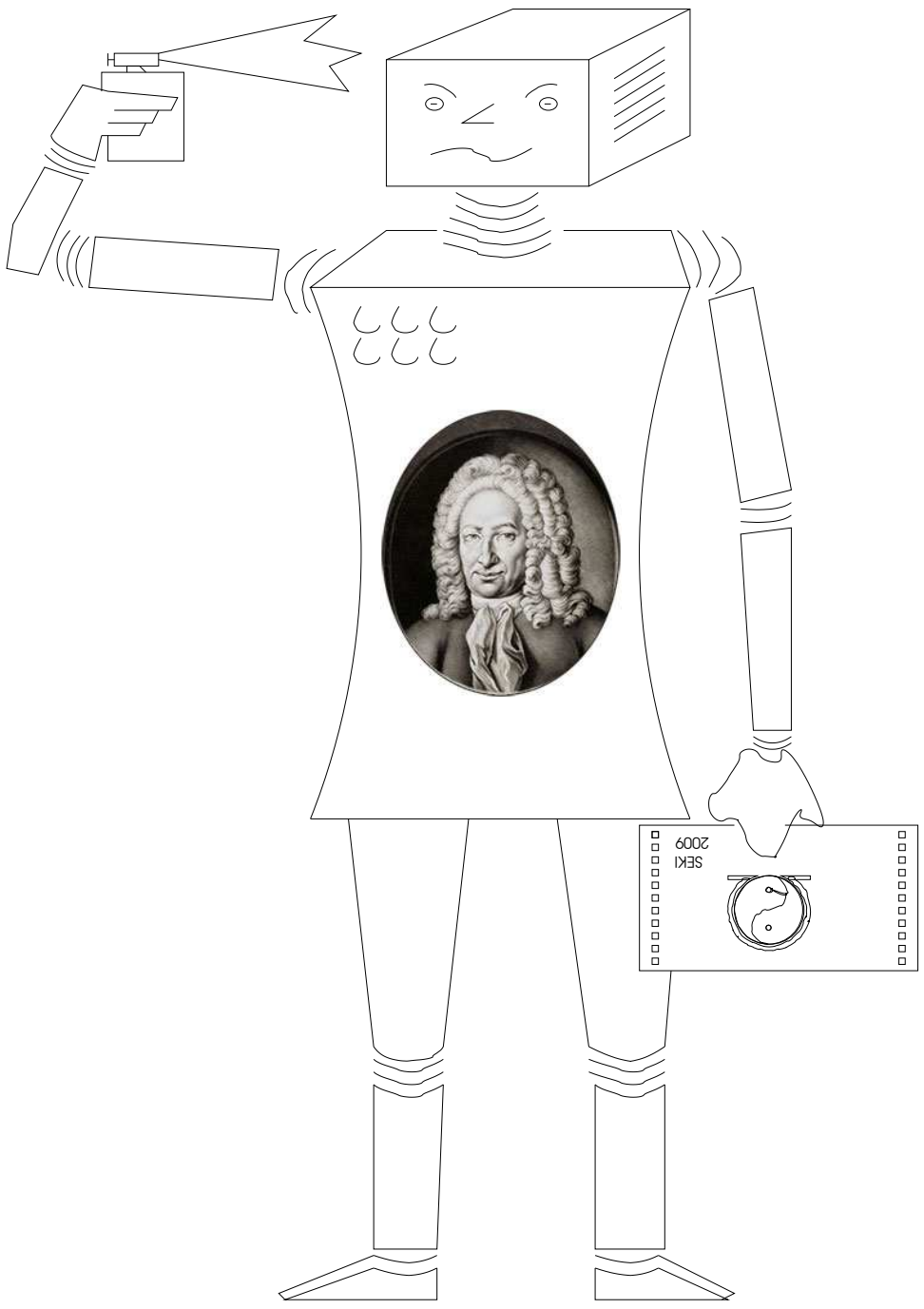

# A Simplified and Improved Free-Variable Framework for HILBERT's epsilon as an Operator of Indefinite Committed Choice

CLAUS-PETER WIRTH

Dept. of Computer Sci., Saarland Univ.,
66123 Saarbrücken, Germany

wirth@logic.at

# A Simplified and Improved Free-Variable Framework for HILBERT's epsilon as an Operator of Indefinite Committed Choice

CLAUS-PETER WIRTH

Dept. of Computer Sci., Saarland Univ., 66123 Saarbrücken, Germany

wirth@logic.at

**Abstract**

Free variables occur frequently in mathematics and computer science with *ad hoc* and altering semantics. We present here the most recent version of our free-variable framework for two-valued logics with properly improved functionality, but only two kinds of free variables left (instead of three): implicitly universally and implicitly existentially quantified ones, now simply called "free atoms" and "variables", respectively. The quantificational expressiveness and the problem-solving facilities of our framework exceed standard first-order logic and even higher-order modal logics, and directly support FERMAT's *descente infinie.* With the improved version of our framework, we can now model also HENKIN quantification, neither using any binders (such as quantifiers or epsilons) nor raising (SKOLEMization). Based only on the traditional $\varepsilon$-formula of HILBERT–BERNAYS, we present our flexible and elegant semantics for HILBERT's $\varepsilon$ as a choice operator with the following features: We avoid overspecification (such as right-uniqueness), but admit indefinite choice, committed choice, and classical logics. Moreover, our semantics for the $\varepsilon$ supports reductive proof search optimally.

# Contents

# 1 Overview

## 1.1 Localizing Our Work in the Field of Artificial Intelligence (AI)

We wish human-assisted and automatic theorem proving to turn into a practical method by means of *human-style calculi*, so that *artificial intelligence* can learn from human mathematicians how to prove — i.e. to develop, formalize and verify — hard mathematical theorems more and more automatically and with lesser and lesser human guidance.

In the middle of the 1990s, the exciting developments of AI during the second half of the 20th century were gathered in the outstanding textbook [Russell & Norvig, 1995] in the form of standard technologies nowadays taught already to the freshmen in informatics and computer science. At the same time, after over twenty years of intensive ingenious development, AI was just celebrating its biggest success in automated theorem proving up to today with ACL2, an industrial-strength automated theorem prover with particular strength in the area of the inductive data types of modern programming languages and computer hardware, where inductive theorem proving was required for the verification of software and hardware, cf. [Moore & Wirth, 2017].

Nowadays, the term "AI" most of the time addresses something completely different, namely statistical methods that have reached unforeseen abilities in assessing natural-language input for translation and query answering; and it seems to have been completely forgotten that these statistical methods have no form of knowledge related to any form of classical, mathematical or modern logic. To pass the border to which these statistical methods converge, we will need logics that are much better than those that are known nowadays: They must be, on the one hand, closer to automatic theorem proving and the statistical AI and, on the other hand, more human-oriented in the sense that they must be closer to the demands of deep natural-language analysis and to the actual practice of human mathematicians. And we will present here a small but crucial step toward the latter of these two goals.

## 1.2 New in the First Edition of 2011 compared to [Wirth, 2006b]

Driven by a weakness in representing Henkin quantification described in [Wirth, 2006b, § 6.4.1] and inspired by nominal terms (cf. e.g. [Urban &al., 2004]), in this paper we significantly simplify and improve our semantic free-variable framework for two-valued logics:

1. We replace the two-layered construction of free $\delta^+$-variables on top of free $\gamma$-variables over free $\delta^-$-variables of [Wirth, 2004; 2006b; 2008] with a one-layered construction of *variables* over *free atoms*:
   - Free atoms now play the former rôle of the $\delta^-$-variables.
   - Variables without choice-condition play the former rôle of the $\gamma$-variables.
   - Variables with choice-condition play the former rôle of the $\delta^+$-variables.
2. As a consequence, the proofs of the lemmas and theorems have shortened by more than a factor of 2. Therefore, we can now present all the proofs in this paper and make it self-contained in this aspect; whereas in [Wirth, 2006b; 2008], we had to point to [Wirth, 2004] for most of the proofs.

3. The difference between variables and free atoms and their names are now more standard and more clear than those of the different free variables before; cf. § 2.1.
4. Compared to [Wirth, 2004], besides shortening the proofs, we have made the meta-level presuppositions more explicit in this paper; cf. § 7.1.
5. Last but not least, we can now treat Henkin quantification in a direct way; cf. § 6.

Taking all these points together, the version of our free-variable framework presented in this paper is the version we recommend for further reference, development, and application: it is indeed much easier to handle than its predecessors.

And so we found it appropriate to present most of the material from [Wirth, 2006b; 2008] in this paper in the improved form; we have omitted only the discussions on the tailoring of operators similar to our $\varepsilon$, and on the analysis of natural-language semantics. The material on mathematical induction in the style of Fermat's *descente infinie* in our framework of [Wirth, 2004] is to be reorganized accordingly in a later publication.

## 1.3 New in This Edition of 2024 compared to the First Edition

This tenth edition has added two dozen pages compared to the ninth edition [Wirth, 2017c], which already had added two dozen pages to the first edition of 2011. These extra pages were strictly required to make this text sufficiently comprehensible for the large practically or theoretically oriented audience of mathematicians, logicians and AI communities. Besides countless corrections, removals of sloppiness, and improvement of representation, most of the effort was directed to make this tenth edition much more easily accessible to a larger audience. The next step will be to add Fermat's *descente infinie* as found in [Wirth, 2004] explicitly to this text and to develop it into an advanced textbook.

## 1.4 Organization

This paper is organized as follows. There are three introductory sections: to our variables and free atoms (§ 2), to their relation to our reductive inference rules (§ 3), and to Hilbert's $\varepsilon$ (§ 4). After formalizing all our syntactic ingredients (§ 5), we discuss our most interesting example — which shows that we now can even formalize Henkin quantification without raising and IF-logic quantifiers with our new positive/negative variable-conditions — in § 6. Mostly for the skeptics and the developers of logic, but also for the admires of technically difficult model theory, we formalize our novel approach to the semantics of our variables, free atoms, and the $\varepsilon$ after all in § 7. Of practical interest again is § 8, where explain the consequences of our semantics for and reasoning with our reductive inference rules, including lemma and induction-hypothesis application. Finally, we summarize and discuss our whole approach in § 9, and conclude in § 10. In the Appendix, you can find a surprising hint that the liberalized $\delta^+$-rules sometimes might be less liberalized than the old-fashioned $\delta^-$-rules (§ A), and discussions of the literature on extended semantics given to Hilbert's $\varepsilon$-operator in the 2nd half of the 20th century (§ B), as well as of the design questions of variable-conditions (§ C). Moreover, at the very end of the Appendix, you can find a single common index and, before that, all references and all proofs — except the beautiful high-level proof of Theorem 8.6.

# 2 Introduction to Free Atoms and Variables

## 2.1 Essential Notions and Notation

Free atoms and variables frequently occur in practice of mathematics and computer science. The logical function of these free symbols varies locally; it is typically determined *ad hoc* by the context. And the intended semantics is given only *implicitly* and varies from context to context. In this paper, however, we will make the semantics of our free atoms and variables *explicit* by using *disjoint* sets of symbols for different semantic functions; namely we will use the following sets of symbols:

$\mathbb{V}$ (the set of (free) *variables*),
$\mathbb{A}$ (the set of *free atoms*),
$\mathbb{B}$ (the set of *bound atoms*).

A *variable*, in the sense we use the word in this paper, is a place-holder in a proof attempt or in a discourse, which may gather and store information and may be concretized later on, by a definition or a description. The name "(free) variable" for such a place-holder comes from *free-variable semantic tableaus*; cf. [Fitting, 1990; 1996].

In our paper here, variables are always free; only bound atoms (i.e. elements of $\mathbb{B}$) can be bound by *binders*, such as quantifiers or operators. Bound symbols had better be called "bindable" instead of "bound", because we will have to treat some unbound occurrences of bound atoms occasionally. When the notion of bound symbols (actually, in German: "gebundene Variablen", cf. § 4 of [Hilbert & Bernays, 1934; 2017b]) became standard around the year 1930, however, neither "bindable" nor the German "bindbar" were considered to be proper words of their respective languages, and so the past participle was chosen.

An *atom* stands for an arbitrary object in a proof attempt or a discourse. Atoms cannot gather information and are invariant under renaming. And we will never want to know anything about a possible atom but whether it is an atom, and — if yes — whether it is identical to another atom or not, and possibly some elementary syntactical properties, such as its name, sort or type. In our context here, for reasons of convenience and efficiency, we would also like to know whether an atom is a free atom or a bound atom, i.e. whether it is from $\mathbb{A}$ or $\mathbb{B}$. The name "atom" for such an object has a tradition in set theories with atoms. In German, nouns are always capitalized and an atom is also called an *urelement* in the context of set theories, but that alternative name puts some emphasis on the origin of creation, in which we are not interested for our atoms. Thus, in this paper, we use the name "urelement" only in the context of set theories and "atom" only in the context of our framework of variables and atoms.

The classification as a (free) variable, free atom, or bound atom will be indicated by adjoining a "$\mathbb{V}$", an "$\mathbb{A}$", or a "$\mathbb{B}$", respectively, as a label to the upper right of the meta-variable for the symbol. If a meta-variable stands for a symbol of the union of some of these sets, we will indicate this by listing all possible sets; e.g. "$x^{\mathbb{VA}}$" is a meta-variable for a symbol that may be either a variable or a free atom. Of course, meta-variables with disjoint labels always denote different symbols; e.g. "$x^{\mathbb{V}}$" and "$x^{\mathbb{A}}$" will always denote different symbols, whereas "$x^{\mathbb{VA}}$" may denote the same symbol as "$x^{\mathbb{A}}$". In formal discussions, also "$x^{\mathbb{A}}$" and "$y^{\mathbb{A}}$" may denote the same symbol. In concrete examples, however, we will implicitly assume that different meta-variables denote different symbols.

## 2.2 Bound and Free Atoms

The following was already noted by Bertrand Russell in [Russell, 1919, p.155].

In mathematical practice, *free symbols* often have an obviously universal intention, such as the free symbols $m$, $p$, and $q$ in the formula

$$(m)^{(p+q)} = (m)^{(p)} * (m)^{(q)}.$$

Then, this formula is not meant to denote a propositional function (say, from integers to truth values), but actually stands for the explicitly universally quantified, closed formula

$$\forall m^{\mathbb{B}}, p^{\mathbb{B}}, q^{\mathbb{B}}.\ \Big(\ (m^{\mathbb{B}})^{(p^{\mathbb{B}}+q^{\mathbb{B}})} = (m^{\mathbb{B}})^{(p^{\mathbb{B}})} * (m^{\mathbb{B}})^{(q^{\mathbb{B}})}\ \Big)$$

with symbols $m^{\mathbb{B}}$, $p^{\mathbb{B}}$, and $q^{\mathbb{B}}$ bound by the universal quantifier $\forall$.

In our framework, we call such symbols *bound atoms*, and bound atoms are the only symbols that may be bound by any binders. Moreover, a bound atom may occur in a formula only in the scope of a binder on it.

In this paper, improving mathematical practice, we write the first formula as

$$(m^{\mathbb{A}})^{(p^{\mathbb{A}}+q^{\mathbb{A}})} = (m^{\mathbb{A}})^{(p^{\mathbb{A}})} * (m^{\mathbb{A}})^{(q^{\mathbb{A}})},$$

which is a formula with *free atoms.* Independent of its context, it is logically equivalent to the explicitly universally quantified formula, but also admits the reference to the free atoms, which is required for mathematical induction in the style of Fermat's *descente infinie*, and also beneficial for solving reference problems in the analysis of natural language. So the third version combines the practical advantages of the first version with the semantic clarity of the second.

## 2.3 Variables

In mathematical practice, it is somehow clear that the linear system of the formula

$$\begin{pmatrix} 2 & 3 \\ 5 & 7 \end{pmatrix} \begin{pmatrix} x \\ y \end{pmatrix} = \begin{pmatrix} 11 \\ 13 \end{pmatrix}$$

asks us to find the set of solutions for the symbols $x$ and $y$, say $(x, y) \in \{(-38, 29)\}$. The mere existence of such a solution is expressed by the explicitly existentially quantified, closed formula

$$\exists x^{\mathbb{B}}, y^{\mathbb{B}}.\ \left(\ \begin{pmatrix} 2 & 3 \\ 5 & 7 \end{pmatrix} \begin{pmatrix} x^{\mathbb{B}} \\ y^{\mathbb{B}} \end{pmatrix} = \begin{pmatrix} 11 \\ 13 \end{pmatrix}\ \right).$$

In this paper, improving mathematical practice, we write the first formula as

$$\begin{pmatrix} 2 & 3 \\ 5 & 7 \end{pmatrix} \begin{pmatrix} x^{\mathbb{V}} \\ y^{\mathbb{V}} \end{pmatrix} = \begin{pmatrix} 11 \\ 13 \end{pmatrix},$$

which is a formula with *(free) variables.* Independent of its context, it is logically equivalent to the explicitly existentially quantified formula, but it admits also the reference to the variables, which is required for retrieving solutions for $x^{\mathbb{V}}$ and $y^{\mathbb{V}}$ as instantiations for $x^{\mathbb{V}}$ and $y^{\mathbb{V}}$ chosen in a formal proof (or a *query* in logical programming). So the third version again combines the practical advantages of the first with the semantic clarity of the second.

## 2.4 Instantiation of Variables and Free Atoms

Both variables and atoms may be instantiated with terms. Only variables, however, may refer to free atoms or other variables, or may *depend* on them as a working mathematician would say. Moreover, only variables have the following properties w.r.t. instantiation:

**Syntactic Restrictions:** If a variable $x^{\mathrm{V}}$ is instantiated with a term $t$, then this affects *all* occurrences of $x^{\mathrm{V}}$ in the entire state of the proof attempt (i.e. $x^{\mathrm{V}}$ is *rigid* in the terminology of semantic tableaus). Thus — provided that the instantiation is executed eagerly — the variable must be replaced *globally* in all terms and formulas of the entire state of the proof attempt with the same term $t$, and afterward the variable can be completely eliminated from the current state of the proof attempt, provided that the variable $x^{\mathrm{V}}$ itself does not occur in the term $t$.

**Overall effects:** While the elimination of the variable $x^{\mathrm{V}}$ after its eager instantiation with such a term $t$ cannot have any effect on the possibility to complete the proof attempt into a successful proof, the choice of a wrong term $t$ may well turn the proof attempt into a necessarily failing one, in particular if the instantiation falsifies the input proposition to be shown (or the query to be answered). On the other hand, the instantiation may be relevant for the consequences of a successful proof because the input proposition (or query) may be strengthened by the global replacement of one of its variables with a term providing an actual witness for the existential property of the original proposition (or an answer to the query).

By contrast to these properties of variables, atoms cannot refer to any other symbols, nor depend on them in any form. Moreover, free atoms have the following properties w.r.t. instantiation:

**Syntactic Restrictions:** A free atom may be

1. globally renamed to a fresh free atom, or else
2. locally and possibly repeatedly instantiated with arbitrary different terms in the application of lemmas or induction hypotheses (provided that the instantiation is admissible in the sense of Theorem 8.5(7)).

**Overall effects:** Neither the possibility to complete a proof attempt, nor the consequences of a successful proof can be changed by any of these two forms of instantiating a free atom.

# 3 Introduction to Reductive Inference Rules

We will now present the essential reductive inference rules for our free-variable framework. Regarding form and notation, please note the following:

- To combine the following four aspects, we have to present our framework here with a *sequent-calculus*: Our calculus must be easy to grasp and — at the same time — admit explicit indication of $\gamma$-multiplicity, rule isomorphism, and eliminability of formulas. As we restrict ourselves to two-valued logics here, we just take the right-hand side of standard sequents. This means that our *sequents* are just disjunctive lists of formulas.
- We assume that all binders have minimal scope; e.g. $\forall x^{\mathbb{B}}, y^{\mathbb{B}}.\ A \wedge B$ reads $(\forall x^{\mathbb{B}}.\ \forall y^{\mathbb{B}}.\ A) \wedge B$
- Our reductive inference rules will be written "reductively" in the sense that passing the line means *reduction*, the reverse of deduction. Note that in the good old days when trees grew upward, GERHARD GENTZEN would have inverted the inference rules such that passing the line means consequence. In our case, passing the line means reduction, and trees grow downward.
- RAYMOND M. SMULLYAN has classified reductive inference rules into *$\alpha$-, $\beta$-, $\gamma$-, and $\delta$-rules*, and invented a *uniform notation* for them; cf. [SMULLYAN, 1968].

In the following rules, let $s$, $t$ be terms, $A$, $B$ formulas, and $\Gamma$ and $\Pi$ sequents. The notions of a *principal formula* (in German: Hauptformel) and a *side formula* (Seitenformel) were introduced in [GENTZEN, 1935] and refined in [SCHMIDT-SAMOA, 2006]. Roughly speaking, the principal formula of an inference rule is the formula that is reduced by that rule, and the side formulas are the resulting pieces replacing the principal formula. In our reductive inference rules here, the principal formulas are the formulas above the lines except the ones in $\Gamma$, $\Pi$ (which are called *parametric formulas*, in German: Nebenformeln), and the side formulas are the formulas below the lines except the ones in $\Gamma$, $\Pi$.

## 3.1 Tautologies

Tautologies form reductive inference rules reducing a sequent to zero sequents. Applied to a leaf of a proof tree, this leaf will not count as a leaf anymore, which becomes clear if we see a rule application to a leaf as putting a node below this leaf, indicating the rule's name and actual parameters, which again has the new sequents as offspring. As tautologies we take those sequents that list as formulas (among others) either two formulas of form $A$ and $\neg A$, or, in case we have a primitive equality symbol '$=$' in our language, the formula $(t\!=\!t)$.

## 3.2 $\alpha$-Rules

Rules that reduce a sequent to a *single* sequent are called *non-branching.* The primitive ones of the non-branching rules count as $\alpha$-rules, such as the cancellation rule for double negation and one elimination rule for each binary propositional operator of our language.

If the only such operator is implication, then the propositional such rules are the following **$\alpha$-rules:**

$$\frac{\Gamma \quad \neg\neg A \quad \Pi}{\Gamma \quad A \quad \Pi} \qquad\qquad \frac{\Gamma \quad A \Rightarrow B \quad \Pi}{\Gamma \quad \neg A \quad B \quad \Pi}$$

Sometimes also other primitive non-branching rules count as $\alpha$-rules, though these rules are not propositional. For instance, in case we have a primitive equality symbol '$=$' in our language, a very useful rule that we should include among the $\alpha$-rules is the the following.

**Primitive contextual-rewriting rule:** If $\neg(s=t)$ or $\neg(t=s)$ is listed as a formula in $\Gamma$ or $\Pi$, then:

$$\frac{\Gamma \quad A[s] \quad \Pi}{\Gamma \quad A[t] \quad \Pi}$$

Here, $A[s]$ denotes a formula context with a hole that is filled by the term $s$.

## 3.3 $\beta$-Rules

**$\beta$-rules** are the branching propositional rules, which reduce a sequent to several sequents, such as

$$\frac{\Gamma \quad \neg(A \Rightarrow B) \quad \Pi}{\begin{array}{lccl}\Gamma & A & & \Pi\\ \Gamma & & \neg B & \Pi\end{array}}$$

## 3.4 $\gamma$-Rules

Suppose we want to prove an existential proposition $\exists y^{\mathbb{B}}.\ A$. Here "$y^{\mathbb{B}}$" is a bound variable according to standard terminology, but as it is an atom according to our classification of § 2.1, we will speak of a "bound atom" instead. Then the $\gamma$-rules of old-fashioned inference systems (such as [Gentzen, 1935] or [Smullyan, 1968]) enforce the choice of a witnessing term $t$ as a substitution for the bound atom *immediately* when incrementing the *multiplicity of a $\gamma$-quantification*, here indicated by an occurrence of $\exists$ or $\neg\forall$ at the root of a formula.

**$\gamma$-rules:** Let $t$ be any term:

$$\frac{\Gamma \quad \exists y^{\mathbb{B}}.\ A \quad \Pi}{A\{y^{\mathbb{B}}{\mapsto}t\} \quad \Gamma \quad \exists y^{\mathbb{B}}.\ A \quad \Pi} \qquad \frac{\Gamma \quad \neg\forall y^{\mathbb{B}}.\ A \quad \Pi}{\neg A\{y^{\mathbb{B}}{\mapsto}t\} \quad \Gamma \quad \neg\forall y^{\mathbb{B}}.\ A \quad \Pi}$$

More modern inference systems (such as the ones in [Fitting, 1996]) enable us to delay the crucial choice of the term $t$ until the state of the proof attempt may provide more information to make a successful decision. This delay is achieved by introducing a special kind of variable.

This special kind of variable is called "dummy" in [Prawitz, 1960] and [Kanger, 1963], "free variable" in [Fitting, 1990; 1996] and in Footnote 11 of [Prawitz, 1960], "meta variable" in the field of planning and constraint solving, and "free $\gamma$-variable" in [Wirth, 2004; 2006a; 2008; 2012a; 2006b; 2014] and [Wirth &al., 2009; 2014].

In this paper, we call these free variables simply "variables" and write them like "$x^{\mathbb{V}}$". If such additional variables are available, we can choose a fresh variable $y^{\mathbb{V}}$ for the arbitrary term $t$ of the, say, first $\gamma$-rule, reduce $\Gamma\ \exists y^{\mathbb{B}}.\ A\ \Pi$ first to $A\{y^{\mathbb{B}}{\mapsto}y^{\mathbb{V}}\}\ \ \Gamma\ \exists y^{\mathbb{B}}.\ A\ \Pi$, and then anytime later in the proof we may globally replace $y^{\mathbb{V}}$ with an appropriate term.

These variables join the inductive construction of admissible terms and therefore complicate the notation of the $\delta$-rules, but do not necessarily affect the above old-fashioned form of *notation of the $\gamma$-rules*: Although we could restrict the term $t$ to be a fresh variable, for convenience we may still admit arbitrary terms; but now including variables, of course.

## 3.5 $\delta^-$-Rules

A $\delta$-rule may introduce either a fresh free atom ($\delta^-$*-rule*) or an $\varepsilon$-constrained fresh variable ($\delta^+$*-rule,* cf. § 3.6).

**$\delta^-$-rules:** Let $x^{\mathbb{A}}$ be a fresh free atom:

$$\frac{\Gamma \quad \forall x^{\mathbb{B}}.\ A \quad \Pi}{A\{x^{\mathbb{B}}\mapsto x^{\mathbb{A}}\} \quad \Gamma \quad \Pi} \qquad \mathbb{V}(\Gamma\ \ \forall x^{\mathbb{B}}.\ A\ \ \Pi) \times \{x^{\mathbb{A}}\}$$

$$\frac{\Gamma \quad \neg\exists x^{\mathbb{B}}.\ A \quad \Pi}{\neg A\{x^{\mathbb{B}}\mapsto x^{\mathbb{A}}\} \quad \Gamma \quad \Pi} \qquad \mathbb{V}(\Gamma\ \ \neg\exists x^{\mathbb{B}}.\ A\ \ \Pi) \times \{x^{\mathbb{A}}\}$$

Note that, say, "$\mathbb{V}(\Gamma\ \ \forall x^{\mathbb{B}}.\ A\ \ \Pi)$" stands for the set of all variables (i.e. symbols from the set $\mathbb{V}$) that occur in the sequent $\Gamma\ \ \forall x^{\mathbb{B}}.\ A\ \ \Pi$.

Let us recall that a free atom typically stands for an arbitrary object in a discourse of which nothing else is known. The free atom $x^{\mathbb{A}}$ introduced by the $\delta^-$-rules is sometimes also called "parameter", "eigenvariable", or "free $\delta$-variable". In HILBERT-calculi, however, this free atom is called a "free *variable*", because the non-reductive (i.e. generative) deduction in HILBERT-calculi admits its unrestricted instantiation by the substitution rule, cf. p. 63 of [HILBERT & BERNAYS, 1934] or p. 62 of [HILBERT & BERNAYS, 1968; 2017b]. The analogues of the $\delta^-$-rules in HILBERT–BERNAYS' predicate calculus are Schemata ($\alpha$) and ($\beta$), to be found on p. 103f. of [HILBERT & BERNAYS, 1934] or on p. 102f. of [HILBERT & BERNAYS, 1968; 2017b].

In our calculi, however, the occurrence of the free atom $x^{\mathbb{A}}$ of the $\delta^-$-rules must be disallowed in the terms that may be used to replace those variables which have already been in use when $x^{\mathbb{A}}$ was introduced by application of the $\delta^-$-rule, i.e. the variables of the upper sequent to which the $\delta^-$-rule was applied. The reason for this restriction of instantiation of variables is that the dependencies (or scoping) of the quantifiers must be somehow reflected in the admissible dependencies of the variables on the free atoms. In our framework, these dependencies are to be captured in a binary relation on the variables and the free atoms, called *variable-condition.*

Indeed, it is sometimes unsound to admit the instantiation of a variable $y^{\mathbb{V}}$ with a term containing a free atom $x^{\mathbb{A}}$ that was introduced later than $y^{\mathbb{V}}$:

**Example 3.1 (Soundness of $\delta^-$-rule)** The formula $\exists y^{\mathbb{B}}.\ \forall x^{\mathbb{B}}.\ (y^{\mathbb{B}} = x^{\mathbb{B}})$ is valid in structures with only one single object, but not in general. We can start a reductive proof attempt of it as follows:

| | | |
|---|---|---|
| $\gamma$-step: | $\forall x^{\mathbb{B}}.\ (y^{\mathbb{V}} = x^{\mathbb{B}})$, | $\exists y^{\mathbb{B}}.\ \forall x^{\mathbb{B}}.\ (y^{\mathbb{B}} = x^{\mathbb{B}})$ |
| $\delta^-$-step: | $(y^{\mathbb{V}} = x^{\mathbb{A}})$, | $\exists y^{\mathbb{B}}.\ \forall x^{\mathbb{B}}.\ (y^{\mathbb{B}} = x^{\mathbb{B}})$ |

Now, if the variable $y^{\mathbb{V}}$ could be instantiated with the free atom $x^{\mathbb{A}}$, then we would get the tautology $(x^{\mathbb{A}} = x^{\mathbb{A}})$, i.e. we would have proved an invalid formula. To prevent this, as indicated to the lower right of the bar of the first of the $\delta^-$-rules, the $\delta^-$-step has to record

$$\mathbb{V}(\forall x^{\mathbb{B}}.\ (y^{\mathbb{V}} = x^{\mathbb{B}}),\ \exists y^{\mathbb{B}}.\ \forall x^{\mathbb{B}}.\ (y^{\mathbb{B}} = x^{\mathbb{B}})) \times \{x^{\mathbb{A}}\} \quad = \quad \{(y^{\mathbb{V}}, x^{\mathbb{A}})\}$$

in a variable-condition, where $(y^{\mathbb{V}}, x^{\mathbb{A}})$ means that the variable $y^{\mathbb{V}}$ must not depend on the free atom $x^{\mathbb{A}}$, or that $y^{\mathbb{V}}$ is somehow "necessarily older" than $x^{\mathbb{A}}$, so that we may never instantiate the variable $y^{\mathbb{V}}$ with a term containing the free atom $x^{\mathbb{A}}$, simply because $x^{\mathbb{A}}$ was not available when $y^{\mathbb{V}}$ was born.

Starting with an empty variable-condition, we extend the variable-condition during proof attempts by $\delta$-steps and by global instantiations of variables. Roughly speaking, such an instantiation is *consistent* if the resulting variable-condition (seen as a directed graph) has *no cycle* after adding, for each variable $y^{\mathbb{V}}$ instantiated with a term $t$ and for each variable or free atom $x^{\mathbb{VA}}$ occurring in $t$, the pair $(x^{\mathbb{VA}}, y^{\mathbb{V}})$.

If we try to instantiate $y^{\mathbb{V}}$ with $x^{\mathbb{A}}$ in our example proof, however, this consistency requirement is violated by the immediate cycle between $y^{\mathbb{V}}$ and $x^{\mathbb{A}}$ and therefore this instantiation is not admissible and the proof attempt fails to show the invalid input proposition. In this way, soundness of the $\delta^{-}$-rules is given by the sets of pairs to their lower right.

## 3.6 $\delta^{+}$-Rules

There are basically two kinds of $\delta$-rules: standard $\delta^{-}$-rules (also simply called *$\delta$-rules*) and $\delta^{+}$-rules (also called *liberalized $\delta$-rules*). They differ in the kind of symbol they introduce and — crucially — in the way they enlarge the variable-condition, depicted to the lower right of the bar:

**$\delta^{+}$-rules:** Let $x^{\mathbb{V}}$ be a fresh variable:

$$\frac{\Gamma \quad \forall x^{\mathbb{B}}.\ A \quad \Pi}{A\{x^{\mathbb{B}}\mapsto x^{\mathbb{V}}\} \quad \Gamma \quad \Pi} \qquad \begin{array}{l}(\ x^{\mathbb{V}},\ \ \varepsilon x^{\mathbb{B}}.\ \neg A\ )\\ \mathbb{VA}(\forall x^{\mathbb{B}}.\ A) \times \{x^{\mathbb{V}}\}\end{array}$$

$$\frac{\Gamma \quad \neg\exists x^{\mathbb{B}}.\ A \quad \Pi}{\neg A\{x^{\mathbb{B}}\mapsto x^{\mathbb{V}}\} \quad \Gamma \quad \Pi} \qquad \begin{array}{l}(\ x^{\mathbb{V}},\ \ \varepsilon x^{\mathbb{B}}.\ A\ )\\ \mathbb{VA}(\neg\exists x^{\mathbb{B}}.\ A) \times \{x^{\mathbb{V}}\}\end{array}$$

While in the, say, first $\delta^{-}$-rule, $\mathbb{V}(\Gamma\ \forall x^{\mathbb{B}}.\ A\ \Pi)$ denotes the set of variables occurring in the *entire* upper sequent, in the first $\delta^{+}$-rule, $\mathbb{VA}(\forall x^{\mathbb{B}}.\ A)$ denotes the set of variables and free atoms occurring in only one of its formulas, namely the *principal formula* $\forall x^{\mathbb{B}}.\ A$. Therefore, the variable-conditions generated by the $\delta^{+}$-rules are typically smaller than the ones generated by the $\delta^{-}$-rules. Smaller variable-conditions permit additional proofs. Indeed, the $\delta^{+}$-rules enable additional proofs on the same level of *$\gamma$-multiplicity* (i.e. the maximal number of repeated $\gamma$-steps applied to the identical principal formula); cf. e.g. [WIRTH, 2004, Example 2.8, p. 21]. For certain classes of theorems, these proofs are exponentially and even non-elementarily shorter than the shortest proofs that apply only $\delta^{-}$-rules; for a short survey see [WIRTH, 2004, § 2.1.5]. Moreover, the $\delta^{+}$-rules provide additional proofs that are not only shorter but also more natural and easier to find, both automatically and for human beings; see the discussion on design goals for inference systems in [WIRTH, 2004, § 1.2.1], and the formal proof of the limit theorem for + in [WIRTH, 2006a; 2012b]. Although we explain in § A of the Appendix why $\delta^{+}$ may not always be more liberal than $\delta^{-}$, the name "liberalized" is justified: $\delta^{+}$-rules provide more freedom to the prover in practice.

Moreover, the pairs indicated to the upper right of the bar of the $\delta^{+}$-rules are to augment another global binary relation besides the variable-condition, namely a function called the *choice-condition.* Roughly speaking, the addition of an element $(x^{\mathbb{V}},\ \varepsilon x^{\mathbb{B}}.\ \neg A)$ to the current choice-condition — as required by the first of the $\delta^{+}$-rules — is to be interpreted as the addition of the equational constraint $x^{\mathbb{V}} = \varepsilon x^{\mathbb{B}}.\ \neg A$. To preserve the soundness of the $\delta^{+}$-step under subsequent global instantiation of the variable $x^{\mathbb{V}}$, this constraint must be observed in such instantiations. What this actually means will be explained in § 4.14.

For a replay of Example 3.1 for the $\delta^{+}$- instead of the $\delta^{-}$-rule, see Example 4.13 in § 4.14.

## 3.7 Global and Practical Aspects of Inference Systems

Assume that our calculus consists in the structural tautologies of the forms $\Gamma\ \ A\ \ \Delta\ \ \neg A\ \ \Pi$, $\Gamma\ \ \neg A\ \ \Delta\ \ A\ \ \Pi$ [as well as $\Gamma\ \ t{=}t\ \ \Delta$], as described in § 3.1, and also in the $\gamma$-rules and the primitive propositional $\beta$- and $\alpha$-rules for all given propositional operators [as well as the primitive contextual-rewriting rule], as described in §§ 3.4, 3.3, and 3.2. Then we can obtain a sound and complete calculus (i.e. a calculus that derives only true formulas and all of them) for classical classical first-order logic [with equality] by adding one of the following three sets of rules:

(1) only the $\delta^{-}$-rules,
(2) only the $\delta^{+}$-rules,
(3) *both* the $\delta^{-}$- and $\delta^{+}$-rules.

Note that, while soundness is a must for any calculus, completeness is a merely theoretical cachet, because the ability to derive any true formula in principle does not mean that we can do so within time. Soundness is typically a local property of a calculus and thus rather easy to show; completeness, however, is always a global property and harder to show. Although completeness proofs are among the favorites of the ivory-tower watchmen, we will stay far away from them in the following, because our overall goals here are merely practical ones, no matter how hard our proofs of strengthened soundness properties may get.

Indeed, even if we have the rare luck that there is a complete calculus (just as the one described by our $\alpha$-, $\beta$-, $\gamma$-, and $\delta$-rules for classical first-order logic), uninformed search with such a calculus may not be more reasonable than novel writing with the British Museum algorithm. Neither recursively enumerating all possible proofs nor all possible sequences of characters, ordered by size, is a practical method by any means — although the former may be what some people still understand under the academic discipline of "automated theorem proving".

Moreover, for more advanced logics, recursive enumerability of the valid formulas cannot always be taken for granted.

Although recursive enumerability is given for classical first-order logic, even for this logic — as it is typical for non-trivial logics — the *invalid* formulas cannot be enumerated, and thus the validity of a formula cannot be decided. This means that any sound prover will have to run forever occasionally in case it tries to prove an invalid formula. This also means that, in general, we cannot give a time limit for running a prover on a valid formula successfully.

But how can this non-enumerability of invalid formulas actually occur with our pretty trivial $\alpha$-, $\beta$-, $\gamma$-, and $\delta$-rules in classical first-order logic? If you do not find out yourself, we will answer this question to you in § 3.8.

Furthermore note that even decidability would not necessary bring us efficiency and tractability.

Last but not least, for advanced practical applications, it is definitely inappropriate to confine oneself in the straitjackets of classical first-order logic without mathematical induction, or of even more restrictive logics, such as "description logics", a name which actually cheats because these languages have only very minor descriptive power.

## 3.8 $\gamma$-Multiplicity

The answer to the question left open in § 3.7 is given by the technical term "*$\gamma$-multiplicity*".

When we apply an $\alpha$-, $\beta$-, or $\delta$-rule to an occurrence of a principal formula, then this occurrence disappears from the leaves of our proof tree. Indeed, such a principal formula is completely replaced with its side formulas and the occurrence of the principal logical operator disappears, such that the application of these rules must terminate.

This is different, however, with the $\gamma$-rules, where the principal formula must not be deleted in the resulting sequent in addition to its side formula — for completeness reasons.

Indeed, repeatedly increasing the multiplicity of the identical $\gamma$-quantified formula cannot be avoided in principle, as shown in the following simple example.

**Example 3.2**
Suppose we want to prove in the natural numbers the one-formula sequent

$$\forall x^{\mathbb{B}}.\ (x^{\mathbb{B}} \leq x^{\mathbb{B}}{+}1)\ \Rightarrow\ 0 \leq 1000.$$

Further suppose that all we know is how to add 1 to the decimal representation of a natural number and that $\leq$ is a reflexive ordering. We can reduce this sequent in one $\alpha$- and two $\gamma$-steps (choosing $t$ to be 0 and 1, resp.) to

$$\neg(0 \leq 1) \quad \neg(1 \leq 2) \quad \neg\forall x^{\mathbb{B}}.\ (x^{\mathbb{B}} \leq x^{\mathbb{B}}{+}1) \quad (0 \leq 1000).$$

For transitivity of $\leq$ to obtain a $\leq$-tautology here, we will need to apply the $\gamma$-rule as before again 998 times to the same $\gamma$-formula $\neg\forall x^{\mathbb{B}}.\ (x^{\mathbb{B}} \leq x^{\mathbb{B}}{+}1)$. Note that this procedure is already informed by our knowledge, which instances of the bound atom $x^{\mathbb{B}}$ will have to be chosen for the proof to succeed. Of course, this is not a state-of-the-art specification of natural numbers, but we have actually experienced that most famous automated-theorem-proving systems typically fail in a very similar way when bigger natural numbers occur.

## 3.9 Skolemization

Note that there is a popular alternative to variable-conditions, namely SKOLEMization, where the $\delta^-$- and $\delta^+$-rules introduce functions (i.e. the logical order of the replacements for the bound atoms is incremented) which are given the variables of $\mathbb{V}(\Gamma\ \ \forall x^{\mathbb{B}}.\ A\ \ \Pi)$ and $\mathbb{V}(\forall x^{\mathbb{B}}.\ A)$ as initial arguments, respectively. Then, the occur-check of unification implements the restrictions on the instantiation of variables, which are required for soundness. In some inference systems, however, SKOLEMization is unsound (e.g. for higher-order systems such as the one in [KOHLHASE, 1998] or the system in [WIRTH, 2004] for *descente infinie*) or inappropriate (e.g. in the matrix systems of [WALLEN, 1990]).

We prefer inference systems that include variable-conditions to inference systems that offer only SKOLEMization. Indeed, this inclusion provides a more general and often simpler approach, which does not necessarily reduce efficiency in any case. Moreover, note that variable-conditions cannot add unnecessary complications here:

- If, in some application, variable-conditions are superfluous, then we can work with empty variable-conditions as if there would be no variable-conditions at all.
- We will need the variable-conditions anyway for our choice-conditions, which again are needed to formalize our approach to HILBERT's $\varepsilon$-operator.

# 4 Introduction to HILBERT's $\varepsilon$

## 4.1 Motivation

HILBERT's $\varepsilon$-symbol is an operator or binder that forms terms, just like PEANO's $\iota$-symbol. Roughly speaking, the term $\varepsilon x^{\mathbb{B}}.\ A$, formed from a bound atom $x^{\mathbb{B}}$ and a formula $A$, denotes *just some* object that is *chosen* such that — if possible — $A$ (seen as a predicate on $x^{\mathbb{B}}$) holds for this object.

For ACKERMANN, BERNAYS, and HILBERT, the $\varepsilon$ was an intermediate tool in proof theory, to be eliminated in the end. Instead of giving a model-theoretic semantics for the $\varepsilon$, they just specified those axioms which were essential in their proof transformations. These axioms did not provide a complete definition, but left the $\varepsilon$ underspecified.

*Descriptive terms* such as $\varepsilon x^{\mathbb{B}}.\ A$ and $\iota x^{\mathbb{B}}.\ A$ are of general interest and applicability. Our more elegant and flexible treatment turns out to be useful in many areas where logic is designed or applied as a tool for description and reasoning.

## 4.2 Requirements Specification

For the usefulness of such descriptive terms we consider the following requirements to be the most important ones.

**Requirement I (Indication of Commitment):**
The syntax must clearly express where exactly a *commitment* to a choice of a particular object is required, and where, to the contrary, different objects corresponding with the description may be chosen for different occurrences of the same descriptive term.

**Requirement II (Reasoning):**
It must be possible to replace a descriptive term with a term that corresponds with its description. The correctness of such a replacement must be expressible and should be verifiable in the original calculus.

**Requirement III (Semantics):**
The semantics should be simple, straightforward, natural, formal, and model-based. Overspecification should be carefully avoided. Furthermore, the semantics should be modular and abstract in the sense that it adds the operator to a variety of logics, independent of the details of a concrete logic.

Our more elegant and flexible, indefinite treatment of the $\varepsilon$-operator is compatible with HILBERT's original one and satisfies these requirements. As it involves novel semantic techniques, it may also serve as the paradigm for the design of similar operators.

## 4.3 Overview

In § B of the Appendix, the reader can find an update of our review form [WIRTH, 2008; 2006b] of the literature on extended semantics given to HILBERT's $\varepsilon$-operator in the $2^{\text{nd}}$ half of the $20^{\text{th}}$ century. In the current § 4, we will now introduce to the $\iota$ and the $\varepsilon$ (§§ 4.4 and 4.5), to the $\varepsilon$'s proof-theoretic origin (§ 4.6), and to our more general semantic objective (§ 4.8) with its emphasis on *indefinite* and *committed choice* (§ 4.9).

## 4.4 From the $\iota$ to the $\varepsilon$

As the $\varepsilon$-operator was developed as an improvement over the still very popular $\iota$-operator, a careful discussion of the $\iota$ in this section is required for a deeper understanding of the $\varepsilon$.

### 4.4.1 The Symbols for the $\iota$-Operator

The probably first descriptive $\iota$-operator occurs in [FREGE, 1893/1903, Vol. I], written as a boldface backslash. As a *boldface* version of the backslash is not easily available in standard typesetting, we will use a simple backslash "\" in § 4.4.4.

A slightly different $\iota$-operator occurs in [PEANO, 1896f.], written as "$\bar{\iota}$", i.e. as an overlined $\iota$. In its German translation [PEANO, 1899b], we also find an alternative symbol with the same denotation, namely an upside-down $\iota$-symbol. Both symbols are meant to indicate the inverse of PEANO's $\iota$-function, which constructs the set of its single argument.

Nowadays, however, "$\{y\}$" is written for PEANO's "$\iota y$", and thus — as a simplifying convention to avoid problems in typesetting and automatic indexing — a simple $\iota$ should be used to designate the descriptive $\iota$-operator, without overlining or inversion.

### 4.4.2 The Essential Idea of the $\iota$-Operator

Let us define the quantifier of *unique existence* by

$$\exists! x^{\mathbb{B}}.\ A \quad := \quad \exists y^{\mathbb{B}}.\ \forall x^{\mathbb{B}}.\ ((y^{\mathbb{B}}{=}x^{\mathbb{B}}) \Leftrightarrow A),$$

for some fresh $y^{\mathbb{B}}$. All the slightly differing specifications of the $\iota$-operator agree in the following point: If there is the unique $x^{\mathbb{B}}$ such that the formula $A$ (seen as a predicate on $x^{\mathbb{B}}$) holds, then the $\iota$-term $\iota x^{\mathbb{B}}.\ A$ denotes this unique object:

$$\exists! x^{\mathbb{B}}.\ A \quad \Rightarrow \quad A\{x^{\mathbb{B}} \mapsto \iota x^{\mathbb{B}}.\ A\} \qquad (\iota_0)$$

or in different notation $(\exists! x^{\mathbb{B}}.\ (A(x^{\mathbb{B}}))) \quad \Rightarrow \quad A(\iota x^{\mathbb{B}}.\ (A(x^{\mathbb{B}})))$.

**Example 4.1 ($\iota$-operator)**
For an informal introduction to the $\iota$-operator, consider $\mathsf{Father}$ to be a predicate for which $\mathsf{Father}(\mathsf{Heinrich\,III}, \mathsf{Heinrich\,IV})$ holds, i.e. "Heinrich III is father of Heinrich IV".
Now, "*the* father of Heinrich IV" is designated by $\iota x^{\mathbb{B}}.\ \mathsf{Father}(x^{\mathbb{B}}, \mathsf{Heinrich\,IV})$, and because this is nobody but Heinrich III, i.e. $\mathsf{Heinrich\,III} = \iota x^{\mathbb{B}}.\ \mathsf{Father}(x^{\mathbb{B}}, \mathsf{Heinrich\,IV})$, we know that $\mathsf{Father}(\iota x^{\mathbb{B}}.\ \mathsf{Father}(x^{\mathbb{B}}, \mathsf{Heinrich\,IV}), \mathsf{Heinrich\,IV})$. Similarly,

$$\mathsf{Father}(\iota x^{\mathbb{B}}.\ \mathsf{Father}(x^{\mathbb{B}}, \mathsf{Adam}), \mathsf{Adam}), \qquad (4.1.1)$$

and thus $\exists y^{\mathbb{B}}.\ \mathsf{Father}(y^{\mathbb{B}}, \mathsf{Adam})$, but, oops! Adam and Eve do not have any fathers.
If you do not agree, you probably appreciate the following problem that occurs when somebody has God as an additional father.

$$\mathsf{Father}(\mathsf{Holy\,Ghost}, \mathsf{Jesus}) \quad \wedge \quad \mathsf{Father}(\mathsf{Joseph}, \mathsf{Jesus}). \qquad (4.1.2)$$

Then the Holy Ghost is *the* father of Jesus and Joseph is *the* father of Jesus:

$$\mathsf{Holy\,Ghost} = \iota x^{\mathbb{B}}.\ \mathsf{Father}(x^{\mathbb{B}}, \mathsf{Jesus}) \ \wedge\ \mathsf{Joseph} = \iota x^{\mathbb{B}}.\ \mathsf{Father}(x^{\mathbb{B}}, \mathsf{Jesus}) \qquad (4.1.3)$$

This implies something *the* Pope may not accept, namely $\mathsf{Holy\,Ghost} = \mathsf{Joseph}$,
and he anathematized Heinrich IV in the year 1076:

$$\mathsf{Anathematized}(\iota x^{\mathbb{B}}.\ \mathsf{Pope}(x^{\mathbb{B}}), \mathsf{Heinrich\,IV}, 1076). \qquad (4.1.4)$$

### 4.4.3 Elementary Semantics Without Straightforward Overspecification

Semantics without a straightforward form of overspecification can be given to the $\iota$-terms in the following three elementary ways:

**Russell's non-referring $\iota$-operator** [RUSSELL, 1905]

In *Principia Mathematica* [1910–1913] by BERTRAND RUSSELL and ALFRED NORTH WHITEHEAD, an $\iota$-term is given a meaning only in form of quantifications over contexts $C[\cdots]$ of the occurrences of the $\iota$-term, *mutatis mutandis* by

$$C[\iota x^{\mathbb{B}}.\ A] \text{ is defined as a short form for } \exists y^{\mathbb{B}}.\ \big(\forall x^{\mathbb{B}}.\ \big((y^{\mathbb{B}}{=}x^{\mathbb{B}}) \Leftrightarrow A\big) \wedge C[y^{\mathbb{B}}]\big).$$

This definition is peculiar because the *definiens* is not of the expected form $C[t]$ (for some term $t$), and because an $\iota$-term on its own — i.e. without a context $C[\cdots]$ — cannot *directly refer* to an object that it may be intended to denote.

This was first presented as a linguistic theory of descriptions in [RUSSELL, 1905] — but without using any symbol for the $\iota$.

RUSSELL's *On denoting* [1905] became so popular that the term "non-referring" had to be introduced to make aware of the fact that RUSSELL's $\iota$-terms are not denoting (in spite of the title), and that RUSSELL's theory of descriptions ignores the fundamental reference aspect of descriptive terms, cf. STRAWSON's *On referring* [1950].

**Hilbert–Bernays' presuppositional $\iota$-operator** [HILBERT & BERNAYS, 1934]

To overcome the complex difficulties of RUSSELL's non-referring semantics, in § 8 of the first volume of the two-volume monograph *Foundations of Mathematics* (*Grundlagen der Mathematik,* 1st edn. 1934, 2nd edn. 1968) by DAVID HILBERT and PAUL BERNAYS, a completed proof of $\exists! x^{\mathbb{B}}.\ A$ is required to precede each formation of a term $\iota x^{\mathbb{B}}.\ A$, which otherwise is not considered a well-formed term at all.

This way of defining the $\iota$ is nowadays called *"presuppositional".* This word occurs in relation to HILBERT–BERNAYS' $\iota$ in [SLATER, 2007a] and [SLATER, 2009, §§ 1, 6, and 8f.], but it does not occur in [STRAWSON, 1950], and we do not know where it occurs first with this meaning.

**Peano's partially specified $\iota$-operator** [PEANO, 1896f.]

Since HILBERT–BERNAYS' presuppositional treatment makes the $\iota$ quite impractical and the formal syntax of logic undecidable in general, in § 1 of the second volume of HILBERT–BERNAYS' *Foundations of Mathematics* (1st edn. 1939, 2nd edn. 1970), HILBERT's $\varepsilon$, however, is already given a more flexible treatment: The simple idea is to leave the $\varepsilon$-terms uninterpreted. This will be described below. In this paper, we will present this more flexible treatment also for the $\iota$.

After all, this treatment is the original one of PEANO's $\iota$, found already in the article *Studii di Logica Matematica* [1896f.] by GUISEPPE PEANO.[1]

It should come as no surprise that PEANO (unlike RUSSELL and HILBERT–BERNAYS!) invented the only practical specification of $\iota$-terms: After all he was most interested in written languages for specification and communication — but hardly in calculi — and created also the spoken artificial language *Latino sine flexione*, cf. e.g. [KENNEDY, 2002].

Moreover, by the partiality of his specification, PEANO avoided also the other pitfall, namely overspecification, and all its unintended consequences (unlike FREGE and QUINE, cf. § 4.4.4). As the symbol "$\bar{\iota}$" was invented by PEANO as well (cf. § 4.4.1), we have good reason to speak of "PEANO's $\iota$", at least as much as we have reason to speak of "HILBERT's $\varepsilon$".

It must not be overlooked that PEANO's $\iota$ — in spite of its partiality — always denotes: It is not a partial operator, it is just partially specified.

At least in non-modal classical logics, it is a well justified standard that *each term denotes.* More precisely — in each model or structure $\mathcal{S}$ under consideration — each occurrence of a proper term denotes an object in the universe of $\mathcal{S}$. Following that standard, to be able to write down $\iota x^{\mathbb{B}}.\, A$ without further consideration, we have to treat $\iota x^{\mathbb{B}}.\, A$ as an uninterpreted term about which we only know axiom $(\iota_0)$ from § 4.4.2.

With $(\iota_0)$ as the only axiom for the $\iota$, the term $\iota x^{\mathbb{B}}.\, A$ has to satisfy $A$ (seen as a predicate on $x^{\mathbb{B}}$) only if there exists a unique object such that $A$ holds for it. The price, however, we have to pay for the avoidance of non-referringness, presuppositionality, and overspecification is that — roughly speaking — the term $\iota x^{\mathbb{B}}.\, A$ is of no use unless the unique existence $\exists! x^{\mathbb{B}}.\, A$ can be derived.

Finally, let us come back to Example 4.1 of § 4.4.2. The problems presented there do not actually appear if $(\iota_0)$ is the whole specification for the $\iota$, because then (4.1.1), (4.1.3), and (4.1.4) are not generally valid. Indeed, the description of (4.1.1) lacks existence and the descriptions of (4.1.3) and (4.1.4) lack uniqueness.

### 4.4.4 Overspecified $\iota$-Operators

From FREGE to QUINE, we find a multitude of $\iota$-operators with definitions that overspecify the $\iota$ in different ways for the sake of *complete definedness* and *syntactic eliminability.*

As we already stated in Requirement III (Semantics) of § 4.2, overspecification should be carefully avoided. Indeed, any overspecification leads to puzzling, arbitrary consequences, which may cause harm to the successful application of descriptive operators in practice.

---

[1](History of Peano and his $\iota$)
In [PEANO, 1896f.], PEANO wrote $\bar{\iota}$ instead of the $\iota$ of Example 4.1, and $\bar{\iota}\{\, x \mid A \,\}$ instead of $\iota x.\, A$. (We have changed the class notation to modern standard here: PEANO actually wrote $\overline{x \varepsilon A}$ instead of $\{\, x \mid A \,\}$.) The bar above the $\iota$ (just as the alternative inversion of the $\iota$) were to indicated that $\bar{\iota}$ was implicitly defined as the inverse of the operator $\iota$ defined by $\iota y := \{y\}$, which occurred already in [PEANO, 1890] and still in [QUINE, 1981].
The definition of $\bar{\iota}$ reads literally [PEANO, 1896f., Definition 22]:

$$a \in K \,.\, \exists a : x, y \in a \,.\, \supset_{x,y} .\, x = y : \supset : x = \bar{\iota} a \,.=.\, a = \iota x$$

This straightforwardly translates into more modern notation as follows:

$$\text{For any class } a\text{:} \quad a \neq \emptyset \wedge \forall x, y.\ (x, y \in a \ \Rightarrow\ x = y) \ \Rightarrow\ \forall x.\ (x = \bar{\iota} a \ \Leftrightarrow\ a = \iota x)$$

Giving up the flavor of an explicit definition of "$x = \bar{\iota} a$", this can be simplified to the following equivalent form:

$$\text{For any class } a\text{:} \quad \exists! x.\ x \in a \ \Rightarrow\ \bar{\iota} a \in a \qquad (\bar{\iota}_0)$$

Besides notational difference, this is $(\iota_0)$ of our § 4.4.2.

**Frege's haphazardly overspecified $\iota$-operator** [FREGE, 1893/1903]

The first occurrence of a descriptive $\iota$-operator in the literature seems to be in 1893, namely in § 11 of the first volume of the two-volume monograph *Grundgesetze der Arithmetik — Begriffsschriftlich abgeleitet* [1893/1903] by GOTTLOB FREGE:

For $A$ seen as a function from objects to truth values, $\backslash A$ (in our notation $\iota x^{\mathbb{B}}.\ A$) is defined to be the object $\Delta$ if $A$ is extensionally equal to the function that checks for equality to $\Delta$, i.e. if $A = \lambda x^{\mathbb{B}}.\ (\Delta = x^{\mathbb{B}})$.

In the case that there is no such $\Delta$, FREGE overspecified his $\iota$-operator pretty haphazardly by defining $\backslash A$ to be $A$, which is not even an object, but a function.

(Note that FREGE actually wrote an $\varepsilon$ (having nothing to do with the $\varepsilon$-operator) instead of our $x^{\mathbb{B}}$, and a *spiritus lenis* over it instead of a modern $\lambda$-operator before and a dot after it. Moreover, he wrote a $\xi$ for the $A$.)

**Quine's overspecified $\iota$-operator** [QUINE, 1981]

In set theories without urelements, such as in [QUINE, 1981], the $\iota$-operator can be defined by something like

$$\iota x^{\mathbb{B}}.\ A \quad := \quad \left\{\ z^{\mathbb{B}}\ \middle|\ \exists y^{\mathbb{B}}.\ \big(\ \forall x^{\mathbb{B}}.\ ((y^{\mathbb{B}}{=}x^{\mathbb{B}}) \Leftrightarrow A) \quad \wedge \quad z^{\mathbb{B}} \in y^{\mathbb{B}}\ \big)\ \right\},$$

for fresh $y^{\mathbb{B}}$ and $z^{\mathbb{B}}$.

This is again an overspecification resulting in $\iota x^{\mathbb{B}}.\, A \ = \ \emptyset$ if there is no such $y^{\mathbb{B}}$ (which otherwise is always unique).

### 4.4.5 A Completely Defined, but Not Overspecified $\iota$-Operator

The complete definitions of the $\iota$ in § 4.4.4 take place in *possibly inconsistent* logical frameworks, namely FREGE's Begriffsschrift and QUINE's version of standard set theory without urelements.

That neither overspecification nor possible inconsistency is necessary for complete definitions of the $\iota$ is witnessed by the following complete, but non-elementary definition of the $\iota$, which is also referring and non-presuppositional.

**The $\varepsilon$-calculus' $\iota$-operator** [HILBERT & BERNAYS, 1939]

In the $\varepsilon$-calculus, which is a conservative extension of first-order predicate calculus, first elaborated in the second volume of HILBERT–BERNAYS' *Foundations of Mathematics* [1939], we can define the $\iota$ simply by

$$\iota x^{\mathbb{B}}.\ A \quad := \quad \varepsilon y^{\mathbb{B}}.\ \forall x^{\mathbb{B}}.\ ((y^{\mathbb{B}} = x^{\mathbb{B}}) \Leftrightarrow A)$$

(for a fresh $y^{\mathbb{B}}$), i.e. as a unique $x^{\mathbb{B}}$ such that $A$ holds (provided there is such an $x^{\mathbb{B}}$).

Note that the simple definition $\iota x^{\mathbb{B}}.\ A \ := \ \varepsilon x^{\mathbb{B}}.\ A$, however, would already be an overspecification in case $A$ has multiple solutions.

This definition is non-elementary, however, in the sense that it introduces $\varepsilon$-terms, which cannot be eliminated in first-order logic in general.

If the $\varepsilon$ is given, this definition is the most useful and elegant way to introduce the $\iota$, although it is somehow *ex eventu,* because the development of the $\varepsilon$ was started two dozen years after the first publications on FREGE's and PEANO's $\iota$-operators.

## 4.5 The $\varepsilon$ as an Improvement over the $\iota$

Compared to the $\iota$, the $\varepsilon$ is more useful because — instead of $(\iota_0)$ — it comes with the stronger axiom

$$\exists x^{\mathbb{B}}.\ A \ \Rightarrow\ A\{x^{\mathbb{B}} \mapsto \varepsilon x^{\mathbb{B}}.\ A\} \qquad (\varepsilon_0)$$

More specifically, as the formula $\exists x^{\mathbb{B}}.\ A$ (which has to be true to guarantee an interpretation of the $\varepsilon$-term $\varepsilon x^{\mathbb{B}}.\ A$ that is meaningful in the sense that it satisfies its formula $A$) is weaker than the corresponding formula $\exists! x^{\mathbb{B}}.\ A$ (for the respective $\iota$-term), the area of useful application is wider for the $\varepsilon$- than for the $\iota$-operator. Indeed, we have already seen in § 4.4.5 that the $\iota$ can be defined in terms of the $\varepsilon$, but not vice versa.

Moreover, in case of $\exists! x^{\mathbb{B}}.\ A$, the $\varepsilon$-operator picks the same element as the $\iota$-operator:

$$\exists! x^{\mathbb{B}}.\ A \ \Rightarrow\ \big(\ \varepsilon x^{\mathbb{B}}.\ A \ =\ \iota x^{\mathbb{B}}.\ A\ \big).$$

Thus, unless eliminability is relevant, we should replace all useful occurrences of the $\iota$ with the $\varepsilon$: As a consequence, among other advantages, the arising proof obligations become weaker and both human and automated generation and generalization of proofs become more efficient.

## 4.6 On the $\varepsilon$'s Proof-Theoretic Origin

### 4.6.1 The $\varepsilon$-Formula and the Historical Sources of the $\varepsilon$

The main historical source on the $\varepsilon$ is the second volume of the *Foundations of Mathematics* [Hilbert & Bernays, 1934; 1939; 1968; 1970], the fundamental work which summarizes the foundational and proof-theoretic contributions of David Hilbert and his mathematical-logic group.

The preferred specification for Hilbert's $\varepsilon$ in proof-theoretic investigations is not the axiom $(\varepsilon_0)$, but *mutatis mutandis* the following formula:

$$A\{x^{\mathbb{B}} \mapsto x^{\mathbb{A}}\} \ \Rightarrow\ A\{x^{\mathbb{B}} \mapsto \varepsilon x^{\mathbb{B}}.\ A\} \qquad (\varepsilon\text{-}formula)$$

The $\varepsilon$-formula is equivalent to $(\varepsilon_0)$, but gets along without any quantifier.

The name "$\varepsilon$-formula" originates in [Hilbert & Bernays, 1939, p. 13], where the $\varepsilon$-operator is simply called *Hilbert's $\varepsilon$-symbol.*

For historical correctness, note that the notation in the original is closer to

$$A^{\mathbb{A}}(x^{\mathbb{A}}) \Rightarrow A^{\mathbb{A}}(\varepsilon x^{\mathbb{B}}.\ A^{\mathbb{A}}(x^{\mathbb{B}})),$$

where the $A^{\mathbb{A}}$ is a concrete singulary predicate atom (called *formula variable* in the original) and comes with several extra rules for its instantiation, cf. [Hilbert & Bernays, 1939, p. 13f.]. *Verbatim* the notation actually is

$$A(a) \Rightarrow A(\varepsilon_x\, A(x)),$$

and the deductive equivalence to $(\varepsilon_0)$, i.e. *verbatim* to

$$(Ex)\, A(x) \Rightarrow A(\varepsilon_x\, A(x)),$$

is straightforward, cf. [Hilbert & Bernays, 1939, pp. 13–15]. In our notation, however, $(\varepsilon_0)$ and the $\varepsilon$-formula are axiom *schemata* where the $A$ is a meta-variable for a formula.

Note that those concrete formulas, which may be denoted by our meta-variable $A$ and which likewise may be used as a substitution for HILBERT–BERNAYS' formula variable $A$, may contain extra occurrences of $x^{\mathbb{A}}$ and $a$, respectively, beside the one explicitly indicated in the $\varepsilon$-formula. The instances of the $\varepsilon$-formula with such extra occurrences, however, do not add to its strength, because they follow from the versions with fresh symbols at these occurrences by instantiation of the fresh symbols with $x^{\mathbb{A}}$ in our framework and with $a$ in HILBERT–BERNAYS' predicate calculus, respectively.

The $\varepsilon$-formula already occurs, however under different names, in the pioneering papers on the $\varepsilon$, i.e. in [ACKERMANN, 1925] as "transfinite axiom 1", in [HILBERT, 1926] as "axiom of choice" (in the operator form $A(a) \Rightarrow A(\varepsilon A)$, where the $\varepsilon$ is called "transfinite logical choice function"), and in [HILBERT, 1928] as "logical $\varepsilon$-axiom" (again in operator form, where the $\varepsilon$ is called "logical $\varepsilon$-function").

### 4.6.2 The Original Explanation of the $\varepsilon$

As the basic methodology of HILBERT's program is to treat all symbols as meaningless, no semantics is required besides the one given by the single axiom ($\varepsilon_0$). To further the understanding, however, we read on p.12 of [HILBERT & BERNAYS, 1939; 1970]:

> $\varepsilon x^{\mathbb{B}}.\ A$ ... "ist ein Ding des Individuenbereichs, und zwar ist dieses Ding gemäß der inhaltlichen Übersetzung der Formel ($\varepsilon_0$) *ein solches, auf das jenes Prädikat A zutrifft, vorausgesetzt, daß es überhaupt auf ein Ding des Individuenbereichs zutrifft.*"
>
> $\varepsilon x^{\mathbb{B}}.\ A$ ... "is a thing of the domain of individuals for which — according to the contentual translation of the formula ($\varepsilon_0$) — *the predicate A holds, provided that A holds for any thing of the domain of individuals at all.*"

**Example 4.2 ($\varepsilon$ instead of $\iota$)** *(continuing Example 4.1 of § 4.4.2)*
Just as for the $\iota$, for the $\varepsilon$ we have $\mathsf{Heinrich\,III} = \varepsilon x^{\mathbb{B}}.\ \mathsf{Father}(x^{\mathbb{B}}, \mathsf{Heinrich\,IV})$ and $\mathsf{Father}(\varepsilon x^{\mathbb{B}}.\ \mathsf{Father}(x^{\mathbb{B}}, \mathsf{Heinrich\,IV}), \mathsf{Heinrich\,IV})$.
But, from the contrapositive of ($\varepsilon_0$) and $\neg\mathsf{Father}(\varepsilon x^{\mathbb{B}}.\ \mathsf{Father}(x^{\mathbb{B}}, \mathsf{Adam}), \mathsf{Adam})$, we now conclude that $\neg\exists y^{\mathbb{B}}.\ \mathsf{Father}(y^{\mathbb{B}}, \mathsf{Adam})$.

### 4.6.3 Defining the Quantifiers via the $\varepsilon$

HILBERT and BERNAYS did not need any semantics or precise intention for the $\varepsilon$-symbol because it was introduced merely as a formal syntactic device to facilitate proof-theoretic investigations, motivated by the possibility to get rid of the existential and universal quantifiers via two direct consequences of axiom ($\varepsilon_0$):

$$\exists x^{\mathbb{B}}.\ A \quad \Leftrightarrow \quad A\{x^{\mathbb{B}} \mapsto \varepsilon x^{\mathbb{B}}.\ A\} \tag{$\varepsilon_1$}$$

$$\forall x^{\mathbb{B}}.\ A \quad \Leftrightarrow \quad A\{x^{\mathbb{B}} \mapsto \varepsilon x^{\mathbb{B}}.\ \neg A\} \tag{$\varepsilon_2$}$$

These equivalences can be seen as definitions of the quantifiers because innermost rewriting with $(\varepsilon_1)$, $(\varepsilon_2)$ yields a normal form after as many steps as there are quantifiers in the input formula. Moreover, also arbitrary rewriting is confluent and terminating, cf. [WIRTH, 2016; 2017a].

It should be noted, however, that rewriting with $(\varepsilon_1)$, $(\varepsilon_2)$ must not be taken for granted *under* modal operators, at least not under the assumption that $\varepsilon$-terms are to remain *rigid,* i.e. independent in their interpretation from their modal contexts. For this assumption there are very good reasons, nicely explained e.g. in [SLATER, 2007a; 2009].

**Example 4.3** Consider the first-order modal logic formula $\Box\exists x^{\mathbb{B}}.\ A$. Moreover, to simplify matters, let us assume that we have constant domains, i.e. that all modal contexts have the same domain of individuals.

Under this condition and for a formula of this structure, it is suggested in [SLATER, 2007a, p.153] to apply $(\varepsilon_1)$ to the considered formula, resulting in $\Box A\{x^{\mathbb{B}} \mapsto \varepsilon x^{\mathbb{B}}.\ A\}$, from which we can doubtlessly conclude $\exists x^{\mathbb{B}}.\ \Box A$, e.g. by Formula (a) in § 4.6.4.

Let us interpret the $\Box$ as "believes" and $A$ as "$x^{\mathbb{B}}$ is the number of rice corns in my car", and let our constant domain be the one of the standard model of the natural numbers. Note that I do not believe of any concrete and definite number that it numbers the rice corns in my car just because I believe that their number is finite.

This interpretation shows that our rewriting with $(\varepsilon_1)$ *under* the operator $\Box$ is incorrect for modal logic in general, at least for rigid $\varepsilon$-terms.

On the other hand, rewriting with $(\varepsilon_1)$, $(\varepsilon_2)$ *above* modal operators is uncritical: $\exists x^{\mathbb{B}}.\ \Box A$ is indeed equivalent to $\Box A\{x^{\mathbb{B}} \mapsto \varepsilon x^{\mathbb{B}}.\ \Box A\}$.

### 4.6.4 The $\varepsilon$-Theorems

When we remove all quantifiers in a derivation of the HILBERT-style predicate calculus of the *Foundations of Mathematics* along $(\varepsilon_1)$ and $(\varepsilon_2)$, the following transformations occur:

Tautologies are turned into tautologies.

The axioms

$$A\{x^{\mathbb{B}}\mapsto x^{\mathbb{A}}\} \quad\Rightarrow\quad \exists x^{\mathbb{B}}.\ A \qquad (\textit{Formula (a)})$$

and

$$\forall x^{\mathbb{B}}.\ A \quad\Rightarrow\quad A\{x^{\mathbb{B}}\mapsto x^{\mathbb{A}}\} \qquad (\textit{Formula (b)})$$

(cf. p. 100f. of [HILBERT & BERNAYS, 1934] or on p. 99f. of [HILBERT & BERNAYS, 1968; 2017b]) are turned into the $\varepsilon$-formula (cf. § 4.6.1) and, roughly speaking, its contrapositive, respectively. Indeed, for the case of Formula (b), we can replace first all $A$ with $\neg A$, and after applying $(\varepsilon_2)$, replace $\neg\neg A$ with $A$, and thus obtain the contrapositive of the $\varepsilon$-formula.

The inference steps are turned into inference steps: the inference schema into the inference schema; the substitution rule for free atoms as well as quantifier introduction (Schemata $(\alpha)$ and $(\beta)$ on p. 103f. of [HILBERT & BERNAYS, 1934] or on p. 102f. of [HILBERT & BERNAYS, 1968; 2017b]) into the substitution rule including $\varepsilon$-terms.

Finally, the $\varepsilon$-formula is taken as a new axiom scheme instead of $(\varepsilon_0)$ because it has the advantage of being free of quantifiers.

The argumentation of the previous paragraphs is actually part of the proof transformation that constructively proves the first of HILBERT–BERNAYS' two theorems on $\varepsilon$-elimination in first-order logic, the so-called *1st $\varepsilon$-Theorem.* In its sharpened form, this theorem can be stated as follows. Note that the original speaks of "bound variables" instead of "bound atoms" and of "formula variables" instead of "predicate atoms", because what we call "variables" is not part of the formula languages of HILBERT–BERNAYS.

**Theorem 4.4 (Sharpened 1st $\varepsilon$-Thm.)** (p.79f. of [HILBERT & BERNAYS, 1939; 1970])
*From a derivation of $\exists x^{\mathbb{B}}_1. \ldots \exists x^{\mathbb{B}}_r. A$ (containing no bound atoms besides the ones bound by the prefix $\exists x^{\mathbb{B}}_1. \ldots \exists x^{\mathbb{B}}_r.$ ) from the formulas $P_1, \ldots, P_k$ (containing neither predicate atoms nor bound atoms) in the predicate calculus (incl. the $\varepsilon$-formula and $=$-substitutability as axiom schemes, plus $=$-reflexivity), we can construct a (finite) disjunction of the form $\bigvee_{i=0}^{s} A\{x^{\mathbb{B}}_1, \ldots, x^{\mathbb{B}}_r \mapsto t_{i,1}, \ldots, t_{i,r}\}$ and a derivation of it*

- *in which bound atoms do not occur at all*
- *from $P_1, \ldots, P_k$ and $=$-axioms (containing neither predicate atoms nor bound atoms)*
- *in the quantifier-free predicate calculus*
  *(i.e. tautologies plus the inference schema and the substitution rule).*

*Note that $r, s$ range over natural numbers including $0$, and that $A$, $t_{i,j}$, and $P_i$ are $\varepsilon$-free because otherwise they would have to include (additional) bound atoms.*

Moreover, the *2nd $\varepsilon$-Theorem* (in [HILBERT & BERNAYS, 1939; 1970]) states that the $\varepsilon$ (just as the $\iota$, cf. [HILBERT & BERNAYS, 1934; 1968]) is a conservative extension of the predicate calculus in the sense that each formal proof of an $\varepsilon$-free formula can be transformed into a formal proof that does not use the $\varepsilon$ at all.

For logics different from classical axiomatic first-order predicate logic, however, it is not necessarily a conservative extension when we add the $\varepsilon$ either with ($\varepsilon_0$), with ($\varepsilon_1$), or with the $\varepsilon$-formula to other first-order logics — may they be weaker such as *intuitionistic* first-order logic,[2] or stronger such as first-order set theories with axiom schemes over arbitrary terms *including the $\varepsilon$*; cf. [WIRTH, 2008, § 3.1.3]. Moreover, even in classical first-order logic there is no translation from the formulas containing the $\varepsilon$ to formulas not containing it.

[2](Consequences of the $\varepsilon$-Formula in Intuitionistic Logic)
Adding the $\varepsilon$ either with ($\varepsilon_0$), with ($\varepsilon_1$), or with the $\varepsilon$-formula (cf. § 4.6) to intuitionistic first-order logic is equivalent on the $\varepsilon$-free fragment to adding PLATO*'s Principle,* i.e. $\exists y^{\mathbb{B}}. (\exists x^{\mathbb{B}}. A \;\Rightarrow\; A\{x^{\mathbb{B}} \mapsto y^{\mathbb{B}}\})$ with $y^{\mathbb{B}}$ not occurring in $A$, cf. [MEYER-VIOL, 1995, § 3.3].
Moreover, the non-trivial direction of ($\varepsilon_2$) is $\forall x^{\mathbb{B}}. A \;\Leftarrow\; A\{x^{\mathbb{B}} \mapsto \varepsilon x^{\mathbb{B}}. \neg A\}$.
Even intuitionistically, this entails its contrapositive $\neg\forall x^{\mathbb{B}}. A \;\Rightarrow\; \neg A\{x^{\mathbb{B}} \mapsto \varepsilon x^{\mathbb{B}}. \neg A\}$,
and then, e.g. by the trivial direction of ($\varepsilon_1$) (when $A$ is replaced with $\neg A$)

$$\neg\forall x^{\mathbb{B}}. A \;\Rightarrow\; \exists x^{\mathbb{B}}. \neg A \tag{Q2}$$

which is not valid in intuitionistic logic in general. Thus, in intuitionistic logic, the universal quantifier becomes strictly weaker by the inclusion of ($\varepsilon_2$) or anything similar for the universal quantifier, such as HILBERT's $\tau$-operator (cf. [HILBERT, 1923a]). More specifically, adding

$$\forall x^{\mathbb{B}}. A \;\Leftarrow\; A\{x^{\mathbb{B}} \mapsto \tau x^{\mathbb{B}}. A\} \tag{$\tau_0$}$$

is equivalent on the $\tau$-free theory to adding $\exists y^{\mathbb{B}}. (\forall x^{\mathbb{B}}. A \;\Leftarrow\; A\{x^{\mathbb{B}} \mapsto y^{\mathbb{B}}\})$ with $y^{\mathbb{B}}$ not occurring in $A$, which again implies (Q2), cf. [MEYER-VIOL, 1995, § 3.4.2].
From a semantic point of view (cf. [GABBAY, 1981]), the intuitionistic $\forall$ may be eliminated, however, by first applying the GÖDEL translation into the modal logic S4 with classical $\forall$ and $\neg$, cf. e.g. [FITTING, 1999], and then adding the $\varepsilon$ conservatively, e.g. by avoiding substitutions via $\lambda$-abstraction as in [FITTING, 1975].

## 4.7 Quantifier Elimination and Subordinate $\varepsilon$-terms

Before we can introduce to our treatment of the $\varepsilon$, we had better get a bit more acquainted with the $\varepsilon$ in general and its well-known features.

The elimination of $\forall$- and $\exists$-quantifiers with the help of $\varepsilon$-terms (cf. § 4.6) may be more difficult than expected when some $\varepsilon$-terms become "subordinate" to others.

**Definition 4.5 (Subordinate)** An $\varepsilon$-term $\varepsilon v^{\mathbb{B}}.\ B$ (or, more generally, a binder on $v^{\mathbb{B}}$ together with its scope $B$) is *superordinate* to an (occurrence of an) $\varepsilon$-term $\varepsilon x^{\mathbb{B}}.\ A$ if

1. $\varepsilon x^{\mathbb{B}}.\ A$ is a subterm of $B$ and
2. an occurrence of the bound atom $v^{\mathbb{B}}$ in $\varepsilon x^{\mathbb{B}}.\ A$ is free in $B$
(i.e. the binder on $v^{\mathbb{B}}$ binds an occurrence of $v^{\mathbb{B}}$ in $\varepsilon x^{\mathbb{B}}.\ A$ ).

An (occurrence of an) $\varepsilon$-term $\varepsilon x^{\mathbb{B}}.\ A$ is *subordinate* to an $\varepsilon$-term $\varepsilon v^{\mathbb{B}}.\ B$ (or, more generally, to a binder on $v^{\mathbb{B}}$ together with its scope $B$) if $\varepsilon v^{\mathbb{B}}.\ B$ is superordinate to $\varepsilon x^{\mathbb{B}}.\ A$.

On p. 24 of [Hilbert & Bernays, 1939; 1970], these subordinate $\varepsilon$-terms, which are responsible for the difficulty to prove the $\varepsilon$-theorems constructively, are called "untergeordnete $\varepsilon$-Ausdrücke". Note that — contrary to Hilbert–Bernays — we do not use a special name for $\varepsilon$-terms with free occurrences of bound atoms here — such as "$\varepsilon$-Ausdrücke" ("$\varepsilon$-expressions" or "quasi $\varepsilon$-terms") instead of "$\varepsilon$-Terme" ("$\varepsilon$-terms") — but simply call them "$\varepsilon$-terms" as well.

**Example 4.6 (Quantifier Elimination and Subordinate $\varepsilon$-Terms)**
Let us repeat the formulas ($\varepsilon_1$) and ($\varepsilon_2$) from § 4.6 here:

$$\exists x^{\mathbb{B}}.\ A \quad\Leftrightarrow\quad A\{x^{\mathbb{B}} \mapsto \varepsilon x^{\mathbb{B}}.\ A\} \qquad (\varepsilon_1)$$
$$\forall x^{\mathbb{B}}.\ A \quad\Leftrightarrow\quad A\{x^{\mathbb{B}} \mapsto \varepsilon x^{\mathbb{B}}.\ \neg A\} \qquad (\varepsilon_2)$$

Let us consider the formula

$$\exists w^{\mathbb{B}}.\ \forall x^{\mathbb{B}}.\ \exists y^{\mathbb{B}}.\ \forall z^{\mathbb{B}}.\ \mathsf{P}(w^{\mathbb{B}}, x^{\mathbb{B}}, y^{\mathbb{B}}, z^{\mathbb{B}})$$

and apply ($\varepsilon_1$) and ($\varepsilon_2$) to remove the four quantifiers completely in an *equivalence transformation.*

We introduce the following abbreviations, where $w^{\mathbb{B}}$, $x^{\mathbb{B}}$, $y^{\mathbb{B}}$, $z^{\mathbb{B}}$ are bound atoms, $w_a$, $x_b$, $y_d$, $z_h$ are terms and $x_a$, $y_a$, $z_a$ are meta-level functions from terms to $\varepsilon$-terms:

$$\begin{array}{lcl}
z_a &=& \lambda w^{\mathbb{B}}, x^{\mathbb{B}}, y^{\mathbb{B}}.\ \varepsilon z^{\mathbb{B}}.\ \neg\mathsf{P}(w^{\mathbb{B}}, x^{\mathbb{B}}, y^{\mathbb{B}}, z^{\mathbb{B}}) \\
y_a &=& \lambda w^{\mathbb{B}}, x^{\mathbb{B}}.\ \varepsilon y^{\mathbb{B}}.\ \mathsf{P}(w^{\mathbb{B}}, x^{\mathbb{B}}, y^{\mathbb{B}}, z_a(w^{\mathbb{B}}, x^{\mathbb{B}}, y^{\mathbb{B}})) \\
x_a &=& \lambda w^{\mathbb{B}}.\ \varepsilon x^{\mathbb{B}}.\ \neg\mathsf{P}(w^{\mathbb{B}}, x^{\mathbb{B}}, y_a(w^{\mathbb{B}}, x^{\mathbb{B}}), z_a(w^{\mathbb{B}}, x^{\mathbb{B}}, y_a(w^{\mathbb{B}}, x^{\mathbb{B}}))), \\
w_a &=& \varepsilon w^{\mathbb{B}}.\ \mathsf{P}(w^{\mathbb{B}}, x_a(w^{\mathbb{B}}), y_a(w^{\mathbb{B}}, x_a(w^{\mathbb{B}})), z_a(w^{\mathbb{B}}, x_a(w^{\mathbb{B}}), y_a(w^{\mathbb{B}}, x_a(w^{\mathbb{B}})))), \\
x_b &=& x_a(w_a) \\
y_d &=& y_a(w_a, x_b) \\
z_h &=& z_a(w_a, x_b, y_d)
\end{array}$$

Innermost rewriting with ($\varepsilon_1$) and ($\varepsilon_2$) results in a unique normal form after at most as many steps as there are quantifiers (cf. [Wirth, 2016; 2017a]). Thus, we eliminate inside-out, i.e. we start with the elimination of $\forall z^{\mathbb{B}}$. The equivalence transformation is:

$$\begin{array}{ll}
\exists w^{\mathbb{B}}.\ \forall x^{\mathbb{B}}.\ \exists y^{\mathbb{B}}.\ \forall z^{\mathbb{B}}. & \mathsf{P}(w^{\mathbb{B}}, x^{\mathbb{B}}, y^{\mathbb{B}}, z^{\mathbb{B}}),\\
\exists w^{\mathbb{B}}.\ \forall x^{\mathbb{B}}.\ \exists y^{\mathbb{B}}. & \mathsf{P}(w^{\mathbb{B}}, x^{\mathbb{B}}, y^{\mathbb{B}}, z_a(w^{\mathbb{B}}, x^{\mathbb{B}}, y^{\mathbb{B}})),\\
\exists w^{\mathbb{B}}.\ \forall x^{\mathbb{B}}. & \mathsf{P}(w^{\mathbb{B}}, x^{\mathbb{B}}, y_a(w^{\mathbb{B}}, x^{\mathbb{B}}), z_a(w^{\mathbb{B}}, x^{\mathbb{B}}, y_a(w^{\mathbb{B}}, x^{\mathbb{B}}))),\\
\exists w^{\mathbb{B}}. & \mathsf{P}(w^{\mathbb{B}}, x_a(w^{\mathbb{B}}), y_a(w^{\mathbb{B}}, x_a(w^{\mathbb{B}})), z_a(w^{\mathbb{B}}, x_a(w^{\mathbb{B}}), y_a(w^{\mathbb{B}}, x_a(w^{\mathbb{B}})))),\\
& \mathsf{P}(w_a, x_a(w_a), y_a(w_a, x_a(w_a)), z_a(w_a, x_a(w_a), y_a(w_a, x_a(w_a)))),\\
& \mathsf{P}(w_a, x_b, y_d, z_h).
\end{array}$$

Note that the resulting formula $\mathsf{P}(w_a, x_b, y_d, z_h)$ is quite deep and has more than one thousand occurrences of the $\varepsilon$-binder. Indeed, in general, $n$ nested quantifiers result in an $\varepsilon$-nesting depth of $2^n-1$. How can this be? After all, we wrote down only four $\varepsilon$-terms in the three meta-level functions $z_a$, $y_a$, and $x_a$ and the term $w_a$, didn't we? Well, these functions occur in many different instances in the formula $\mathsf{P}(w_a, x_b, y_d, z_h)$, yielding different $\varepsilon$-terms, each expressing a different choice with its own commitment.

To understand this, let us have a closer look a the structure of the resulting formula

$$\mathsf{P}(w_a, x_b, y_d, z_h)$$

and display each of the top structures of its different $\varepsilon$-terms explicitly by means of a meta-level function from terms to $\varepsilon$-terms. Then we have:

$$\begin{array}{ll}
z_a = \lambda w^{\mathbb{B}}, x^{\mathbb{B}}, y^{\mathbb{B}}.\ \varepsilon z^{\mathbb{B}}.\ \neg\mathsf{P}(w^{\mathbb{B}}, x^{\mathbb{B}}, y^{\mathbb{B}}, z^{\mathbb{B}})\\
y_a = \lambda w^{\mathbb{B}}, x^{\mathbb{B}}.\ \varepsilon y^{\mathbb{B}}.\ \mathsf{P}(w^{\mathbb{B}}, x^{\mathbb{B}}, y^{\mathbb{B}}, z_a(w^{\mathbb{B}}, x^{\mathbb{B}}, y^{\mathbb{B}}))\\
z_b = \lambda w^{\mathbb{B}}.\ \lambda x^{\mathbb{B}}.\ \varepsilon z^{\mathbb{B}}.\ \neg\mathsf{P}(w^{\mathbb{B}}, x^{\mathbb{B}}, y_a(w^{\mathbb{B}}, x^{\mathbb{B}}), z^{\mathbb{B}})\\
x_a = \lambda w^{\mathbb{B}}.\ \varepsilon x^{\mathbb{B}}.\ \neg\mathsf{P}(w^{\mathbb{B}}, x^{\mathbb{B}}, y_a(w^{\mathbb{B}}, x^{\mathbb{B}}), z_b(w^{\mathbb{B}}, x^{\mathbb{B}}))\\
z_c = \lambda w^{\mathbb{B}}, y^{\mathbb{B}}.\ \varepsilon z^{\mathbb{B}}.\ \neg\mathsf{P}(w^{\mathbb{B}}, x_a(w^{\mathbb{B}}), y^{\mathbb{B}}, z^{\mathbb{B}}),\\
y_b = \lambda w^{\mathbb{B}}.\ \varepsilon y^{\mathbb{B}}.\ \mathsf{P}(w^{\mathbb{B}}, x_a(w^{\mathbb{B}}), y^{\mathbb{B}}, z_c(w^{\mathbb{B}}, y^{\mathbb{B}}))\\
z_d = \lambda w^{\mathbb{B}}.\ \varepsilon z^{\mathbb{B}}.\ \neg\mathsf{P}(w^{\mathbb{B}}, x_a(w^{\mathbb{B}}), y_b(w^{\mathbb{B}}), z^{\mathbb{B}})\\
w_a = \varepsilon w^{\mathbb{B}}.\ \mathsf{P}(w^{\mathbb{B}}, x_a(w^{\mathbb{B}}), y_b(w^{\mathbb{B}}), z_d(w^{\mathbb{B}}))
\end{array}$$

$$\begin{array}{l}
z_e = \lambda x^{\mathbb{B}}, y^{\mathbb{B}}.\ \varepsilon z^{\mathbb{B}}.\ \neg\mathsf{P}(w_a, x^{\mathbb{B}}, y^{\mathbb{B}}, z^{\mathbb{B}})\\
y_c = \lambda x^{\mathbb{B}}.\ \varepsilon y^{\mathbb{B}}.\ \mathsf{P}(w_a, x^{\mathbb{B}}, y^{\mathbb{B}}, z_e(x^{\mathbb{B}}, y^{\mathbb{B}}))\\
z_f = \lambda x^{\mathbb{B}}.\ \varepsilon z^{\mathbb{B}}.\ \neg\mathsf{P}(w_a, x^{\mathbb{B}}, y_c(x^{\mathbb{B}}), z^{\mathbb{B}})\\
x_b = \varepsilon x^{\mathbb{B}}.\ \neg\mathsf{P}(w_a, x^{\mathbb{B}}, y_c(x^{\mathbb{B}}), z_f(x^{\mathbb{B}}))\\
z_g = \lambda y^{\mathbb{B}}.\ \varepsilon z^{\mathbb{B}}.\ \neg\mathsf{P}(w_a, x_b, y^{\mathbb{B}}, z^{\mathbb{B}})\\
y_d = \varepsilon y^{\mathbb{B}}.\ \mathsf{P}(w_a, x_b, y^{\mathbb{B}}, z_g(y^{\mathbb{B}}))\\
z_h = \varepsilon z^{\mathbb{B}}.\ \neg\mathsf{P}(w_a, x_b, y_d, z^{\mathbb{B}})
\end{array}$$

First of all, note that the already defined symbols $z_a$, $y_a$, $x_a$, $w_a$, $x_b$, $y_d$, $z_h$ still denote the same terms as before, although the equations for $x_a$, $w_a$, $x_b$, $y_d$, $z_h$ differ from the previous notation by using more specific terms.

Moreover, the $\varepsilon$-terms described by the above equations are exactly those that require a commitment to their choice. This means that each of $z_a$, $z_b$, $z_c$, $z_d$, $z_e$, $z_f$, $z_g$, $z_h$, each of $y_a$, $y_b$, $y_c$, $y_d$, and each of $x_a$, $x_b$ may be chosen differently without affecting soundness of the equivalence transformation. Note that they are strictly nested into each other; so we must choose in the order of

$$z_a,\ y_a,\ z_b,\ x_a,\ z_c,\ y_b,\ z_d,\ w_a,\ z_e,\ y_c,\ z_f,\ x_b,\ z_g,\ y_d,\ z_h.$$

Furthermore, for all $\varepsilon$-terms except $w_a$, $x_b$, $y_d$, $z_h$, we actually have to choose a function instead of a simple object, which does not really matter on the actual term-level of the huge formula $\mathsf{P}(w_a, x_b, y_d, z_h)$, because there each of these functions is always applied to the same arguments. Indeed, each $z_i$ occurs only once in the directly following term, denoting an $\varepsilon$-term; and, in the same way, each $y_j$ only in the two following terms, applied to the same arguments; and each $x_k$ only in the four following terms applied to the same argument.

In Hilbert–Bernays' view, however, there are neither functions nor objects at all, but only terms (and expressions with free occurrences of bound atoms). For instance, in non-abbreviating notation the term $x_a(w^{\mathbb{B}})$ reads:

$$\varepsilon x^{\mathbb{B}}.\ \ \neg\mathsf{P}\left(\begin{array}{l} w^{\mathbb{B}},\\ x^{\mathbb{B}},\\ \varepsilon y^{\mathbb{B}}.\ \mathsf{P}\big(\ w^{\mathbb{B}},\ \ x^{\mathbb{B}},\ \ y^{\mathbb{B}},\ \ \varepsilon z^{\mathbb{B}}.\ \neg\mathsf{P}(w^{\mathbb{B}},x^{\mathbb{B}},y^{\mathbb{B}},z^{\mathbb{B}})\ \big),\\ \varepsilon z_b^{\mathbb{B}}.\ \neg\mathsf{P}\Big(\ w^{\mathbb{B}},\ \ x^{\mathbb{B}},\ \ \varepsilon y^{\mathbb{B}}.\ \mathsf{P}\big(\ w^{\mathbb{B}},\ \ x^{\mathbb{B}},\ \ y^{\mathbb{B}},\ \ \varepsilon z^{\mathbb{B}}.\ \neg\mathsf{P}(w^{\mathbb{B}},x^{\mathbb{B}},y^{\mathbb{B}},z^{\mathbb{B}})\ \big),\ \ z_b^{\mathbb{B}}\ \Big)\end{array}\right).$$

Note that we have renamed a bound atom $z^{\mathbb{B}}$ to $z_b^{\mathbb{B}}$, because we do not want to have a bound variable to be bound a second time in its scope. We chose $z_b^{\mathbb{B}}$ because this subterm corresponds to $z_b$. Moreover, $y_b(w^{\mathbb{B}})$ reads

$$\varepsilon y_b^{\mathbb{B}}.\ \mathsf{P}\left(\begin{array}{l} w^{\mathbb{B}},\\ \varepsilon x^{\mathbb{B}}.\ \ \neg\mathsf{P}\left(\begin{array}{l} w^{\mathbb{B}},\\ x^{\mathbb{B}},\\ \varepsilon y^{\mathbb{B}}.\ \mathsf{P}\big(\ w^{\mathbb{B}},\ \ x^{\mathbb{B}},\ \ y^{\mathbb{B}},\ \ \varepsilon z^{\mathbb{B}}.\ \neg\mathsf{P}(w^{\mathbb{B}},x^{\mathbb{B}},y^{\mathbb{B}},z^{\mathbb{B}})\ \big),\\ \varepsilon z_b^{\mathbb{B}}.\ \neg\mathsf{P}\Big(\ w^{\mathbb{B}},\ \ x^{\mathbb{B}},\ \ \varepsilon y^{\mathbb{B}}.\ \mathsf{P}\big(\ w^{\mathbb{B}},\ \ x^{\mathbb{B}},\ \ y^{\mathbb{B}},\ \ \varepsilon z^{\mathbb{B}}.\ \neg\mathsf{P}(w^{\mathbb{B}},x^{\mathbb{B}},y^{\mathbb{B}},z^{\mathbb{B}})\ \big),\ \ z_b^{\mathbb{B}}\ \Big)\end{array}\right),\\ y_b^{\mathbb{B}},\\ \varepsilon z_c^{\mathbb{B}}.\ \neg\mathsf{P}\left(\begin{array}{l} w^{\mathbb{B}},\\ \varepsilon x^{\mathbb{B}}.\ \ \neg\mathsf{P}\left(\begin{array}{l} w^{\mathbb{B}},\\ x^{\mathbb{B}},\\ \varepsilon y^{\mathbb{B}}.\ \mathsf{P}\big(\ w^{\mathbb{B}},\ \ x^{\mathbb{B}},\ \ y^{\mathbb{B}},\ \ \varepsilon z^{\mathbb{B}}.\ \neg\mathsf{P}(w^{\mathbb{B}},x^{\mathbb{B}},y^{\mathbb{B}},z^{\mathbb{B}})\ \big),\\ \varepsilon z_b^{\mathbb{B}}.\ \neg\mathsf{P}\Big(\ w^{\mathbb{B}},\ \ x^{\mathbb{B}},\ \ \varepsilon y^{\mathbb{B}}.\ \mathsf{P}\big(\ w^{\mathbb{B}},\ \ x^{\mathbb{B}},\ \ y^{\mathbb{B}},\ \ \varepsilon z^{\mathbb{B}}.\ \neg\mathsf{P}(w^{\mathbb{B}},x^{\mathbb{B}},y^{\mathbb{B}},z^{\mathbb{B}})\ \big),\ \ z_b^{\mathbb{B}}\ \Big)\end{array}\right),\\ y_b^{\mathbb{B}},\\ z_c^{\mathbb{B}}\end{array}\right)\end{array}\right).$$

Condensed data on the above terms read as follows:

| | $\varepsilon$-nesting depth | number of $\varepsilon$-binders | Ackermann rank | Ackermann degree |
|---|---|---|---|---|
| $z_a(w^{\mathbb{B}},x^{\mathbb{B}},y^{\mathbb{B}})$ | 1 | 1 | 1 | undefined |
| $y_a(w^{\mathbb{B}},x^{\mathbb{B}})$ | 2 | 2 | 2 | undefined |
| $z_b(w^{\mathbb{B}},x^{\mathbb{B}})$ | 3 | 3 | 1 | undefined |
| $x_a(w^{\mathbb{B}})$ | 4 | 6 | 3 | undefined |
| $z_c(w^{\mathbb{B}},y^{\mathbb{B}})$ | 5 | 7 | 1 | undefined |
| $y_b(w^{\mathbb{B}})$ | 6 | 14 | 2 | undefined |
| $z_d(w^{\mathbb{B}})$ | 7 | 21 | 1 | undefined |
| $w_a$ | 8 | 42 | 4 | 1 |
| $z_e(x^{\mathbb{B}},y^{\mathbb{B}})$ | 9 | 43 | 1 | undefined |
| $y_c(x^{\mathbb{B}})$ | 10 | 86 | 2 | undefined |
| $z_f(x^{\mathbb{B}})$ | 11 | 129 | 1 | undefined |
| $x_b$ | 12 | 258 | 3 | 2 |
| $z_g(y^{\mathbb{B}})$ | 13 | 301 | 1 | undefined |
| $y_d$ | 14 | 602 | 2 | 3 |
| $z_h$ | 15 | 903 | 1 | 4 |
| $\mathsf{P}(w_a,x_b,y_d,z_h)$ | 15 | 1805 | undefined | undefined |

For $\forall w^{\mathbb{B}}.\ \forall x^{\mathbb{B}}.\ \forall y^{\mathbb{B}}.\ \forall z^{\mathbb{B}}.\ \mathsf{P}(w^{\mathbb{B}},x^{\mathbb{B}},y^{\mathbb{B}},z^{\mathbb{B}})$ instead of $\exists w^{\mathbb{B}}.\ \forall x^{\mathbb{B}}.\ \exists y^{\mathbb{B}}.\ \forall z^{\mathbb{B}}.\ \mathsf{P}(w^{\mathbb{B}},x^{\mathbb{B}},y^{\mathbb{B}},z^{\mathbb{B}})$, we get the same exponential growth of nesting depth as in the example above, when we completely eliminate the quantifiers using ($\varepsilon_2$). The only difference is that we get additional occurrences of '$\neg$'. If we have quantifiers of the same kind in a row, however, we had better choose them in parallel; e.g., for $\forall w^{\mathbb{B}}.\ \forall x^{\mathbb{B}}.\ \forall y^{\mathbb{B}}.\ \forall z^{\mathbb{B}}.\ \mathsf{P}(w^{\mathbb{B}},x^{\mathbb{B}},y^{\mathbb{B}},z^{\mathbb{B}})$, we choose

$$v_a \quad := \quad \varepsilon v^{\mathbb{B}}.\ \neg\mathsf{P}(1^{\mathrm{st}}(v^{\mathbb{B}}), 2^{\mathrm{nd}}(v^{\mathbb{B}}), 3^{\mathrm{rd}}(v^{\mathbb{B}}), 4^{\mathrm{th}}(v^{\mathbb{B}})),$$

and then take $\mathsf{P}(1^{\mathrm{st}}(v_a), 2^{\mathrm{nd}}(v_a), 3^{\mathrm{rd}}(v_a), 4^{\mathrm{th}}(v_a))$ as result of the elimination.

Roughly speaking, in today's automated theorem proving, the exponential explosion of term depth of the example above is avoided by an outside-in removal of $\delta$-quantifiers *without removing the quantifiers below $\varepsilon$-binders* and by a replacement of $\gamma$-quantified bound atoms with variables without choice-conditions. For the formula of Example 4.6, this yields $\mathsf{P}(w^{\mathbb{V}}, \overline{x}, y^{\mathbb{V}}, \overline{z})$ with $\overline{x} = \varepsilon x^{\mathbb{B}}.\ \neg\exists y^{\mathbb{B}}.\ \forall z^{\mathbb{B}}.\ \mathsf{P}(w^{\mathbb{V}}, x^{\mathbb{B}}, y^{\mathbb{B}}, z^{\mathbb{B}})$ and $\overline{z} = \varepsilon z^{\mathbb{B}}.\ \neg\mathsf{P}(w^{\mathbb{V}}, \overline{x}, y^{\mathbb{V}}, z^{\mathbb{B}})$. Thus, in general, the nesting of binders for the complete elimination of a prenex of $n$ quantifiers does not become deeper than $\frac{1}{4}(n{+}1)^2$.

Moreover, if we are only interested in reduction and not in equivalence transformation, we can get rid of the $\varepsilon$-terms by reducing to one of the formulas $\mathsf{P}(w^{\mathbb{V}}, x^{\mathbb{A}}(w^{\mathbb{V}}), y^{\mathbb{V}}, z^{\mathbb{A}}(w^{\mathbb{V}}, y^{\mathbb{V}}))$ with SKOLEM terms or $\mathsf{P}(w^{\mathbb{V}}, x^{\mathbb{A}}, y^{\mathbb{V}}, z^{\mathbb{A}})$ with $\{(w^{\mathbb{V}}, x^{\mathbb{A}}),\ (w^{\mathbb{V}}, z^{\mathbb{A}}),\ (y^{\mathbb{V}}, z^{\mathbb{A}})\}$ as an extension to the variable-condition. Note that neither with SKOLEMization nor variable-conditions we have any non-constant growth of nesting depth, and the same will be the case for our approach to $\varepsilon$-terms.

## 4.8 Our Objective

While the historiographical and technical research on the $\varepsilon$-theorems is still going on and the methods of $\varepsilon$-elimination and $\varepsilon$-substitution did not die with HILBERT's program, this is not our subject here. We are less interested in HILBERT's formal program and the consistency of mathematics than in the powerful use of logic in creative processes. And, instead of the tedious syntactic proof transformations, which easily lose their usefulness and elegance within their technical complexity and which — more importantly — can only refer to an already existing logic, we look for *model-theoretic* means for finding new logics and new applications. And the question that still has to be answered in this field is:

*What would be a proper semantics for Hilbert's $\varepsilon$ ?*

## 4.9 Indefinite and Committed Choice

Just as the $\iota$-symbol is usually taken to be the referential interpretation of the *definite* articles in natural languages, it is our opinion that the $\varepsilon$-symbol should be that of the *indefinite* determiners (articles and pronouns) such as "a(n)" or "some".

**Example 4.7 ($\varepsilon$ instead of $\iota$ again)** *(continuing Example 4.1)*
It may well be the case that

$$\mathsf{Holy\,Ghost} \;=\; \varepsilon x^{\mathbb{B}}.\ \mathsf{Father}(x^{\mathbb{B}}, \mathsf{Jesus}) \quad\wedge\quad \mathsf{Joseph} \;=\; \varepsilon x^{\mathbb{B}}.\ \mathsf{Father}(x^{\mathbb{B}}, \mathsf{Jesus})$$

i.e. that "The Holy Ghost is <u>*a*</u> father of Jesus and Joseph is <u>*a*</u> father of Jesus." But this does not bring us into trouble with the Pope because we do not know whether all fathers of Jesus are equal. This will become clearer when we reconsider this in Example 4.15.

Closely connected to indefinite choice (also called "indeterminism" or "don't care nondeterminism") is the notion of *committed choice.* For example, when we have a new telephone, we typically *don't care* which number we get, but once a number has been chosen for our telephone, we will insist on a *commitment to this choice*, so that our phone number is not changed between two incoming calls.

**Example 4.8 (Committed choice)**
Suppose we want to prove $\exists x^{\mathbb{B}}.\ (x^{\mathbb{B}} \neq x^{\mathbb{B}})$
According to ($\varepsilon_1$) from § 4.6 this reduces to $\varepsilon x^{\mathbb{B}}.\ (x^{\mathbb{B}} \neq x^{\mathbb{B}}) \neq \varepsilon x^{\mathbb{B}}.\ (x^{\mathbb{B}} \neq x^{\mathbb{B}})$
Since there is no solution to $x^{\mathbb{B}} \neq x^{\mathbb{B}}$ we can replace
$\varepsilon x^{\mathbb{B}}.\ (x^{\mathbb{B}} \neq x^{\mathbb{B}})$ with anything. Thus, the above reduces to $\mathsf{0} \neq \varepsilon x^{\mathbb{B}}.\ (x^{\mathbb{B}} \neq x^{\mathbb{B}})$
and then, by exactly the same argumentation, to $\mathsf{0} \neq \mathsf{1}$
which is true in the natural numbers.

Thus, we would have proved our original formula $\exists x^{\mathbb{B}}.\ (x^{\mathbb{B}} \neq x^{\mathbb{B}})$, which, however, is false. What went wrong? Well, we have to commit to our choice for all occurrences of the $\varepsilon$-term introduced when eliminating the existential quantifier: If we choose $\mathsf{0}$ on the left-hand side, we have to commit to the choice of $\mathsf{0}$ on the right-hand side as well.

## 4.10 Do not be afraid of Indefiniteness!

From the discussion in § 4.9, one could get the impression that an indefinite logical treatment of the $\varepsilon$ is not easy to find. Indeed, at first sight, there is the problem that some standard axiom schemes cannot be taken for granted, such as substitutability

$$s = t \qquad \Rightarrow \qquad f(s) = f(t)$$

and reflexivity

$$t = t$$

Note that substitutability is similar to the highly controversial *extensionality axiom*

$$\forall x^{\mathbb{B}}.\ (A_0 \Leftrightarrow A_1) \qquad \Rightarrow \qquad \varepsilon x^{\mathbb{B}}.\ A_0 \ = \ \varepsilon x^{\mathbb{B}}.\ A_1 \tag{E2}$$

(cf. § 8.4) when we take logical equivalence as equality. Moreover, note that

$$\varepsilon x^{\mathbb{B}}.\ \mathsf{true} \ = \ \varepsilon x^{\mathbb{B}}.\ \mathsf{true} \tag{Reflex}$$

is an instance of reflexivity.

Thus, it seems that — in case of an indefinite $\varepsilon$ — the replacement of a subterm with an equal term is problematic, and so is the equality of syntactically equal terms.

It may be interesting to see that — in computer programs — we are quite used to committed choice and to an indefinite behavior of choosing, and that the violation of substitutability and even reflexivity is no problem there:

**Example 4.9 (Violation of Substitutability and Reflexivity in Programs)**
In the implementation of the specification of the web-based hypertext system of [Mattick & Wirth, 1999], we needed a function that chooses an element from a set implemented as a list. Its ML code is:

```
fun choose s = case s of Set (i :: _) => i | _ => raise Empty;
```

And, of course, it simply returns the first element of the list. For another set that is equal — but where the list may have another order — the result may be different. Thus, the behavior of the function `choose` is indefinite for a given set, but any time it is called for an implemented set, it chooses a particular element and *commits to this choice,* i.e.: when called again, it returns the same value. In this case we have `choose s` = `choose s`, but `s` = `t` does not imply `choose s` = `choose t`. In an implementation where some parallel reordering of lists may take place, even `choose s` = `choose s` may be wrong.

From this example we may learn that the question of a commitment of `choose s` is not settled until its choice step has actually been implemented. *This is also the case with our* $\varepsilon$*.* Indeed, in our treatment of an $\varepsilon$-term, the choice step may be implemented by a representation with commitment of choice or by one without.

Thus, on the one hand, when we want to prove

$$\varepsilon x^{\mathbb{B}}.\,\mathsf{true} \;=\; \varepsilon x^{\mathbb{B}}.\,\mathsf{true}$$

we can choose $\mathsf{0}$ for both occurrences of $\varepsilon x^{\mathbb{B}}.\,\mathsf{true}$, get $\mathsf{0}=\mathsf{0}$, and the proof is successful.

On the other hand, suppose that we want to prove

$$\varepsilon x^{\mathbb{B}}.\,\mathsf{true} \;\neq\; \varepsilon x^{\mathbb{B}}.\,\mathsf{true}$$

*without commitment*, or, equivalently,

$$\varepsilon x^{\mathbb{B}}.\,\mathsf{true} \;\neq\; \varepsilon y^{\mathbb{B}}.\,\mathsf{true},$$

because we consider equality of terms and formulas only up to renaming of bound atoms. Then we can choose $\mathsf{0}$ for one occurrence and $\mathsf{1}$ for the other get $\mathsf{0}\neq\mathsf{1}$, and the proof is successful again.

This procedure seems wondrous, but it is very similar to something altogether commonplace for the case of variables instead of $\varepsilon$-terms:

On the one hand, when we want to prove

$$x^{\mathbb{V}} = y^{\mathbb{V}}$$

we can choose $\mathsf{0}$ to replace both $x^{\mathbb{V}}$ and $y^{\mathbb{V}}$, get $\mathsf{0}=\mathsf{0}$, and the proof is successful.

On the other hand, when we want to prove

$$x^{\mathbb{V}} \neq y^{\mathbb{V}}$$

we can choose $\mathsf{0}$ to replace $x^{\mathbb{V}}$ and $\mathsf{1}$ to replace $y^{\mathbb{V}}$, get $\mathsf{0}\neq\mathsf{1}$, and the proof is successful again.

There is an important difference, however, between the inequations $\varepsilon x^{\mathbb{B}}.\,\mathsf{true} \neq \varepsilon x^{\mathbb{B}}.\,\mathsf{true}$ and $x^{\mathbb{V}} \neq y^{\mathbb{V}}$: The latter does not violate the reflexivity axiom!

And we are going to cure the violation of the former immediately with the help of our variables, but now with *non-empty* choice-conditions. Instead of $\varepsilon x^{\mathbb{B}}.\,\mathsf{true} \neq \varepsilon x^{\mathbb{B}}.\,\mathsf{true}$ we write $x^{\mathbb{V}} \neq y^{\mathbb{V}}$ and remember for what these variables stand, by storing this information into a function $C$, called a *choice-condition*:

$$\begin{array}{rcl} C(x^{\mathbb{V}}) & := & \varepsilon x^{\mathbb{B}}.\,\mathsf{true}, \\ C(y^{\mathbb{V}}) & := & \varepsilon x^{\mathbb{B}}.\,\mathsf{true}. \end{array}$$

In the following, we will now describe how to do this in general.

## 4.11 Replacing $\varepsilon$-Terms with Variables

For a first step, suppose that our $\varepsilon$-terms are not subordinate to any outside binder (cf. Definition 4.5). Then, we can replace any $\varepsilon$-term $\varepsilon z^{\mathbb{B}}.\,A$ with a fresh variable $z^{\mathbb{V}}$ and extend the partial function $C$ by

$$C(z^{\mathbb{V}}) \;:=\; \varepsilon z^{\mathbb{B}}.\,A.$$

Note that this does not change the syntax of the $\varepsilon$-term at all, which is to be appreciated in particular if this term contains further $\varepsilon$-terms to be replaced in such steps later on.

The overall procedure eliminates all $\varepsilon$-terms in sequents, formulas and terms (except on top of the values of choice-conditions) without loosing any information, but yielding the following two advantages:

- A crucial advantage of replacing $\varepsilon$-terms with fresh variables is that these variables will clearly indicate whether a committed choice is required: We can express a **committed choice** by repeatedly using the same variable, and a **choice without commitment** by several variables with the same choice-condition.
- As another consequence of this elimination, the substitutability and reflexivity axioms are immediately regained, and the problems discussed in § 4.10 disappear.

**Example 4.10**
This also solves our problems with committed choice of Example 4.8 of § 4.9:
Now, again using $(\varepsilon_1)$, $\exists x^{\mathbb{B}}.\,(x^{\mathbb{B}} \neq x^{\mathbb{B}})$ reduces to $x^{\mathbb{V}} \neq x^{\mathbb{V}}$ with

$$C(x^{\mathbb{V}}) \;:=\; \varepsilon x^{\mathbb{B}}.\,(x^{\mathbb{B}} \neq x^{\mathbb{B}})$$

and the proof attempt immediately fails because of the now regained reflexivity axiom.

As the second step, we still have to explain what to do with subordinate $\varepsilon$-terms. If the $\varepsilon$-term $\varepsilon v_l^{\mathbb{B}}.\,A$ contains free occurrences of exactly the distinct bound atoms $v_0^{\mathbb{B}}, \ldots, v_{l-1}^{\mathbb{B}}$, then we have to replace this $\varepsilon$-term with the application term $z^{\mathbb{V}}(v_0^{\mathbb{B}}, \ldots, v_{l-1}^{\mathbb{B}})$ (of the same type as $v_l^{\mathbb{B}}$) (for a fresh variable $z^{\mathbb{V}}$) and to extend the choice-condition $C$ by

$$C(z^{\mathbb{V}}) \;:=\; \lambda v_0^{\mathbb{B}}, \ldots, v_{l-1}^{\mathbb{B}}.\,\varepsilon v_l^{\mathbb{B}}.\,A.$$

Notice that — even after subsequent renaming — the functional variable $z^{\mathbb{V}}$ will occur in our terms and formulas and in the ranges of our choice-conditions only followed by a tuple providing all its $l$ arguments, i.e. in the form $z^{\mathbb{V}}(t_0, \ldots, t_{l-1})$, but never without such an $l$-tuple, say in an equation like $z^{\mathbb{V}} = x^{\mathbb{V}}$ or $z^{\mathbb{V}} = \lambda v_0^{\mathbb{B}}, \ldots, v_{l-1}^{\mathbb{B}}.\,t$. Thus, we do not introduce higher-order language by our elimination of $\varepsilon$-terms, but use $\lambda$-notation only at top level in the values $C(z^{\mathbb{V}})$ of our choice-conditions — to bind the free occurrences of the bound atoms in the following $\varepsilon$-term.

Moreover, notice that no renaming of atoms or variables takes place in the formula $A$ and only fresh variables are introduced, so that the result of the complete replacement of all $\varepsilon$-terms does not depend on any strategy, such as innermost or outermost.

**Example 4.11 (Choice-Condition)** *(continuing Example 4.6 of § 4.7)*
Let us now do the elimination of quantifiers by innermost rewriting with $(\varepsilon_1)$ and $(\varepsilon_2)$ from the beginning of Example 4.6 again, but now eliminate each $\varepsilon$-term immediately by replacing it with a fresh variable with proper choice-condition. Then the equivalence transformation looks hardly any different: The transformation reads

$$\begin{array}{ll}
\exists w^{\mathbb{B}}.\ \forall x^{\mathbb{B}}.\ \exists y^{\mathbb{B}}.\ \forall z^{\mathbb{B}}. & \mathsf{P}(w^{\mathbb{B}}, x^{\mathbb{B}}, y^{\mathbb{B}}, z^{\mathbb{B}}),\\
\exists w^{\mathbb{B}}.\ \forall x^{\mathbb{B}}.\ \exists y^{\mathbb{B}}. & \mathsf{P}(w^{\mathbb{B}}, x^{\mathbb{B}}, y^{\mathbb{B}}, z^{\mathbb{V}}_a(w^{\mathbb{B}}, x^{\mathbb{B}}, y^{\mathbb{B}})),\\
\exists w^{\mathbb{B}}.\ \forall x^{\mathbb{B}}. & \mathsf{P}(w^{\mathbb{B}}, x^{\mathbb{B}}, y^{\mathbb{V}}_a(w^{\mathbb{B}}, x^{\mathbb{B}}), z^{\mathbb{V}}_a(w^{\mathbb{B}}, x^{\mathbb{B}}, y^{\mathbb{V}}_a(w^{\mathbb{B}}, x^{\mathbb{B}}))),\\
\exists w^{\mathbb{B}}. & \mathsf{P}(w^{\mathbb{B}}, x^{\mathbb{V}}_a(w^{\mathbb{B}}), y^{\mathbb{V}}_a(w^{\mathbb{B}}, x^{\mathbb{V}}_a(w^{\mathbb{B}})), z^{\mathbb{V}}_a(w^{\mathbb{B}}, x^{\mathbb{V}}_a(w^{\mathbb{B}}), y^{\mathbb{V}}_a(w^{\mathbb{B}}, x^{\mathbb{V}}_a(w^{\mathbb{B}})))),\\
 & \mathsf{P}(w^{\mathbb{V}}_a, x^{\mathbb{V}}_a(w^{\mathbb{V}}_a), y^{\mathbb{V}}_a(w^{\mathbb{V}}_a, x^{\mathbb{V}}_a(w^{\mathbb{V}}_a)), z^{\mathbb{V}}_a(w^{\mathbb{V}}_a, x^{\mathbb{V}}_a(w^{\mathbb{V}}_a), y^{\mathbb{V}}_a(w^{\mathbb{V}}_a, x^{\mathbb{V}}_a(w^{\mathbb{V}}_a)))),
\end{array}$$

and comes with the following choice-condition $C$:

$$\begin{array}{lcl}
C(z^{\mathbb{V}}_a) & := & \lambda w^{\mathbb{B}}, x^{\mathbb{B}}, y^{\mathbb{B}}.\ \varepsilon z^{\mathbb{B}}.\ \neg\mathsf{P}(w^{\mathbb{B}}, x^{\mathbb{B}}, y^{\mathbb{B}}, z^{\mathbb{B}})\\
C(y^{\mathbb{V}}_a) & := & \lambda w^{\mathbb{B}}, x^{\mathbb{B}}.\ \varepsilon y^{\mathbb{B}}.\ \mathsf{P}(w^{\mathbb{B}}, x^{\mathbb{B}}, y^{\mathbb{B}}, z^{\mathbb{V}}_a(w^{\mathbb{B}}, x^{\mathbb{B}}, y^{\mathbb{B}}))\\
C(x^{\mathbb{V}}_a) & := & \lambda w^{\mathbb{B}}.\ \varepsilon x^{\mathbb{B}}.\ \neg\mathsf{P}(w^{\mathbb{B}}, x^{\mathbb{B}}, y^{\mathbb{V}}_a(w^{\mathbb{B}}, x^{\mathbb{B}}), z^{\mathbb{V}}_a(w^{\mathbb{B}}, x^{\mathbb{B}}, y^{\mathbb{V}}_a(w^{\mathbb{B}}, x^{\mathbb{B}})))\\
C(w^{\mathbb{V}}_a) & := & \varepsilon w^{\mathbb{B}}.\ \mathsf{P}(w^{\mathbb{B}}, x^{\mathbb{V}}_a(w^{\mathbb{B}}), y^{\mathbb{V}}_a(w^{\mathbb{B}}, x^{\mathbb{V}}_a(w^{\mathbb{B}})), z^{\mathbb{V}}_a(w^{\mathbb{B}}, x^{\mathbb{V}}_a(w^{\mathbb{B}}), y^{\mathbb{V}}_a(w^{\mathbb{B}}, x^{\mathbb{V}}_a(w^{\mathbb{B}}))))
\end{array}$$

If we again want to have an explicit representation of each and every $\varepsilon$-term that comes with commitment to its choice, then this again looks hardly any different from Example 4.6: The result is

$$\mathsf{P}(w^{\mathbb{V}}_a, x^{\mathbb{V}}_b, y^{\mathbb{V}}_d, z^{\mathbb{V}}_h),$$

and it comes with the following alternative choice-condition $C'$:

$$\begin{array}{lcrl}
C'(z^{\mathbb{V}}_a) & := & \lambda w^{\mathbb{B}}, x^{\mathbb{B}}, y^{\mathbb{B}}.\ \varepsilon z^{\mathbb{B}}. & \neg\mathsf{P}(w^{\mathbb{B}}, x^{\mathbb{B}}, y^{\mathbb{B}}, z^{\mathbb{B}})\\
C'(y^{\mathbb{V}}_a) & := & \lambda w^{\mathbb{B}}, x^{\mathbb{B}}.\ \varepsilon y^{\mathbb{B}}. & \mathsf{P}(w^{\mathbb{B}}, x^{\mathbb{B}}, y^{\mathbb{B}}, z^{\mathbb{V}}_a(w^{\mathbb{B}}, x^{\mathbb{B}}, y^{\mathbb{B}}))\\
C'(z^{\mathbb{V}}_b) & := & \lambda w^{\mathbb{B}}, x^{\mathbb{B}}.\ \varepsilon z^{\mathbb{B}}. & \neg\mathsf{P}(w^{\mathbb{B}}, x^{\mathbb{B}}, y^{\mathbb{V}}_a(w^{\mathbb{B}}, x^{\mathbb{B}}), z^{\mathbb{B}})\\
C'(x^{\mathbb{V}}_a) & := & \lambda w^{\mathbb{B}}.\ \varepsilon x^{\mathbb{B}}. & \neg\mathsf{P}(w^{\mathbb{B}}, x^{\mathbb{B}}, y^{\mathbb{V}}_a(w^{\mathbb{B}}, x^{\mathbb{B}}), z^{\mathbb{V}}_b(w^{\mathbb{B}}, x^{\mathbb{B}}))\\
C'(z^{\mathbb{V}}_c) & := & \lambda w^{\mathbb{B}}, y^{\mathbb{B}}.\ \varepsilon z^{\mathbb{B}}. & \neg\mathsf{P}(w^{\mathbb{B}}, x^{\mathbb{V}}_a(w^{\mathbb{B}}), y^{\mathbb{B}}, z^{\mathbb{B}})\\
C'(y^{\mathbb{V}}_b) & := & \lambda w^{\mathbb{B}}.\ \varepsilon y^{\mathbb{B}}. & \mathsf{P}(w^{\mathbb{B}}, x^{\mathbb{V}}_a(w^{\mathbb{B}}), y^{\mathbb{B}}, z^{\mathbb{V}}_c(w^{\mathbb{B}}, y^{\mathbb{B}}))\\
C'(z^{\mathbb{V}}_d) & := & \lambda w^{\mathbb{B}}.\ \varepsilon z^{\mathbb{B}}. & \neg\mathsf{P}(w^{\mathbb{B}}, x^{\mathbb{V}}_a(w^{\mathbb{B}}), y^{\mathbb{V}}_b(w^{\mathbb{B}}), z^{\mathbb{B}})\\
C'(w^{\mathbb{V}}_a) & := & \varepsilon w^{\mathbb{B}}. & \mathsf{P}(w^{\mathbb{B}}, x^{\mathbb{V}}_a(w^{\mathbb{B}}), y^{\mathbb{V}}_b(w^{\mathbb{B}}), z^{\mathbb{V}}_d(w^{\mathbb{B}}))\\
C'(z^{\mathbb{V}}_e) & := & \lambda x^{\mathbb{B}}, y^{\mathbb{B}}.\ \varepsilon z^{\mathbb{B}}. & \neg\mathsf{P}(w^{\mathbb{V}}_a, x^{\mathbb{B}}, y^{\mathbb{B}}, z^{\mathbb{B}})\\
C'(y^{\mathbb{V}}_c) & := & \lambda x^{\mathbb{B}}.\ \varepsilon y^{\mathbb{B}}. & \mathsf{P}(w^{\mathbb{V}}_a, x^{\mathbb{B}}, y^{\mathbb{B}}, z^{\mathbb{V}}_e(x^{\mathbb{B}}, y^{\mathbb{B}}))\\
C'(z^{\mathbb{V}}_f) & := & \lambda x^{\mathbb{B}}.\ \varepsilon z^{\mathbb{B}}. & \neg\mathsf{P}(w^{\mathbb{V}}_a, x^{\mathbb{B}}, y^{\mathbb{V}}_c(x^{\mathbb{B}}), z^{\mathbb{B}})\\
C'(x^{\mathbb{V}}_b) & := & \varepsilon x^{\mathbb{B}}. & \neg\mathsf{P}(w^{\mathbb{V}}_a, x^{\mathbb{B}}, y^{\mathbb{V}}_c(x^{\mathbb{B}}), z^{\mathbb{V}}_f(x^{\mathbb{B}}))\\
C'(z^{\mathbb{V}}_g) & := & \lambda y^{\mathbb{B}}.\ \varepsilon z^{\mathbb{B}}. & \neg\mathsf{P}(w^{\mathbb{V}}_a, x^{\mathbb{V}}_b, y^{\mathbb{B}}, z^{\mathbb{B}})\\
C'(y^{\mathbb{V}}_d) & := & \varepsilon y^{\mathbb{B}}. & \mathsf{P}(w^{\mathbb{V}}_a, x^{\mathbb{V}}_b, y^{\mathbb{B}}, z^{\mathbb{V}}_g(y^{\mathbb{B}}))\\
C(z^{\mathbb{V}}_h) & := & \varepsilon z^{\mathbb{B}}. & \neg\mathsf{P}(w^{\mathbb{V}}_a, x^{\mathbb{V}}_b, y^{\mathbb{V}}_d, z^{\mathbb{B}})
\end{array}$$

Both representations here are much smaller and easier to understand than any of those in Example 4.6. Indeed, the $\lambda$-binders are not anymore a notation for huge meta-level terms, but just part of our syntax for choice-conditions. Moreover, the formula $\mathsf{P}(w^{\mathbb{V}}_a, x^{\mathbb{V}}_b, y^{\mathbb{V}}_d, z^{\mathbb{V}}_h)$ has a term depth of 1 and no $\varepsilon$-binders, whereas $\mathsf{P}(w_a, x_b, y_d, z_h)$ in Example 4.6 had already a mere $\varepsilon$-depth of 15 and 1805 $\varepsilon$-binders.

## 4.12 Why We Do Not Abandon the $\varepsilon$-Symbol

By our just described rewriting procedure, we can replace explicit representations of $\varepsilon$-terms in formulas and sequents completely with variables. The only places where the $\varepsilon$ still occurs is outermost under at most one $\lambda$-operator in the range of the choice-condition $C$; and also there it is not essential because, instead of

$$C(z^{\mathbb{V}}) \;=\; \lambda v_0^{\mathbb{B}},\ldots,v_{l-1}^{\mathbb{B}}.\ \varepsilon v_l^{\mathbb{B}}.\ A,$$

we could write

$$C(z^{\mathbb{V}}) \;=\; \lambda v_0^{\mathbb{B}},\ldots,v_{l-1}^{\mathbb{B}}.\ A\{v_l^{\mathbb{B}} \mapsto z^{\mathbb{V}}(v_0^{\mathbb{B}},\ldots,v_{l-1}^{\mathbb{B}})\}$$

as we have actually done previously in [Wirth, 2004; 2006a; 2008; 2012b; 2006b].

The old procedure, however, turned out to be a bit clumsy in several aspects, complicating definitions and procedures. For instance, it requires to eliminate innermost $\varepsilon$-terms first, because otherwise the bound atoms $v_0^{\mathbb{B}}$, …, $v_{l-1}^{\mathbb{B}}$ could be newly introduced to the $\varepsilon$-terms subordinate to $\varepsilon v_l^{\mathbb{B}}.\ A$; and the variable for those subordinate $\varepsilon$-terms would then have to take more $\lambda$-arguments than needed, which may make later proofs infeasible. Although innermost rewriting would still yield the same result, there are too many other little, but superfluous complications with our previous presentation of our choice-conditions. Last but not least — although we neither presuppose any $\lambda$- nor any $\varepsilon$-calculus — our way of writing the values of our choice-conditions with $\lambda$ and $\varepsilon$ immediately conveys the actual intuition behind our choice-conditions to any logician.

## 4.13 Crucial Representational Change, but nothing more yet?

After all, Example 4.11 shows that — by combination of choice-conditions and term sharing via variables — in our framework, Hilbert's $\varepsilon$ becomes practically feasible for the first time.

Up to here, however, one might still consider the effect of our free-variable framework on Hilbert's $\varepsilon$ to be a merely representational one.

Indeed, as long as a fresh variable with choice-condition appears only once in a formula, the original tree structure of formulas is kept in our framework up to isomorphism. Computer scientists will see this fresh variable just as a pointer to a central storage named "choice-condition" of all objects of data type "Hilbert's $\varepsilon$", where they can find the sub-tree for this variable's $\varepsilon$-term with its sub-formula. And a user-friendly interface could still be configured to present only the traditional $\varepsilon$-terms to a die-hard conservative user.

As we can glimpse from the last formula of the initial transformation of Example 4.11, however, quantifier-elimination does not stay with single occurrence of any fresh variable. Thus, the effect of committed choice by multiple occurrences of variables does not only sneak in from modern application areas of the $\varepsilon$, but already from the most traditional of its applications! And then we have to admit that the tree structure of formulas has actually turned into a more efficient, well-know data structure called a *DAG* (*directed acyclic graph*).

In the following sections we will also go beyond the mostly representational character of our treatment of Hilbert's $\varepsilon$, which already now comes with the traditionally not available option of $\varepsilon$-terms without committed choice (via two different variables with identical values under the choice-condition) and a nice stepwise rewriting procedure without any strategic restrictions by which we can flexibly and even partially generate our new representation.

## 4.14 Instantiating Choice-Conditioned Variables (“$\varepsilon$-Substitution”)

Having already realized Requirement I (Indication of Commitment) of § 4.2 in § 4.11, we are now going to explain how to satisfy Requirement II (Reasoning). To this end, we have to explain how to replace variables with terms that satisfy their choice-conditions.

The first thing to know about variables with choice-conditions is: Just like the variables without choice-conditions (introduced by $\gamma$-rules e.g.) and contrary to free atoms, the variables with choice-conditions (introduced by $\delta^{+}$-rules e.g.) are *rigid* in the sense that the only way to replace a variable is to do it *globally*, i.e. in all formulas and all choice-conditions with the same term in an atomic transaction.

In *reductive* theorem proving, such as in sequent, tableau, matrix, or indexed-formula-tree calculi, we are in the following situation: While a variable without choice-condition can be replaced with everything (satisfying the current variable-condition), the replacement of a variable with a choice-condition requires some proof work, and a free atom cannot be instantiated at all.

Contrariwise, when formulas are used as tools instead of tasks, free atoms can indeed be replaced — and this even locally (i.e. non-rigidly) and repeatedly. This is the case not only for purely *generative* calculi (such as resolution and paramodulation calculi) and HILBERT-style calculi (such as the predicate calculus of [HILBERT & BERNAYS, 1934; 1939; 1968; 1970]), but also for the lemma and induction hypothesis application in the otherwise reductive calculi of [WIRTH, 2004], cf. [WIRTH, 2004, § 2.5.2].

More precisely — again considering *reductive* theorem proving, where formulas are proof tasks — a variable without choice-condition may be instantiated with any term (of appropriate type) that does not violate the current variable-condition, cf. § 5.5 for formal details. The instantiation of a variable with choice-condition additionally requires some proof work depending on the current choice-condition, cf. Definition 5.12 for formal details. In general, if a substitution $\sigma$ replaces the variable $y^{\mathbb{V}}$ in the domain of the choice-condition $C$, then — to know that the global instantiation of the entire proof attempt with $\sigma$ satisfies the choice-condition $C$ — we have to prove $(Q_C(y^{\mathbb{V}}))\sigma$, where $Q_C$ is given as follows:

**Definition 4.12** ($Q_C, Q'_C$)
$Q_C$ or else $Q'_C$ are the functions that map every $z^{\mathbb{V}} \in \mathrm{dom}(C)$ with

$$C(z^{\mathbb{V}}) = \lambda v_0^{\mathbb{B}}, \ldots, v_{l-1}^{\mathbb{B}}.\ \varepsilon v_l^{\mathbb{B}}.\ B$$

for some types $\alpha_0, \ldots, \alpha_l$, some mutually distinct bound atoms $v_0^{\mathbb{B}}, \ldots, v_l^{\mathbb{B}} \in \mathbb{B}$, and some formula $B$ with $z^{\mathbb{V}} : \alpha_0, \ldots, \alpha_{l-1} \to \alpha_l$, $v_0^{\mathbb{B}} : \alpha_0$, $\ldots$, $v_l^{\mathbb{B}} : \alpha_l$, and $\mathbb{B}(B) \subseteq \{v_0^{\mathbb{B}}, \ldots, v_l^{\mathbb{B}}\}$ (otherwise $Q_C$ and $Q'_C$ are undefined) to the single-formula sequent

$$\forall v_0^{\mathbb{B}}.\ \ldots \forall v_{l-1}^{\mathbb{B}}.\ \big(\ \exists v_l^{\mathbb{B}}.\ B \quad \Rightarrow \quad B\{v_l^{\mathbb{B}} \mapsto z^{\mathbb{V}}(v_0^{\mathbb{B}}, \ldots, v_{l-1}^{\mathbb{B}})\}\ \big),$$

or else to the two-formula sequent

$$\neg B\mu\{v_l^{\mathbb{B}} \mapsto v_l^{\mathbb{A}}\} \qquad B\mu\{v_l^{\mathbb{B}} \mapsto z^{\mathbb{V}}(v_0^{\mathbb{A}}, \ldots, v_{l-1}^{\mathbb{A}})\},$$

for the substitution $\mu = \{v_0^{\mathbb{B}} \mapsto v_0^{\mathbb{A}}, \ldots, v_{l-1}^{\mathbb{B}} \mapsto v_{l-1}^{\mathbb{A}}\}$ and for mutually different fresh free atoms $v_0^{\mathbb{A}}, \ldots, v_l^{\mathbb{A}} \in \mathbb{A}$ with $v_0^{\mathbb{A}} : \alpha_0$, $\ldots$, $v_l^{\mathbb{A}} : \alpha_l$.

First of all, note that $Q_C(z^{\mathbb{V}})$ is nothing but a formulation of HILBERT–BERNAYS’ axiom ($\varepsilon_0$) (cf. § 4.5) in our framework. Moreover, Lemma 7.10 will state the validity of $Q_C(z^{\mathbb{V}})$. This means that the satisfaction of the $\varepsilon$’s specification depends only on the substitution $\sigma$.

Indeed, regarding the $\sigma$-instance of $Q_C(z^{\mathrm{V}})$ whose provability is required, it is only the *arbitrariness* of the substitution $\sigma$ that poses the question of satisfying the choice-condition.

Furthermore, note that $\mathbb{B}(\sigma) = \emptyset = \mathbb{A}(\sigma, B) \cap \{v_0^{\mathbb{A}}, \ldots, v_l^{\mathbb{A}}\}$ and that, by $l{-}1$ $\delta^-$-steps, one $\alpha$-step, and then one more $\delta^-$-step, the sequent $Q_C(z^{\mathrm{V}})$ reduces to the sequent $Q'_C(z^{\mathrm{V}})$ as well as the sequent $(Q_C(z^{\mathrm{V}}))\sigma$ to the sequent $(Q'_C(z^{\mathrm{V}}))\sigma$ with $\mathbb{V}((Q'_C(z^{\mathrm{V}}))\sigma) \times \{v_0^{\mathbb{A}}, \ldots, v_l^{\mathbb{A}}\}$ as the extension to the negative part $N$ of our positive/negative variable-condition $(P, N)$. We will often have to introduce $(Q_C(z^{\mathrm{V}}))\sigma$ as a new lemma to our proof forest and then we prefer its more technical version $(Q'_C(z^{\mathrm{V}}))\sigma$ because this has the first $l{+}1$ steps of its proof already done and — more important — its many free atoms can all be arbitrarily instantiated whenever we will apply the lemma elsewhere, cf. Theorem 8.3(3) and Example 4.14.

Now, as an example for $Q_C$, we can replay Example 3.1 in § 3.5 and use it for a discussion of the $\delta^+$-rule instead of the $\delta^-$-rule:

**Example 4.13 (Soundness of $\delta^+$-rule)** The formula $\exists y^{\mathbb{B}}.\ \forall x^{\mathbb{B}}.\ (y^{\mathbb{B}} = x^{\mathbb{B}})$ is valid in structures with only one single object, but not in general. We can again start a reductive proof attempt of it, but now with a $\delta^+$-step instead of the $\delta^-$-step in Example 3.1:

| | |
|---|---|
| $\gamma$-step: | $\forall x^{\mathbb{B}}.\ (y^{\mathrm{V}} = x^{\mathbb{B}}),\quad \exists y^{\mathbb{B}}.\ \forall x^{\mathbb{B}}.\ (y^{\mathbb{B}} = x^{\mathbb{B}})$ |
| $\delta^+$-step: | $(y^{\mathrm{V}} = x^{\mathrm{V}}),\quad \exists y^{\mathbb{B}}.\ \forall x^{\mathbb{B}}.\ (y^{\mathbb{B}} = x^{\mathbb{B}})$ |

Now, if the variable $y^{\mathrm{V}}$ could be replaced with the variable $x^{\mathrm{V}}$, then we would get the tautology $(x^{\mathrm{V}} = x^{\mathrm{V}})$, i.e. we would have proved an invalid formula. To prevent this, as indicated to the lower right of the bar of the first of the $\delta^+$-rules in § 3.6 on Page 14, the $\delta^+$-step has to record
$$\mathbb{VA}(\forall x^{\mathbb{B}}.\ (y^{\mathrm{V}} = x^{\mathbb{B}})) \times \{x^{\mathrm{V}}\} \quad = \quad \{(y^{\mathrm{V}}, x^{\mathrm{V}})\}$$
in a positive variable-condition, where $(y^{\mathrm{V}}, x^{\mathrm{V}})$ means that "$x^{\mathrm{V}}$ positively depends on $y^{\mathrm{V}}$" (or that "$y^{\mathrm{V}}$ is a subterm of the description of $x^{\mathrm{V}}$"), so that we may never instantiate the variable $y^{\mathrm{V}}$ with a term containing the variable $x^{\mathrm{V}}$, because this instantiation would result in cyclic dependencies (or in a cyclic term).

Contrary to Example 3.1, we have a further opportunity here to complete this proof attempt into a successful proof: If the substitution $\sigma := \{x^{\mathrm{V}} \mapsto y^{\mathrm{V}}\}$ could be applied, then we would get the tautology $(y^{\mathrm{V}} = y^{\mathrm{V}})$, i.e. we would have proved an invalid formula. To prevent this — as indicated to the upper right of the bar of the first of the $\delta^+$-rules in § 3.6 on Page 14 — the $\delta^+$-step has to record
$$\left(\ x^{\mathrm{V}},\ \ \varepsilon x^{\mathbb{B}}.\ \neg(y^{\mathrm{V}} = x^{\mathbb{B}})\ \right)$$
in the choice-condition $C$. If we take this pair as an equation, then the intuition behind the above statement that $y^{\mathrm{V}}$ is somehow a subterm of the description of $x^{\mathrm{V}}$ becomes immediately clear. If we take it as an element of the graph of the function $C$, however, then we can compute $(Q_C(x^{\mathrm{V}}))\sigma$ and try to prove it. $Q_C(x^{\mathrm{V}})$ is $\exists x^{\mathbb{B}}.\ \neg(y^{\mathrm{V}} = x^{\mathbb{B}}) \ \Rightarrow\ \neg(y^{\mathrm{V}} = x^{\mathrm{V}})$;
so $(Q_C(x^{\mathrm{V}}))\sigma$ is $\exists x^{\mathbb{B}}.\ \neg(y^{\mathrm{V}} = x^{\mathbb{B}}) \ \Rightarrow\ \neg(y^{\mathrm{V}} = y^{\mathrm{V}})$.
In classical logic with equality this is equivalent to $\exists x^{\mathbb{B}}.\ \neg(y^{\mathrm{V}} = x^{\mathbb{B}}) \ \Rightarrow\ \mathsf{false}$, and then to $\forall x^{\mathbb{B}}.\ (y^{\mathrm{V}} = x^{\mathbb{B}})$. If we were able to show the truth of this formula, then it would be sound to apply the substitution $\sigma$ to prove the above sequent resulting from the $\gamma$-step. That sequent, however, already lists this formula as an element of its disjunction. Thus, no progress is possible by means of the $\delta^+$-rules here; and so this example is not a counterexample to the soundness of the $\delta^+$-rules.

**Example 4.14 (Predecessor Function)**

Suppose that our domain is natural numbers and that $y^{\mathbb{V}}$ has the choice-condition

$$C(y^{\mathbb{V}}) \quad = \quad \lambda v_0^{\mathbb{B}}.\ \varepsilon v_1^{\mathbb{B}}.\ \big(v_0^{\mathbb{B}} = v_1^{\mathbb{B}} + 1\big).$$

Then the single-formula sequent $Q_C(y^{\mathbb{V}})$ reads

$$\forall v_0^{\mathbb{B}}.\ \Big(\exists v_1^{\mathbb{B}}.\ \big(v_0^{\mathbb{B}} = v_1^{\mathbb{B}} + 1\big) \Rightarrow \big(v_0^{\mathbb{B}} = y^{\mathbb{V}}(v_0^{\mathbb{B}}) + 1\big)\Big).$$

Then, before we may instantiate $y^{\mathbb{V}}$ with the symbol $\mathsf{p}$ for the predecessor function, partially specified by the single defining equation

$$\mathsf{p}(x^{\mathbb{A}}{+}1) \ = \ x^{\mathbb{A}},$$

we have to prove the single-formula sequent $(Q_C(y^{\mathbb{V}}))\{y^{\mathbb{V}}{\mapsto}\mathsf{p}\}$, which reads

$$\forall v_0^{\mathbb{B}}.\ \Big(\exists v_1^{\mathbb{B}}.\ \big(v_0^{\mathbb{B}} = v_1^{\mathbb{B}} + 1\big) \Rightarrow \big(v_0^{\mathbb{B}} = \mathsf{p}(v_0^{\mathbb{B}}) + 1\big)\Big).$$

Let us move out the existential quantifier one step. Then, by contextual rewriting of $\mathsf{p}(v_0^{\mathbb{B}})$, first with $v_0^{\mathbb{B}} = v_1^{\mathbb{B}} + 1$ and then with the defining equation for $\mathsf{p}$, which functions here as a lemma and thus admits the instantiation via $\{x^{\mathbb{A}} \mapsto v_1^{\mathbb{B}}\}$, we can reduce this proof task to

$$\forall v_0^{\mathbb{B}}.\ \exists v_1^{\mathbb{B}}.\ \Big(\big(v_0^{\mathbb{B}} = v_1^{\mathbb{B}} + 1\big) \Rightarrow \big(v_0^{\mathbb{B}} = v_1^{\mathbb{B}} + 1\big)\Big),$$

which reduces to the tautology $A \Rightarrow A$. This completes the proof and we now may globally apply the substitution $\{y^{\mathbb{V}}{\mapsto}\mathsf{p}\}$, because our variable-condition is not changed by this substitution because its value is only a constant without any variables or free atoms.

For a comparison and as an exercise, let us do this proof again, but start — instead of the single-formula sequent $(Q_C(y^{\mathbb{V}}))\{y^{\mathbb{V}}{\mapsto}\mathsf{p}\}$ — directly with its reduct, the two formula sequent $(Q'_C(y^{\mathbb{V}}))\{y^{\mathbb{V}}{\mapsto}\mathsf{p}\}$. Here $(Q'_C(y^{\mathbb{V}}))\{y^{\mathbb{V}}{\mapsto}\mathsf{p}\}$ reads

$$\neg\big(v_0^{\mathbb{B}} = v_1^{\mathbb{B}} + 1\big)\{v_0^{\mathbb{B}}{\mapsto}v_0^{\mathbb{A}}\}\{v_1^{\mathbb{B}}{\mapsto}v_1^{\mathbb{A}}\}\{y^{\mathbb{V}}{\mapsto}\mathsf{p}\} \qquad \big(v_0^{\mathbb{B}} = v_1^{\mathbb{B}} + 1\big)\{v_0^{\mathbb{B}}{\mapsto}v_0^{\mathbb{A}}\}\{v_1^{\mathbb{B}}{\mapsto}y^{\mathbb{V}}(v_0^{\mathbb{A}})\}\{y^{\mathbb{V}}{\mapsto}\mathsf{p}\},$$

i.e.

$$\neg\big(v_0^{\mathbb{A}} = v_1^{\mathbb{A}} + 1\big) \qquad \big(v_0^{\mathbb{A}} = \mathsf{p}(v_0^{\mathbb{A}}) + 1\big).$$

By our primitive contextual-rewriting rule of § 3.2, we can rewrite the second formula first with the first one and then with the defining equation for $\mathsf{p}$, and obtain the tautology

$$\neg\big(v_0^{\mathbb{A}} = v_1^{\mathbb{A}} + 1\big) \qquad \big(v_0^{\mathbb{A}} = v_1^{\mathbb{A}} + 1\big).$$

Thus, we have seen that proving $(Q'_C(y^{\mathbb{V}}))\{y^{\mathbb{V}}{\mapsto}\mathsf{p}\}$ formally is indeed a bit simpler than proving $(Q_C(y^{\mathbb{V}}))\{y^{\mathbb{V}}{\mapsto}\mathsf{p}\}$ and that — with its two free atoms — the former is useful when activated as a rewrite lemma, whereas the latter is hardly of any use as a lemma.

Finally, note that the fact that $\mathsf{p}(0)$ is not specified here is no problem at all, simply because the value of $\mathsf{p}(0)$ is not required in this proof. In general, it is not necessary to specify it in any equation, as the total function $\mathsf{p}$ may be left just as underspecified as our $\varepsilon$. For an elaborate framework of partial specification with positive/negative-conditional equations, see [Wirth, 2009].

Before we end § 4, our introduction to HILBERT's $\varepsilon$, let us jump back to the first example of it, namely Example 4.1 in § 4.4.2.

This first example, as you will have noticed, referred to the omnipresent legend on what happened in Canossa in the year 1077, after the Pope had anathematized the German king Heinrich IV, which was a highly debated subject among leading historians for nearly thousand years, until [FRIED, 2012] clearly settled the possibilities and their likelihoods in a spectacular way.

Instead of our first example, which dealt with the $\iota$ yet, we will now finally reconsider the $\varepsilon$-version of this example, namely Example 4.7 in § 4.9.

**Example 4.15 (Canossa 1077)** *(continuing Example 4.1 (§ 4.4.2) and Example 4.7 (§ 4.9)*

After complete elimination of the $\varepsilon$-terms in the formula displayed in Example 4.7 by our rewriting procedure, this formula reads:

$$\mathsf{Holy\,Ghost} \;=\; z_0^{\mathbb{V}} \quad\wedge\quad \mathsf{Joseph} \;=\; z_1^{\mathbb{V}} \tag{4.15.1}$$

with $\quad C(z_0^{\mathbb{V}}) = \varepsilon x^{\mathbb{B}}.\ \mathsf{Father}(x^{\mathbb{B}}, \mathsf{Jesus}),$

and $\quad C(z_1^{\mathbb{V}}) = \varepsilon x^{\mathbb{B}}.\ \mathsf{Father}(x^{\mathbb{B}}, \mathsf{Jesus}).$

This does not bring us into the old trouble with the Pope because nobody knows whether $z_0^{\mathbb{V}} = z_1^{\mathbb{V}}$ holds or not. Indeed, the identical value of $z_0^{\mathbb{V}}$ and $z_1^{\mathbb{V}}$ under the choice-condition $C$ does not at all imply any commitment of choice here! When the identical $\varepsilon$-term occurred in the traditional framework of HILBERT–BERNAYS, however, it always expressed a committed choice for all occurrences. This was necessary to get the quantifier elimination by means of the $\varepsilon$ going (cf. Example 4.8), but it was not a good style: Even the conservative traditionalists never agreed on whether the commitment remains active after renaming of bound atoms inside the $\varepsilon$-term. We would definitely say that renaming of atoms must not do any harm to a given commitment.

On the one hand, knowing (4.1.2) from Example 4.1 of § 4.4.2, we can even prove (4.15.1) as follows: Let us replace $z_0^{\mathbb{V}}$ with Holy Ghost because, for $\sigma_0 := \{z_0^{\mathbb{V}} \mapsto \mathsf{Holy\,Ghost}\}$, from Father(Holy Ghost, Jesus) we conclude

$$\exists x^{\mathbb{B}}.\ \mathsf{Father}(x^{\mathbb{B}}, \mathsf{Jesus}) \quad\Rightarrow\quad \mathsf{Father}(\mathsf{Holy\,Ghost}, \mathsf{Jesus}),$$

which is nothing but the required $(Q_C(z_0^{\mathbb{V}}))\sigma_0$.

Analogously, we replace $z_1^{\mathbb{V}}$ with Joseph because, for $\sigma_1 := \{z_1^{\mathbb{V}} \mapsto \mathsf{Joseph}\}$, from (4.1.2) we conclude the required $(Q_C(z_1^{\mathbb{V}}))\sigma_1$. After these replacements, (4.15.1) becomes the tautology

$$\mathsf{Holy\,Ghost} \;=\; \mathsf{Holy\,Ghost} \quad\wedge\quad \mathsf{Joseph} \;=\; \mathsf{Joseph}$$

On the other hand, if we want to have trouble, we can apply the substitution

$$\sigma' \;=\; \{z_0^{\mathbb{V}} \mapsto \mathsf{Joseph},\ z_1^{\mathbb{V}} \mapsto \mathsf{Joseph}\}$$

to (4.15.1) because both $(Q_C(z_0^{\mathbb{V}}))\sigma'$ and $(Q_C(z_1^{\mathbb{V}}))\sigma'$ are equal to $(Q_C(z_1^{\mathbb{V}}))\sigma_1$. Then our task is to show

$$\mathsf{Holy\,Ghost} \;=\; \mathsf{Joseph} \quad\wedge\quad \mathsf{Joseph} \;=\; \mathsf{Joseph}.$$

Note that this course of action is stupid, even under the aspect of theorem proving alone.

# 5 Formal Presentation of Our Syntax

After some preliminary subsections, we formalize our novel positive/negative variable-conditions and their consistency (§ 5.4), extensions, $\sigma$-updates, and admissibility of substitutions (§ 5.5). Moreover, we formalize or choice-conditions (§ 5.6) and their extensions and $\sigma$-updates (§ 5.7). All in all, we formalize all our required syntactic ingredients here.

## 5.1 Basic Notions and Notation

'$\mathbf{N}$' denotes the set of natural numbers and '$\prec$' the ordering on $\mathbf{N}$. Let $\mathbf{N}_+ := \{\, n\in\mathbf{N} \mid 0\neq n \,\}$. We use '$\uplus$' for the union of disjoint classes and 'id' for the identity function. For classes $R$, $A$, and $B$ we define:

| | | |
|---|---|---|
| $\mathrm{dom}(R)$ | $:= \{\, a \mid \exists b.\ (a,b)\in R \,\}$ | *domain* |
| ${}_A{\upharpoonleft}R$ | $:= \{\, (a,b)\in R \mid a\in A \,\}$ | *(domain-) restriction to* $A$ |
| $\langle A\rangle R$ | $:= \{\, b \mid \exists a\in A.\ (a,b)\in R \,\}$ | *image of* $A$, i.e. $\langle A\rangle R = \mathrm{ran}({}_A{\upharpoonleft}R)$ |

And the dual ones:

| | | |
|---|---|---|
| $\mathrm{ran}(R)$ | $:= \{\, b \mid \exists a.\ (a,b)\in R \,\}$ | *range* |
| $R{\upharpoonright}_B$ | $:= \{\, (a,b)\in R \mid b\in B \,\}$ | *range-restriction to* $B$ |
| $R\langle B\rangle$ | $:= \{\, a \mid \exists b\in B.\ (a,b)\in R \,\}$ | *reverse-image of* $B$, i.e. $R\langle B\rangle = \mathrm{dom}(R{\upharpoonright}_B)$ |

Furthermore, we use '$\emptyset$' to denote the empty set as well as the empty function. Functions are (right-) unique relations, and so the meaning of "$f\circ g$" is extensionally given by $(f\circ g)(x) = g(f(x))$. The *class of total functions from* $A$ *to* $B$ is denoted as $A \to B$. The *class of (possibly) partial functions from* $A$ *to* $B$ is denoted as $A \leadsto B$. Both $\to$ and $\leadsto$ associate to the right, i.e. $A \leadsto B \to C$ reads $A \leadsto (B \to C)$.

Let $R$ be a binary relation. $R$ is said to be a relation *on* $A$ if $\mathrm{dom}(R) \cup \mathrm{ran}(R) \subseteq A$. $R$ is *irreflexive* if $\mathrm{id}\cap R = \emptyset$. It is $A$*-reflexive* if ${}_A{\upharpoonleft}\mathrm{id} \subseteq R$. Speaking of a *reflexive* relation we refer to the largest $A$ that is appropriate in the local context, and referring to this $A$ we write $R^0$ ambiguously to denote ${}_A{\upharpoonleft}\mathrm{id}$. With $R^1 := R$, and $R^{n+1} := R^n\circ R$ for $n \in \mathbf{N}_+$, $R^m$ denotes the $m$-step relation for $R$. The *transitive closure* of $R$ is $R^+ := \bigcup_{n\in\mathbf{N}_+} R^n$. The *reflexive transitive closure* of $R$ is $R^* := \bigcup_{n\in\mathbf{N}} R^n$. The *reverse* of $R$ is $R^{-1} := \{\, (b,a) \mid (a,b)\in R \,\}$. A relation $R$ [on $A$] is *well-founded* if every non-empty class $B$ [$\subseteq A$] has an $R$-minimal element, i.e. $\exists a\in B.\ \neg\exists a'\in B.\ a' R\, a$. A sequence $(s_i)_{i\in\mathbf{N}}$ is *non-terminating in* $R$ if $s_i\ R\ s_{i+1}$ for all $i \in \mathbf{N}$. $R$ is *terminating* if there are no non-terminating sequences in $R$.

A *quasi-ordering* '$\lesssim$' on a class $A$ is an $A$-reflexive and transitive (binary) relation on $A$, and we define its *reverse* by $\gtrsim\ :=\ \lesssim^{-1}$ And its *equivalence* by $\approx\ :=\ \lesssim\cap\gtrsim$. By an *ordering* '$<$' we mean an irreflexive and transitive relation, sometimes called "strict partial ordering" by others. A *reflexive ordering '$\leq$' on* $A$ is an $A$-reflexive, antisymmetric, and transitive relation on $A$, sometimes called "partial ordering" by others. The *ordering* $<$ *of a quasi-ordering or a reflexive ordering* $\lesssim$ is $\lesssim\setminus\gtrsim$, and $\lesssim$ is called *well-founded* if $<$ is well-founded.

**Lemma 5.1** *(Lemma 2.1 in [*Wirth*, 2004, p. 17])*
*For a binary relation $R$ we have the following equivalences:*
*$R$ is well-founded iff $R^+$ is well-founded iff $R^+$ is a well-founded ordering.*

## 5.2 Choice Functions

To be more useful in the context of HILBERT's $\varepsilon$, the standard notion of a "choice function" $f$ needs to be slightly modified to admit $\emptyset \in \mathrm{dom}(f)$ in spite of $f(\emptyset) \notin \emptyset$ :

**Definition 5.2 ([Generalized / Function-] Choice Function)**
$f$ is a *choice function* [*on A*] if $f$ is a function with $[A \subseteq \mathrm{dom}(f)$ and$]$
$$f : \mathrm{dom}(f) \to \bigcup(\mathrm{dom}(f)) \text{ and } \forall Y \in \mathrm{dom}(f).\ (\ f(Y) \in Y\ ).$$
$f$ is a *generalized choice function* [*on A*] if $f$ is a function with $[A \subseteq \mathrm{dom}(f)$ and$]$
$$f : \mathrm{dom}(f) \to \bigcup(\mathrm{dom}(f)) \text{ and } \forall Y \in \mathrm{dom}(f).\ (\ f(Y) \in Y \ \vee\ Y = \emptyset\ ).$$
$f$ is a *function-choice function for a function F* if $f$ is a function with $\mathrm{dom}(F) \subseteq \mathrm{dom}(f)$ and $\forall x \in \mathrm{dom}(F).\ (\ f(x) \in F(x)\ )$.

**Corollary 5.3**
*The empty function $\emptyset$ is both a choice function and a generalized choice function on $\emptyset$.*
*If $\mathrm{dom}(f) = \{\emptyset\}$, then $f$ is neither a choice function nor a generalized choice function.*
*If $\emptyset \notin \mathrm{dom}(f)$, then $f$ is a generalized choice function if and only if $f$ is a choice function.*
*If $\emptyset \in \mathrm{dom}(f)$, then $f$ is a generalized choice function if and only if there is a choice function $f'$ and an $x \in \bigcup(\mathrm{dom}(f'))$ such that $f = f' \uplus \{(\emptyset, x)\}$.*

## 5.3 Variables, Atoms, Constants, and Substitutions

We assume the following sets of symbols to be disjoint:

- $\mathbb{V}$ (free) (rigid) *variables*, which serve as unknowns or the free variables of [FITTING, 1990; 1996]
- $\mathbb{A}$ *free atoms*, which serve as parameters and must not be bound
- $\mathbb{B}$ *bound atoms*, which may be bound
- $\Sigma$ *constants*, i.e. the function and predicate symbols from the signature

We define:

$\mathbb{VA} := \mathbb{V} \uplus \mathbb{A}$
$\mathbb{VAB} := \mathbb{V} \uplus \mathbb{A} \uplus \mathbb{B}$

By slight abuse of notation, for $S \in \{\mathbb{V}, \mathbb{A}, \mathbb{B}, \mathbb{VA}, \mathbb{VAB}\}$, we write "$S(\Gamma)$" to denote the set of symbols from $S$ that have *free* occurrences in $\Gamma$.

Let $\sigma$ be a substitution. $\sigma$ is a *substitution on V* if $\mathrm{dom}(\sigma) \subseteq V$. Unless explicitly stated otherwise, we use only substitutions on subsets of $\mathbb{VAB}$.

The following crucial statement (as simple as it is) will require some discussion.

**The Substitution Statement:**

> We denote with "$\Gamma\sigma$" the result of replacing each free occurrence of a symbol $x \in \mathrm{dom}(\sigma)$ in $\Gamma$ with $\sigma(x)$; possibly after renaming in $\Gamma$ some symbols that are bound in $\Gamma$, in particular because a capture of their free occurrences in $\sigma(x)$ must be avoided.

We bind only symbols from the set $\mathbb{B}$ of bound atoms. And — unless explicitly stated otherwise — we tacitly assume that all occurrences of atoms from $\mathbb{B}$ in a term or formula or in the range of a substitution are *bound occurrences*, i.e. that an atom $x^{\mathbb{B}} \in \mathbb{B}$ occurs in these contexts only in the scope of a binder on $x^{\mathbb{B}}$.

Therefore, a renaming of symbols bound in $\Gamma$, as mentioned in the substitution statement, is hardly required in this paper because, in standard situations, even without renaming, no additional occurrences can become bound (i.e. captured) when applying a substitution; with the exception of non-outermost substitutions, inside the context of a term or formula. For instance, if we implement $\lambda$-reduction in such a context via reduction of $(\lambda x^{\mathbb{B}}.\ s)(t)$ to $s\{x^{\mathbb{B}}\mapsto t\}$, then, for each atom $y^{\mathbb{B}} \in \mathbb{B}(t)$ (i.e. for each bound atom $y^{\mathbb{B}}$ with *free* occurrences in $t$), each of the binders of $y^{\mathbb{B}}$ in $s$ — with a free occurrence of $x^{\mathbb{B}}$ within the scope of this binder — has to be renamed to a fresh bound atom before we apply the substitution $\{x^{\mathbb{B}}\mapsto t\}$.

This situation is avoided in Hilbert–Bernays by requiring (1) exactly the occurrences of bound atoms to be bound (as we do as well) and by (2) forbidding binders of a bound atom in the scope of a binder on the same atom (what we do not forbid). Then indeed, if $y^{\mathbb{B}}$ occurs free in $t$, it must have a binder in the context of $(\lambda x^{\mathbb{B}}.\ s)(t)$, and thus cannot have a binder in $s$ according to requirement (2). Even with Hilbert–Bernays' requirements (1, 2), however, we cannot get rid of renaming of atoms bound in $\Gamma$ completely: Let $\Gamma$ be the formula $\forall x^{\mathbb{B}}.(x^{\mathbb{B}} = y^{\mathbb{V}})$ and $\sigma$ be $\{y^{\mathbb{V}} \mapsto \varepsilon x^{\mathbb{B}}.(x^{\mathbb{B}} = x^{\mathbb{B}})\}$. To maintain Hilbert–Bernays' requirements, $\Gamma\sigma$ must not be $\forall x^{\mathbb{B}}.(x^{\mathbb{B}} = \varepsilon x^{\mathbb{B}}.(x^{\mathbb{B}} = x^{\mathbb{B}}))$, although we accept such a formula in this paper. To satisfy Hilbert–Bernays' requirements, however, $x^{\mathbb{B}}$ must be renamed in $\Gamma$ before substitution, say to $z^{\mathbb{B}}$, so that $\Gamma\sigma$ becomes $\forall z^{\mathbb{B}}.(z^{\mathbb{B}} = \varepsilon x^{\mathbb{B}}.(x^{\mathbb{B}} = x^{\mathbb{B}}))$, which we prefer as well because it is easier to read.

## 5.4 Consistent Positive/Negative Variable-Conditions

Variable-conditions are binary relations on variables and free atoms. They put conditions on the possible substitutions on variables, and on the dependencies of their valuations. To gain clearer expression and higher expressiveness, in this paper, a variable-condition is formalized as a pair $(P, N)$ of binary relations, called a "positive/negative variable-condition":

- $P$, the first component of such a pair, is a binary relation that is meant to express *positive* dependencies. It comes with the intention of transitivity, although it will typically not be closed up to transitivity for reasons of presentation and efficiency. The overall idea is that the occurrence of a pair $(x^{\mathbb{VA}}, y^{\mathbb{V}})$ in this positive relation means something like
  "the admissibility of a value for $y^{\mathbb{V}}$ may depend on $x^{\mathbb{VA}}$"
  or
  "$x^{\mathbb{VA}}$ is considered to be part of the specification of the values admissible for $y^{\mathbb{V}}$".
- $N$, the second component, however, is meant to capture *negative* dependencies. The overall idea is that the occurrence of a pair $(x^{\mathbb{V}}, y^{\mathbb{A}})$ in this negative relation means something like
  "the value of $x^{\mathbb{V}}$ must not depend on $y^{\mathbb{A}}$"
  or
  "$y^{\mathbb{A}}$ is fresh for $x^{\mathbb{V}}$".
  or
  "the variable $x^{\mathbb{V}}$ must not be substituted with a term in which $x^{\mathbb{A}}$ could ever appear — not even after subsequent substitutions on its variables".

Relations similar to this negative relation already occurred as the only component of a variable-condition in [Wirth, 1998], and later — with a completely different motivation — as "*freshness* conditions" also in [Gabbay & Pitts, 2002].

**Definition 5.4 (Positive/Negative Variable-Condition)**
A *positive/negative variable-condition* is a pair $(P, N)$ with

$$P \subseteq \mathbb{VA} \times \mathbb{V}$$

and

$$N \subseteq \mathbb{V} \times \mathbb{A}.$$

In a positive/negative variable-condition $(P, N)$, the relations $P$ and $N$ are always disjoint because their ranges are subsets of the disjoint sets $\mathbb{V}$ and $\mathbb{A}$, respectively. Moreover, note that in this paper, the only changes on the set $N$ come from applications of the $\delta^-$-rules (introducing elements of $\mathbb{V}\times\mathbb{A}$), whereas the only changes on the set $P$ come from applications of the $\delta^+$-rules and from the global instantiations of variables (both introducing elements from $\mathbb{VA}\times\mathbb{V}$).

A relation exactly like this positive relation $P$ was the only component of a variable-condition as defined and used identically throughout [Wirth, 2002; 2004; 2006a; 2008; 2012b; 2006b]. Note, however, that, in these publications, we had to admit this single positive relation to be a subset of $\mathbb{VA}\times\mathbb{VA}$ (instead of the restriction to $\mathbb{VA}\times\mathbb{V}$ of Definition 5.4 in this paper), because it had to simulate the negative relation ($N$) in addition; thereby losing some expressive power as compared to our positive/negative variable-conditions here (cf. Example 6.1). For further considerations on the design of our special form of variable-conditions here, see § C.

In the following definition, the well-foundedness guarantees that all dependencies can be traced back to independent symbols and that no variable may transitively depend on itself, whereas the irreflexivity makes sure that no contradictious dependencies can occur.

**Definition 5.5 (Consistency)**
A pair $(P, N)$ is *consistent* if $P$ is well-founded and $P^+ \circ N$ is irreflexive.

Let $(P, N)$ be a positive/negative variable-condition. Let us think of our (binary) relations $P$ and $N$ as edges of a directed graph whose vertices are the symbols for variables and free atoms currently in use. Then, $P^+ \circ N$ is irreflexive if and only if there is no cycle in $P \cup N$ that contains exactly one edge from $N$. Moreover, in practice, a positive/negative variable-condition $(P, N)$ can always be chosen to be finite in both its components. In the case that $P$ is finite, $P$ is well-founded if and only if $P$ is acyclic. Thus we get:

**Corollary 5.6**
*Let $(P, N)$ be a positive/negative variable-condition with $|P| \in \mathbf{N}$.*
*$(P, N)$ is consistent if and only if*
*each cycle in the directed graph of $P \uplus N$ contains more than one edge from $N$.*
*In case of $|N| \in \mathbf{N}$, the right-hand side of this equivalence can be effectively tested with an asymptotic time complexity of $|P| + |N|$.*

Note that, in the finite case, the test of Corollary 5.6 seems to be both the most efficient and the most human-oriented way to represent the question of consistency of positive/negative variable-conditions.

## 5.5 Extensions, $\sigma$-Updates, and $(P, N)$-Substitutions

Within a progressing reasoning process, positive/negative variable-conditions may be subject to only one kind of transformation, which we simply call an "extension".

**Definition 5.7 ([Weak] Extension)**
$(P', N')$ is an [*weak*] *extension of* $(P, N)$ if $(P', N')$ is a positive/negative variable-condition, $P \subseteq P'$ [or at least $P \subseteq (P')^{+}$], and $N \subseteq N'$.

As immediate corollaries of Definitions 5.7 and 5.5 and Lemma 5.1 we get:

**Corollary 5.8**
*Being an extension is a reflexive ordering.*
*Being a weak extension is a quasi-ordering, and its equivalence is given by identity of both the negative relation and the transitive closure of the positive relation.*

**Corollary 5.9** *If $(P', N')$ is a consistent positive/negative variable-condition and an [weak] extension of $(P, N)$, then $(P, N)$ is a consistent positive/negative variable-condition as well.*

A $\sigma$-update is a special form of an extension:

**Definition 5.10 ($\sigma$-Update, Dependence Relation)**
Let $(P, N)$ be a positive/negative variable-condition and $\sigma$ be a substitution on $\mathbb{V}$.
The *dependence relation of* $\sigma$ is $D_\sigma :=$

$$\left\{ \ (z^{\mathbb{VA}}, x^{\mathbb{V}}) \ \middle| \ x^{\mathbb{V}} \in \mathrm{dom}(\sigma) \ \wedge \ z^{\mathbb{VA}} \in \mathbb{VA}(\sigma(x^{\mathbb{V}})) \ \right\}.$$

The *$\sigma$-update of* $(P, N)$ is $\left( \ P \cup D_\sigma, \ \ N \ \right)$.[3]

**Definition 5.11 ($(P, N)$-Substitution)**
Let $(P, N)$ be a positive/negative variable-condition. $\sigma$ is a $(P, N)$*-substitution* if
$\sigma$ is a substitution on $\mathbb{V}$ and the $\sigma$-update of $(P, N)$ is consistent.

Syntactically, $(x^{\mathbb{V}}, a^{\mathbb{A}}) \in N$ is to express that a $(P, N)$-substitution $\sigma$ must not replace $x^{\mathbb{V}}$ with a term in which $a^{\mathbb{A}}$ could ever occur; i.e. that $a^{\mathbb{A}}$ is fresh for $x^{\mathbb{V}}$: $a^{\mathbb{A}} \mathrel{\#} x^{\mathbb{V}}$. This is indeed guaranteed if any $\sigma$-update $(P', N')$ of $(P, N)$ is again required to be consistent, and so on. We can see this as follows: For $z^{\mathbb{V}} \in \mathbb{V}(\sigma(x^{\mathbb{V}}))$, we get

$$z^{\mathbb{V}} \ \ P' \ \ x^{\mathbb{V}} \ \ N' \ \ a^{\mathbb{A}}.$$

If we now try to apply a second substitution $\sigma'$ with $a^{\mathbb{A}} \in \mathbb{A}(\sigma'(z^{\mathbb{V}}))$ (so that $a^{\mathbb{A}}$ occurs in $(x^{\mathbb{V}}\sigma)\sigma'$, contrary to what we initially expressed as our freshness intention), then $\sigma'$ is not a $(P', N')$-substitution because, for the $\sigma'$-update $(P'', N'')$ of $(P', N')$, we have

$$a^{\mathbb{A}} \ \ P'' \ \ z^{\mathbb{V}} \ \ P'' \ \ x^{\mathbb{V}} \ \ N'' \ \ a^{\mathbb{A}};$$

so $(P'')^{+} \circ N''$ is not irreflexive. All in all, the positive/negative variable-condition

- $(P', N')$ blocks any instantiation of $(x^{\mathbb{V}}\sigma)$ resulting in a term containing $a^{\mathbb{A}}$, just as
- $(P, N)$ blocked $x^{\mathbb{V}}$ before the application of $\sigma$.

## 5.6 Choice-Conditions

In the following we define choice-conditions as syntactic objects. They influence our semantics by a compatibility requirement, which will be described in Definition 7.4.

**Definition 5.12 (Choice-Condition, Choice Type)**
$C$ is a $(P,N)$-*choice-condition* if

- $(P,N)$ is a consistent positive/negative variable-condition and
- $C$ is a partial function on $\mathbb{V}$

such that, for every $y^{\mathbb{V}} \in \mathrm{dom}(C)$, the following items hold for some types $\alpha_0,\ldots,\alpha_l$:

1. The value $C(y^{\mathbb{V}})$ is of the form
$$\lambda v_0^{\mathbb{B}},\ldots,v_{l-1}^{\mathbb{B}}.\ \varepsilon v_l^{\mathbb{B}}.\ B$$
for some formula $B$ and for some mutually distinct bound atoms $v_0^{\mathbb{B}},\ldots,v_l^{\mathbb{B}} \in \mathbb{B}$ with $\mathbb{B}(B) \subseteq \{v_0^{\mathbb{B}},\ldots,v_l^{\mathbb{B}}\}$ and $v_0^{\mathbb{B}} : \alpha_0,\ \ldots,\ v_l^{\mathbb{B}} : \alpha_l$.
2. $y^{\mathbb{V}} : \alpha_0,\ldots,\alpha_{l-1} \rightarrow \alpha_l$.
3. $z^{\mathbb{VA}}\ P^+\ y^{\mathbb{V}}$ for all $z^{\mathbb{VA}} \in \mathbb{VA}(C(y^{\mathbb{V}}))$.

In the situation described, $\alpha_l$ is *the choice type of* $C(y^{\mathbb{V}})$.
$\beta$ is *a choice type of* $C$ if there is a $z^{\mathbb{V}} \in \mathrm{dom}(C)$ such that $\beta$ is the choice type of $C(z^{\mathbb{V}})$.

**Example 5.13 (Choice-Condition)** *(continuing Example 4.11)*

(a) If $(P,N)$ is a consistent positive/negative variable-condition that satisfies
$$z_a^{\mathbb{V}}\ P\ y_a^{\mathbb{V}}\ P\ z_b^{\mathbb{V}}\ P\ x_a^{\mathbb{V}}\ P\ z_c^{\mathbb{V}}\ P\ y_b^{\mathbb{V}}\ P\ z_d^{\mathbb{V}}\ P\ w_a^{\mathbb{V}}\ P\ z_e^{\mathbb{V}}\ P\ y_c^{\mathbb{V}}\ P\ z_f^{\mathbb{V}}\ P\ x_b^{\mathbb{V}}\ P\ z_g^{\mathbb{V}}\ P\ y_d^{\mathbb{V}}\ P\ z_h^{\mathbb{V}},$$
then the $C$ of Example 4.11 is a $(P,N)$-choice-condition, indeed.

---

[3] ($\sigma$-Updates Admitting Variable-Reuse and -Permutation)
For a version of $\sigma$-updates that admits variable-reuse and -permutation as explained in Note 10 of [Wirth, 2004] and executed in Notes 26–30 of [Wirth, 2004], the $\sigma$-update has to forget about the old meaning of the variables in $\mathrm{dom}(\sigma)$. To this end — instead of the simpler $(P \cup D, N)$ — we have to chose a *$\sigma$-update admitting variable-reuse and -permutation* to be
$$\Big(\ \big(\ {}_{\mathbb{VA}\setminus\mathrm{dom}(\sigma)}{\upharpoonleft} P\ \cup\ P' \circ P\ \big){\upharpoonright}_{\mathbb{V}\setminus\mathrm{dom}(\sigma)},\qquad {}_{\mathbb{V}\setminus\mathrm{dom}(\sigma)}{\upharpoonleft} N\ \cup\ {}_{\mathbb{V}}{\upharpoonleft} P' \circ N\ \Big)$$
for $P' := D\ \cup\ {}_{\mathbb{VA}\setminus\mathrm{dom}(\sigma)}{\upharpoonleft}(P{\upharpoonright}_{\mathrm{dom}(\sigma)})^+$.
Note that $P'$ can be simplified to $D$ here by taking as the *$\sigma$-update admitting* $\mathrm{V}_\gamma$*-reuse and -permutation:*
$$\Big(\ \big(\ {}_{\mathbb{A}\cup\mathrm{V}_{\delta^+}\cup(\mathrm{V}_\gamma\setminus\mathrm{dom}(\sigma))}{\upharpoonleft} P\ \cup\ D \circ P\ \cup\ D{\upharpoonright}_{\mathrm{V}_{\delta^+}}\ \big),\quad {}_{\mathrm{V}_{\delta^+}\cup(\mathrm{V}_\gamma\setminus\mathrm{dom}(\sigma))}{\upharpoonleft} N\ \cup\ {}_{\mathbb{V}}{\upharpoonleft} D{\upharpoonright}_{\mathrm{V}_\gamma\cap\mathrm{dom}(\sigma)} \circ N\ \Big),$$
provided that we partition $\mathbb{V}$ into two sets $\mathrm{V}_{\delta^+} \uplus \mathrm{V}_\gamma$, use $\mathrm{V}_{\delta^+}$ as the possible domain of the choice-conditions, and admit variable-reuse and -permutation only on $\mathrm{V}_\gamma$, similar to what we already did in Note 10 of [Wirth, 2004]. (The crucial restriction becomes here the following: For a (positive/negative) $\sigma$-update $(P'',N'')$ admitting $\mathrm{V}_\gamma$-reuse and -permutation we have $P'' \subseteq \mathbb{VA} \times \mathrm{V}_{\delta^+}$ and $N'' \subseteq \mathbb{V} \times \mathbb{A}$ ).
Note, however, that it is actually better to work with the more complicated $P'$, simply because it is more general and because the transitive closure will not be computed in practice, but a graph will be updated just as exemplified in Note 10 of [Wirth, 2004].

(b) If some clever person tried to do the entire quantifier elimination of Example 4.11 by

$$\begin{array}{lcl} C'(z^{\mathbb{V}}_h) & := & \varepsilon z^{\mathbb{B}}_h.\ \neg\mathsf{P}(w^{\mathbb{V}}_a, x^{\mathbb{V}}_b, y^{\mathbb{V}}_d, z^{\mathbb{B}}_h) \\ C'(y^{\mathbb{V}}_d) & := & \varepsilon y^{\mathbb{B}}_d.\ \mathsf{P}(w^{\mathbb{V}}_a, x^{\mathbb{V}}_b, y^{\mathbb{B}}_d, z^{\mathbb{V}}_h) \\ C'(x^{\mathbb{V}}_b) & := & \varepsilon x^{\mathbb{B}}_b.\ \neg\mathsf{P}(w^{\mathbb{V}}_a, x^{\mathbb{B}}_b, y^{\mathbb{V}}_d, z^{\mathbb{V}}_h) \\ C'(w^{\mathbb{V}}_a) & := & \varepsilon w^{\mathbb{B}}_a.\ \mathsf{P}(w^{\mathbb{B}}_a, x^{\mathbb{V}}_b, y^{\mathbb{V}}_d, z^{\mathbb{V}}_h) \end{array}$$

then he would — among other constraints — have to satisfy $z^{\mathbb{V}}_h\ P^+\ y^{\mathbb{V}}_d\ P^+\ z^{\mathbb{V}}_h$, because of item 3 of Definition 5.12 and the values of $C'$ at $y^{\mathbb{V}}_d$ and $z^{\mathbb{V}}_h$. This would make $P$ non-well-founded. Thus, this $C'$ cannot be a $(P,N)$-choice-condition for any $(P,N)$, because the consistency of $(P,N)$ is required in Definition 5.12. Note that the choices required by $C'$ for $y^{\mathbb{V}}_d$ and $z^{\mathbb{V}}_h$ are in an unsolvable conflict, indeed.

(c) For a more elementary example, take

$$C''(x^{\mathbb{V}}) \ := \ \varepsilon x^{\mathbb{B}}.\ (x^{\mathbb{B}} = y^{\mathbb{V}}) \qquad\qquad C''(y^{\mathbb{V}}) \ := \ \varepsilon y^{\mathbb{B}}.\ (x^{\mathbb{V}} \neq y^{\mathbb{B}})$$

Then $x^{\mathbb{V}}$ and $y^{\mathbb{V}}$ form a vicious circle of conflicting choices for which no valuation can be found that is compatible with $C''$. But $C''$ is no choice-condition at all because there is no *consistent* positive/negative variable-condition $(P,N)$ such that $C''$ is a $(P,N)$-choice-condition.

## 5.7 Extending Extensions and $\sigma$-Updates to Choice-Conditions

Just like the positive/negative variable-condition $(P,N)$, the $(P,N)$-choice-condition $C$ may be extended during proofs. This kind of extension plays an important rôle in inference:

**Definition 5.14 (Extended Extension)**
$(C',(P',N'))$ is an *extended extension of* $(C,(P,N))$ if

- $C$ is a $(P,N)$-choice-condition (cf. Definition 5.12),
- $C'$ is a $(P',N')$-choice-condition,
- $(P',N')$ is an extension of $(P,N)$ (cf. Definition 5.7), and
- $C \subseteq C'$.

**Corollary 5.15** *Being an extended extension is a reflexive ordering.*

After global application of a $(P,N)$-substitution $\sigma$, we now have to update both $(P,N)$ and $C$:

**Definition 5.16 (Extended $\sigma$-Update)**
Let $C$ be a $(P,N)$-choice-condition and let $\sigma$ be a $(P,N)$-substitution.
The *extended* $\sigma$*-update* $(C',(P',N'))$ *of* $(C,(P,N))$ is given as follows:

$$C' \ := \ \{\ (x^{\mathbb{V}}, A\sigma) \mid (x^{\mathbb{V}}, A) \in C \wedge x^{\mathbb{V}} \notin \mathrm{dom}(\sigma)\ \},$$

$(P',N')$ is the $\sigma$-update of $(P,N)$ (cf. Definition 5.10).

Note that a $\sigma$-update (cf. Definition 5.10) is an extension (cf. Definition 5.7), whereas an extended $\sigma$-update is not an extended extension in general, because entries of the choice-condition may be modified or even deleted, such that we may have $\mathrm{ran}(C) \not\subseteq \mathrm{ran}(C')$ and $\mathrm{dom}(C) \not\subseteq \mathrm{dom}(C')$. In case of $x^{\mathbb{V}} \in \mathrm{dom}(\sigma) \cap \mathrm{dom}(C)$ we get $x^{\mathbb{V}} \notin \mathrm{dom}(C')$, which is no problem because of $x^{\mathbb{V}} \notin \mathbb{V}(\sigma(x^{\mathbb{V}}))$ (as $\sigma$ is a $(P, N)$-substitution), and thus $x^{\mathbb{V}}$ disappears from the whole proof attempt after global application of $\sigma$. The remaining properties of an extended extension, however, are satisfied:

**Lemma 5.17 (Extended $\sigma$-Update)** *Let $C$ be a $(P, N)$-choice-condition. Let $\sigma$ be a $(P, N)$-substitution. Let $(C', (P', N'))$ be the extended $\sigma$-update of $(C, (P, N))$. Then $C'$ is a $(P', N')$-choice-condition.*

# 6 Example: Henkin Quantification

If the previous examples were sufficient for understanding variable-conditions and there is no interest in Henkin quantification and the extra powers of our *positive/negative* variable-condition, then this § 6 should be skipped; the later sections to not depend on it.

In [Wirth, 2006b, § 6.4.1], we showed that Henkin quantification was problematic for the variable-conditions of that paper, which had only one component, namely the positive one of our positive/negative variable-conditions here: Indeed, there the only way to model an example of a Henkin quantification precisely was to increase the order of some variables by raising. Let us consider the same example here again and show that now we can model its Henkin quantification directly with a *consistent* positive/*negative* variable-condition, even *without raising.*

**Example 6.1 (Henkin Quantification)** In [Hintikka, 1974], quantifiers in first-order logic were found insufficient to give the precise semantics of some English sentences. In [Hintikka, 1996], *IF logic,* i.e. Independence-Friendly logic — a first-order logic with more flexible quantifiers in the sense that their quantifier elimination can result in smaller variable-conditions — was presented to overcome this weakness. In [Hintikka, 1974], we find the following sentence:

> Some relative of each villager and
> some relative of each townsman hate each other. (H0)

Let us first change to a lovelier subject:

> Some potential loved one of each woman and
> some potential loved one of each man could love each other. (H1)

For our purposes here, we consider (H1) to be equivalent to the following sentence, which may be more meaningful and easier to understand:

> For each person, one of those this person could love can be chosen, such that
> the choice for any woman and the choice for any man could love each other.

(H1) can be represented by the following Henkin-quantified IF-logic formula:

$$\forall x_0^{\mathbb{B}}.\ \forall y_0^{\mathbb{B}}.\ \left( \left( \begin{array}{ll} & \mathsf{Female}(x_0^{\mathbb{B}}) \\ \wedge & \mathsf{Male}(y_0^{\mathbb{B}}) \end{array} \right) \ \Rightarrow\ \exists y_1^{\mathbb{B}}/y_0^{\mathbb{B}}.\ \exists x_1^{\mathbb{B}}/x_0^{\mathbb{B}}.\ \left( \begin{array}{ll} & \mathsf{Loves}(x_0^{\mathbb{B}}, y_1^{\mathbb{B}}) \\ \wedge & \mathsf{Loves}(y_0^{\mathbb{B}}, x_1^{\mathbb{B}}) \\ \wedge & \mathsf{Loves}(y_1^{\mathbb{B}}, x_1^{\mathbb{B}}) \\ \wedge & \mathsf{Loves}(x_1^{\mathbb{B}}, y_1^{\mathbb{B}}) \end{array} \right) \right) \tag{H2}$$

Let us refer to the standard game-theoretic semantics for quantifiers (cf. e.g. [Hintikka, 1996]), which is defined as follows: Witnesses have to be picked for the quantified variables outside-in. We have to pick the witnesses for the $\gamma$-quantifiers (i.e., in (H2), for the existential quantifiers), and our opponent in the game picks the witnesses for the $\delta$-quantifiers (i.e. for the universal quantifiers in (H2)). We win if the resulting quantifier-free formula evaluates to true. A formula is true if we have a winning strategy.

Then an IF-logic quantifier such as "$\exists y_1^{\mathbb{B}}/y_0^{\mathbb{B}}$." in (H2) is a special quantifier, which is a bit different from "$\exists y_1^{\mathbb{B}}$.". Game-theoretically, it has the following semantics: It asks us to pick the love $y_1^{\mathbb{B}}$ independently from the choice of the man $y_0^{\mathbb{B}}$ (by our opponent in the game), although the IF-logic quantifier occurs in the scope of the quantifier "$\forall y_0^{\mathbb{B}}$.".

Note that Formula (H2) is already close to anti-prenex form. In fact, if we move its quantifiers closer toward the leaves of the formula tree, this does not admit us to reduce their dependencies. It is more interesting, however, to move the quantifiers of (H2) out — to obtain prenex form — and then to simplify the prenex by using the equivalence of "$\forall y^{\mathbb{B}}_0.\ \exists y^{\mathbb{B}}_1/y^{\mathbb{B}}_0.$" and "$\exists y^{\mathbb{B}}_1.\ \forall y^{\mathbb{B}}_0.$", resulting in:

$$\forall x^{\mathbb{B}}_0.\ \exists y^{\mathbb{B}}_1.\ \forall y^{\mathbb{B}}_0.\ \exists x^{\mathbb{B}}_1/x^{\mathbb{B}}_0.\ \left(\left(\begin{array}{rl} & \mathsf{Female}(x^{\mathbb{B}}_0) \\ \wedge & \mathsf{Male}(y^{\mathbb{B}}_0) \end{array}\right) \Rightarrow \left(\begin{array}{rl} & \mathsf{Loves}(x^{\mathbb{B}}_0, y^{\mathbb{B}}_1) \\ \wedge & \mathsf{Loves}(y^{\mathbb{B}}_0, x^{\mathbb{B}}_1) \\ \wedge & \mathsf{Loves}(y^{\mathbb{B}}_1, x^{\mathbb{B}}_1) \\ \wedge & \mathsf{Loves}(x^{\mathbb{B}}_1, y^{\mathbb{B}}_1) \end{array}\right)\right) \qquad (\mathrm{H2}')$$

Note that this formula is stronger than the following formula with standard quantifiers:

$$\forall x^{\mathbb{B}}_0.\ \exists y^{\mathbb{B}}_1.\ \forall y^{\mathbb{B}}_0.\ \exists x^{\mathbb{B}}_1.\ \left(\left(\begin{array}{rl} & \mathsf{Female}(x^{\mathbb{B}}_0) \\ \wedge & \mathsf{Male}(y^{\mathbb{B}}_0) \end{array}\right) \Rightarrow \left(\begin{array}{rl} & \mathsf{Loves}(x^{\mathbb{B}}_0, y^{\mathbb{B}}_1) \\ \wedge & \mathsf{Loves}(y^{\mathbb{B}}_0, x^{\mathbb{B}}_1) \\ \wedge & \mathsf{Loves}(y^{\mathbb{B}}_1, x^{\mathbb{B}}_1) \\ \wedge & \mathsf{Loves}(x^{\mathbb{B}}_1, y^{\mathbb{B}}_1) \end{array}\right)\right) \qquad (\mathrm{S2}')$$

An alternative way to define the semantics of IF-logic quantifiers is by describing their effect on the equivalent *raised* forms of the formulas in which they occur. *Raising* is a dual of SKOLEMization, cf. [MILLER, 1992]. The raised form of (S2′) is the following:

$$\exists y^{\mathbb{B}}_1.\ \exists x^{\mathbb{B}}_1.\ \forall x^{\mathbb{B}}_0.\ \forall y^{\mathbb{B}}_0.\ \left(\left(\begin{array}{rl} & \mathsf{Female}(x^{\mathbb{B}}_0) \\ \wedge & \mathsf{Male}(y^{\mathbb{B}}_0) \end{array}\right) \Rightarrow \left(\begin{array}{rl} & \mathsf{Loves}(x^{\mathbb{B}}_0, y^{\mathbb{B}}_1(x^{\mathbb{B}}_0)) \\ \wedge & \mathsf{Loves}(y^{\mathbb{B}}_0, x^{\mathbb{B}}_1(y^{\mathbb{B}}_0, x^{\mathbb{B}}_0)) \\ \wedge & \mathsf{Loves}(y^{\mathbb{B}}_1(x^{\mathbb{B}}_0), x^{\mathbb{B}}_1(y^{\mathbb{B}}_0, x^{\mathbb{B}}_0)) \\ \wedge & \mathsf{Loves}(x^{\mathbb{B}}_1(y^{\mathbb{B}}_0, x^{\mathbb{B}}_0), y^{\mathbb{B}}_1(x^{\mathbb{B}}_0)) \end{array}\right)\right) \qquad (\mathrm{S3})$$

For HENKIN-quantified IF-logic formulas, the raised form is defined as usual, besides that a $\gamma$-quantifier, say "$\exists x^{\mathbb{B}}_1.$", followed by a slash as in "$\exists x^{\mathbb{B}}_1/x^{\mathbb{B}}_0.$", is raised such that $x^{\mathbb{B}}_0$ does not appear as an argument to the raising function for $x^{\mathbb{B}}_1$. Accordingly, *mutatis mutandis*, (H2) as well as (H2′) are equivalent to their common raised form (H3) below, where $x^{\mathbb{B}}_0$ does not occur as an argument to the raising function $x^{\mathbb{B}}_1$ — contrary to (S3), which is strictly implied by (H3) because we can choose the love of the woman differently for different men.

$$\exists y^{\mathbb{B}}_1.\ \exists x^{\mathbb{B}}_1.\ \forall x^{\mathbb{B}}_0.\ \forall y^{\mathbb{B}}_0.\ \left(\left(\begin{array}{rl} & \mathsf{Female}(x^{\mathbb{B}}_0) \\ \wedge & \mathsf{Male}(y^{\mathbb{B}}_0) \end{array}\right) \Rightarrow \left(\begin{array}{rl} & \mathsf{Loves}(x^{\mathbb{B}}_0, y^{\mathbb{B}}_1(x^{\mathbb{B}}_0)) \\ \wedge & \mathsf{Loves}(y^{\mathbb{B}}_0, x^{\mathbb{B}}_1(y^{\mathbb{B}}_0)) \\ \wedge & \mathsf{Loves}(y^{\mathbb{B}}_1(x^{\mathbb{B}}_0), x^{\mathbb{B}}_1(y^{\mathbb{B}}_0)) \\ \wedge & \mathsf{Loves}(x^{\mathbb{B}}_1(y^{\mathbb{B}}_0), y^{\mathbb{B}}_1(x^{\mathbb{B}}_0)) \end{array}\right)\right) \qquad (\mathrm{H3})$$

Now, (H3) looks already very much like the following tentative representation of (H1) in our framework of variables and free atoms:

$$\left(\begin{array}{rl} & \mathsf{Female}(x^{\mathbb{A}}_0) \\ \wedge & \mathsf{Male}(y^{\mathbb{A}}_0) \end{array}\right) \Rightarrow \left(\begin{array}{rl} & \mathsf{Loves}(x^{\mathbb{A}}_0, y^{\mathbb{V}}_1) \\ \wedge & \mathsf{Loves}(y^{\mathbb{A}}_0, x^{\mathbb{V}}_1) \\ \wedge & \mathsf{Loves}(y^{\mathbb{V}}_1, x^{\mathbb{V}}_1) \\ \wedge & \mathsf{Loves}(x^{\mathbb{V}}_1, y^{\mathbb{V}}_1) \end{array}\right) \qquad (\mathrm{H1}')$$

with choice-condition $C$ given by

$$\begin{array}{rcl} C(y^{\mathbb{V}}_1) & := & \varepsilon y^{\mathbb{B}}_1.\ (\mathsf{Female}(x^{\mathbb{A}}_0) \Rightarrow \mathsf{Loves}(x^{\mathbb{A}}_0, y^{\mathbb{B}}_1)) \\ C(x^{\mathbb{V}}_1) & := & \varepsilon x^{\mathbb{B}}_1.\ (\mathsf{Male}(y^{\mathbb{A}}_0) \Rightarrow \mathsf{Loves}(y^{\mathbb{A}}_0, x^{\mathbb{B}}_1)) \end{array}$$

which requires our positive/negative variable-condition $(P, N)$ to contain $(x^{\mathbb{A}}_0, y^{\mathbb{V}}_1)$ and $(y^{\mathbb{A}}_0, x^{\mathbb{V}}_1)$ in the positive relation $P$ (by item 3 of Definition 5.12). This choice-condition mirrors the structure of the natural-language sentence (H1) as close as possible. Actually, however, we do not need any non-empty choice-condition here at all. Indeed, to find a representa-tion in our framework, we can also work with an empty choice-condition. Crucial for our discussion, however, is that we can have

$$(x_0^{\mathbb{A}}, y_1^{\mathbb{V}}), (y_0^{\mathbb{A}}, x_1^{\mathbb{V}}) \in P.$$

Indeed: Otherwise the loves could not depend on their lovers. And this means that in a proof attempt for this sentence we cannot choose the woman's love according to the woman we are treating, such that we would have to select the loves of Bloody Mary and Audrey Hepburn by means of the same description. Thus, in the following, let us definitely put these two pairs into $P$, no matter whether we do this because of the choice-conditions supporting natural language representation or because we want to have a chance to complete a proof of this sentence later.

In any case, we can add $(y_1^{\mathbb{V}}, y_0^{\mathbb{A}})$ to the negative relation $N$ here, namely to express that $y_1^{\mathbb{V}}$ must not read $y_0^{\mathbb{A}}$. Then we obtain:

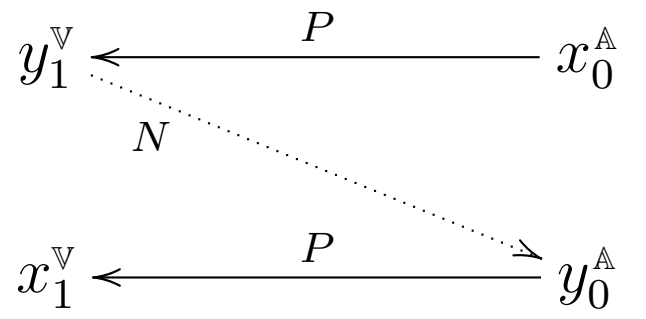

The same variable-condition is also obtained if we start with the empty variable-condition $(P, N) := (\emptyset, \emptyset)$, remove all quantifiers from (S2′) with our $\gamma$- and $\delta^-$-rules, and then add $\{(x_0^{\mathbb{A}}, y_1^{\mathbb{V}}), (y_0^{\mathbb{A}}, x_1^{\mathbb{V}})\}$ to $P$.

The corresponding procedure for (H2′), however, has to add also $(x_1^{\mathbb{V}}, x_0^{\mathbb{A}})$ to $N$ as part of the last $\gamma$-step that removes the IF-logic quantifier "$\exists x_1^{\mathbb{B}}/x_0^{\mathbb{A}}.$" and replaces $x_1^{\mathbb{B}}$ with $x_1^{\mathbb{V}}$. After this procedure, our current positive/negative variable-condition is now given as $(P, N)$ with $P = \{(x_0^{\mathbb{A}}, y_1^{\mathbb{V}}), (y_0^{\mathbb{A}}, x_1^{\mathbb{V}})\}$ and $N = \{(y_1^{\mathbb{V}}, y_0^{\mathbb{A}}), (x_1^{\mathbb{V}}, x_0^{\mathbb{A}})\}$. Thus, we have a single cycle in the graph, namely the following one:

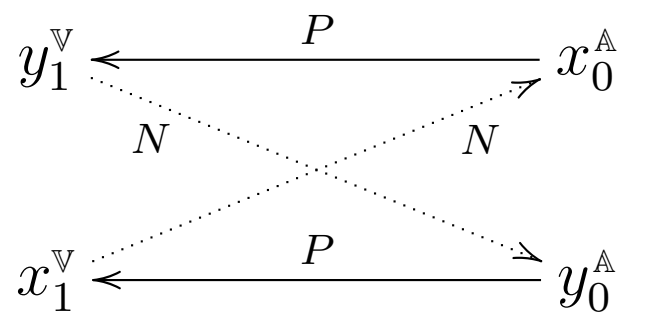

But this cycle necessarily has two edges from the negative relation $N$. Thus, in spite of this cycle, our positive/negative variable-condition $(P, N)$ is consistent by Corollary 5.6.

With the variable-conditions of [Wirth, 2002; 2004; 2006a; 2008; 2012b; 2006b], however, this cycle necessarily destroys the consistency, because they have no distinction between the edges of $N$ and $P$.

Therefore — if the discussion in [Wirth, 2006b, § 6.4.1] is sound — our new framework of this paper with positive/*negative* variable-conditions is the only one among all approaches suitable for describing the semantics of noun phrases in natural languages that admits us to model IF-logic and Henkin quantifiers without raising.

While the rules for $\delta$-quantifiers of IF logic work just like our normal $\delta$-rules (indeed, the law of the excluded middle fails to hold in IF logic in general), we can now formalize the inference rule for the *$\gamma$-quantifiers of IF logic* as follows:

Let $x^{\mathbb{VA}}$ be a variable or a free atom.
Let $t$ be any term not containing $x^{\mathbb{VA}}$ (i.e. $x^{\mathbb{VA}} \notin \mathbb{VA}(t)$):

$$\frac{\quad \varGamma \quad \exists y^{\mathbb{B}}/x^{\mathbb{VA}}.\ A \quad \varPi}{A\{y^{\mathbb{B}}{\mapsto}t\} \quad \varGamma \quad \exists y^{\mathbb{B}}/x^{\mathbb{VA}}.\ A \quad \varPi} \qquad \mathbb{V}(t) \times \{x^{\mathbb{VA}}\}$$

$$\frac{\quad \varGamma \quad \neg\forall y^{\mathbb{B}}/x^{\mathbb{VA}}.\ A \quad \varPi}{\neg A\{y^{\mathbb{B}}{\mapsto}t\} \quad \varGamma \quad \neg\forall y^{\mathbb{B}}/x^{\mathbb{VA}}.\ A \quad \varPi} \qquad \mathbb{V}(t) \times \{x^{\mathbb{VA}}\}$$

Here, $\mathbb{V}(t) \times \{x^{\mathbb{VA}}\}$ should be added to $N$, the negative part of the current positive/negative variable-condition $(P, N)$ — no matter whether we have the case $x^{\mathbb{VA}} \in \mathbb{V}$ or actually $x^{\mathbb{VA}} \in \mathbb{A}$. Note that the first of these two cases may violate our range restriction for the negative part given in Definition 5.4, but this range restriction was chosen here mainly for technical simplification (cf. § C).

Moreover, note that, because $x^{\mathbb{VA}}$ is not fresh but was typically introduced by a previous application of a $\delta^-$- or $\delta^+$-rule, the application of a $\gamma$-rule for IF-logic quantifiers could result in an inconsistent positive/negative variable-condition. Thus, we have to add the requirement for the consistency of the resulting variable-condition as a precondition for the application of these new inference rules.

With these $\gamma$-rules for IF-logic quantifiers, we can obtain the cyclic graph above from (H2) or (H2′) just as we obtained the non-cyclic graph above from (S2′). If we replace the two applications of $\delta^-$-rules here with two applications of $\delta^+$-rules and start from (H2), then the resulting graph becomes

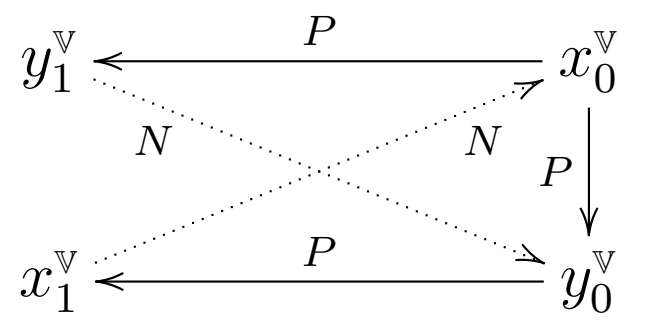

If we start from (H2′), we obtain

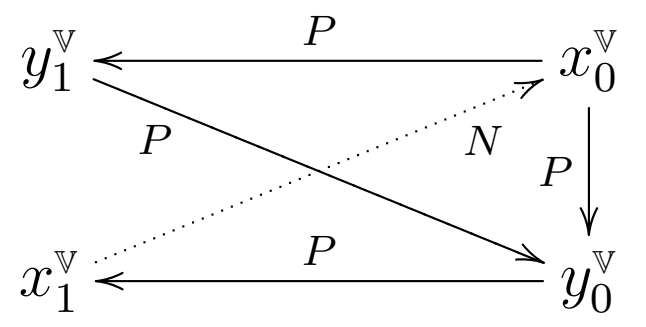

Each of these graphs has the same cycle with only one edge from the negative part $N$, which means that each of the variable-conditions is inconsistent. Thus, it seems that the application of $\delta^+$-rules to $\delta$-quantifiers with IF-logic $\gamma$-quantifiers in their scope is not to be recommended and the $\delta^-$-rules should be used instead, just as for outer $\delta$-quantifiers over which we want to do mathematical induction in the style of *descente infinie.* If we always do so, variables will hardly occur in the second component of IF-logic quantifiers, and then we can get along with the case of $x^{\mathbb{VA}} \in \mathbb{A}$ in the above new $\gamma$-rules and do not have to modify our range restriction on the negative part of our positive/negative variable-conditions.

The other direction in which one might cure the inconsistency is to liberalize our $\delta^+$-rules further. The variables and free atoms these rules introduce to the domain of the positive variable-condition have to cover exactly the ones occurring in the $\varepsilon$-term of its choice-condition. But we can simplify this $\varepsilon$-term in this case to

$$\varepsilon y_0^{\mathbb{B}}.\ \neg\exists x_1^{\mathbb{B}}/x_0^{\mathbb{V}}.\ (\mathsf{Male}(y_0^{\mathbb{B}}) \Rightarrow \mathsf{Loves}(y_0^{\mathbb{B}}, x_1^{\mathbb{B}})).$$

If we managed to liberalize our $\delta^+$-rules to work with this choice-condition here, in which neither $x_0^{\mathbb{V}}$ nor $y_1^{\mathbb{V}}$ occur (as the one in "$/x_0^{\mathbb{V}}$" does not count), then we could remove the downward edges for $P$ in each of the latter two diagrams, and consistency would be regained.

# 7 Formal Presentation of our Semantics

To satisfy Requirement III (Semantics) of § 4.2, we will now present our novel semantics for HILBERT's $\varepsilon$ formally. In §§ 7.1 and 7.2, we explain how to deal with variables. After formalizing the compatibility of choice-conditions (§ 7.3), we define our notion of validity and discuss some examples (§ 7.5). Our interest goes beyond soundness in that we want to have *"preservation of solutions"*. By this we mean the following: All *closing substitutions* for the variables — i.e. all solutions that transform a proof attempt (to which a proposition has been reduced) into a closed proof — are also solutions of the original proposition. This is similar to a proof in PROLOG (cf. [KOWALSKI, 1974], [CLOCKSIN & MELLISH, 2003]), computing answers to a query proposition that contains variables. Therefore, we discuss this solution-preserving notion of *reduction* (§ 8), in particular under the aspect of extensions of choice-conditions, and under the aspect of global instantiation of variables with choice-conditions ("$\varepsilon$-substitution"). Finally, in § 8.1, we show soundness, safeness, and solution-preservation for our $\gamma$-, $\delta^-$, and $\delta^+$-rules of §§ 3.4, 3.5, and 3.6.

All in all, we extend and simplify the presentation of [WIRTH, 2008], which already simplifies and extends the presentation of [WIRTH, 2004] and which is extended with additional linguistic applications in [WIRTH, 2006b]. Note, however, that [WIRTH, 2004] additionally contains some comparative discussions and compatible extensions for *descente infinie*, which also apply to our new version here.

## 7.1 Semantic Presuppositions

Instead of defining truth from scratch, we require some abstract properties typically holding in two-valued model semantics.

Truth is given relative to a $\Sigma$-structure $\mathcal{S}$, which provides some *non-empty set* as the universe (or "carrier", "domain") (for each type). Moreover, we assume that every $\Sigma$-structure $\mathcal{S}$ is not only defined on the predicate and function symbols of the signature $\Sigma$, but is defined also on the symbols $\forall$ and $\exists$ such that $\mathcal{S}(\exists)$ serves as a function-choice function for the universe function $\mathcal{S}(\forall)$ in the sense that, for each type $\alpha$ of $\Sigma$, the universe for the type $\alpha$ is denoted by $\mathcal{S}(\forall)_\alpha$ and $\mathcal{S}(\exists)_\alpha \in \mathcal{S}(\forall)_\alpha$.

For $\mathrm{X} \subseteq \mathbb{VAB}$, we denote the set of *partial $\mathcal{S}$-valuations of* X (i.e. the set of functions mapping a (possibly non-proper) subset of X to objects of the universe of $\mathcal{S}$) with

$$\mathrm{X} \rightsquigarrow \mathcal{S}\,,$$

and the set of *(total) $\mathcal{S}$-valuations of* X with

$$\mathrm{X} \rightarrow \mathcal{S}\,,$$

the subset of those $\delta : \mathrm{X} \rightsquigarrow \mathcal{S}$ that are total on X, i.e. those $\delta \in \mathrm{X} \rightsquigarrow \mathcal{S}$ with $\mathrm{dom}(\delta) = \mathrm{X}$. Here we always expect types to be respected in the sense that, for each $\delta : \mathrm{X} \rightsquigarrow \mathcal{S}$ and for each $x^{\mathbb{VAB}} \in \mathrm{dom}(\delta)$ with $x^{\mathbb{VAB}} : \alpha$ (i.e. $x^{\mathbb{VAB}}$ has type $\alpha$), we have $\delta(x^{\mathbb{VAB}}) \in \mathcal{S}(\forall)_\alpha$.

For $\delta : \mathrm{X} \rightarrow \mathcal{S}$, we denote with "$\mathcal{S} \uplus \delta$" the extension of $\mathcal{S}$ to X. More precisely, we assume some evaluation function "eval" such that $\mathrm{eval}(\mathcal{S} \uplus \delta)$ maps every term whose free-occurring symbols are from $\Sigma \uplus \mathrm{X}$ into the universe of $\mathcal{S}$ (respecting types). Moreover, $\mathrm{eval}(\mathcal{S} \uplus \delta)$ maps every formula $B$ whose free-occurring symbols are from $\Sigma \uplus \mathrm{X}$ to $\mathsf{TRUE}$ or $\mathsf{FALSE}$, such that:

$$B \text{ is true in } \mathcal{S} \uplus \delta \quad \text{iff} \quad \mathrm{eval}(\mathcal{S} \uplus \delta)(B) = \mathsf{TRUE}.$$

We leave open what our formulas and what our Σ-structures exactly are. The latter can range from first-order Σ-structures to higher-order modal Σ-models; provided that the following three properties — which (explicitly or implicitly) belong to the standard of most logic textbooks — hold for every term or formula $B$, every Σ-structure $\mathcal{S}$, and every $\mathcal{S}$-valuation $\delta : \mathbb{VAR} \rightsquigarrow \mathcal{S}$.

Explicitness Lemma

The value of the evaluation of $B$ depends only on the valuation of those variables and atoms that actually have free occurrences in $B$; i.e., for $\mathrm{X} := \mathbb{VAR}(B)$, if $\mathrm{X} \subseteq \mathrm{dom}(\delta)$, then:

$$\mathrm{eval}(\mathcal{S} \uplus \delta)(B) \quad = \quad \mathrm{eval}(\mathcal{S} \ \uplus \ {}_{\mathrm{X}}\!\upharpoonright\!\delta)(B).$$

Substitution [Value] Lemma

Let $\sigma$ be a substitution on $\mathbb{VAR}$. If $\mathbb{VAR}(B\sigma) \subseteq \mathrm{dom}(\delta)$, then:

$$\mathrm{eval}(\mathcal{S} \uplus \delta)(B\sigma) \quad = \quad \mathrm{eval}\Big( \ \mathcal{S} \ \ \uplus \ \ \big( \ ( \ \sigma \ \uplus \ {}_{\mathbb{VAR}\setminus\mathrm{dom}(\sigma)}\!\upharpoonright\!\mathrm{id} \ ) \ \circ \ \mathrm{eval}(\mathcal{S} \uplus \delta) \ \big) \ \Big)\Big( \ B \ \Big).$$

Valuation-Lemma($l \in \mathbf{N}$)

The evaluation function treats application terms from $\mathbb{VAR}$ straightforwardly in the sense that for every $v_0^{\mathbb{VAR}}, \ldots, v_{l-1}^{\mathbb{VAR}}, y^{\mathbb{VAR}} \in \mathrm{dom}(\delta)$ with $v_0^{\mathbb{VAR}} : \alpha_0, \ \ldots, \ v_{l-1}^{\mathbb{VAR}} : \alpha_{l-1}$, $y^{\mathbb{VAR}} : \alpha_0, \ldots, \alpha_{l-1} \to \alpha_l$ for some types $\alpha_0, \ldots, \alpha_{l-1}, \alpha_l$, we have:

$$\mathrm{eval}(\mathcal{S} \uplus \delta)(y^{\mathbb{VAR}}(v_0^{\mathbb{VAR}}, \ldots, v_{l-1}^{\mathbb{VAR}})) \ = \ \delta(y^{\mathbb{VAR}})(\delta(v_0^{\mathbb{VAR}}), \ldots, \delta(v_{l-1}^{\mathbb{VAR}})).$$

In the case of $l = 0$, this equation is meant to be read as $\mathrm{eval}(\mathcal{S} \uplus \delta)(y^{\mathbb{VAR}}) = \delta(y^{\mathbb{VAR}})$. For the case of $l \succ 0$ — a case we only need if choice-conditions with a non-empty $\lambda$-prefix in front of the $\varepsilon$-binder occur, such as in the proof of Theorem 7.5 — the variable $y^{\mathbb{VAR}}$ is a higher-order symbol. Besides this, however, the basic language of the general reasoning framework may well be first-order and does not have to include function application.

Moreover, in the cases where we explicitly refer to quantifiers, implication, or negation, such as in our inference rules of §§ 3.4, 3.5, and 3.6, or in our version of axiom ($\varepsilon_0$) (cf. Definition 4.12), and in the lemmas and theorems that refer to these (namely Lemmas 7.10 and 7.11 and Theorems 8.3, 8.5(6));[4] we have to know that the quantifiers, the implication, and the negation show the standard semantic behavior of classical logic:

∀-Lemma

Assume $\mathbb{VAR}(\forall x^{\mathbb{B}}.\ A) \subseteq \mathrm{dom}(\delta)$. The following two are equivalent:

- $\mathrm{eval}(\mathcal{S} \uplus \delta)(\forall x^{\mathbb{B}}.\ A) = \mathsf{TRUE}$
- $\mathrm{eval}(\mathcal{S} \uplus {}_{\mathbb{VAR}\setminus\{x^{\mathbb{B}}\}}\!\upharpoonright\!\delta \uplus \chi)(A) = \mathsf{TRUE}$ for every $\chi : \{x^{\mathbb{B}}\} \to \mathcal{S}$

∃-Lemma

Assume $\mathbb{VAR}(\exists x^{\mathbb{B}}.\ A) \subseteq \mathrm{dom}(\delta)$. The following two are equivalent:

- $\mathrm{eval}(\mathcal{S} \uplus \delta)(\exists x^{\mathbb{B}}.\ A) = \mathsf{TRUE}$,
- $\mathrm{eval}(\mathcal{S} \uplus {}_{\mathbb{VAR}\setminus\{x^{\mathbb{B}}\}}\!\upharpoonright\!\delta \uplus \chi)(A) = \mathsf{TRUE}$ for some $\chi : \{x^{\mathbb{B}}\} \to \mathcal{S}$

⇒-Lemma

Assume $\mathbb{VAR}(A \Rightarrow B) \subseteq \mathrm{dom}(\delta)$. The following two are equivalent:

- $\mathrm{eval}(\mathcal{S} \uplus \delta)(A \Rightarrow B) = \mathsf{TRUE}$
- $\mathrm{eval}(\mathcal{S} \uplus \delta)(A) = \mathsf{FALSE}$ or $\mathrm{eval}(\mathcal{S} \uplus \delta)(B) = \mathsf{TRUE}$

¬-Lemma

Assume $\mathbb{VAR}(A) \subseteq \mathrm{dom}(\delta)$. The following two are equivalent:

- $\mathrm{eval}(\mathcal{S} \uplus \delta)(A) = \mathsf{TRUE}$
- $\mathrm{eval}(\mathcal{S} \uplus \delta)(\neg A) = \mathsf{FALSE}$

## 7.2 Semantic Relations and $\mathcal{S}$-Raising-Valuations

We now come to some technical definitions required for our semantic (model-theoretic) counterparts of our syntactic $(P, N)$-substitutions.

Let $\mathcal{S}$ be a $\Sigma$-structure. An *$\mathcal{S}$-raising-valuation* $\pi$ plays the rôle of a *raising function*, a dual of a SKOLEM function as defined in [MILLER, 1992]. This means that $\pi$ does not simply map each variable directly to an object of $\mathcal{S}$ (of the same type), but may additionally read the values of some free atoms under an $\mathcal{S}$-valuation $\tau : \mathbb{A} \to \mathcal{S}$. More precisely, we assume that $\pi$ takes some restriction of $\tau$ as a second argument, say $\tau' : \mathbb{A} \leadsto \mathcal{S}$ with $\tau' \subseteq \tau$. In short:

$$\pi : \mathbb{V} \to (\mathbb{A} \leadsto \mathcal{S}) \leadsto \mathcal{S}.$$

Moreover, for each variable $x^{\mathbb{V}}$, we require that the set $\mathrm{dom}(\tau')$ of atoms read by $\pi(x^{\mathbb{V}})$ is identical for all $\tau$. This identical set will be denoted with $S_\pi\langle\!\langle\{x^{\mathbb{V}}\}\rangle\!\rangle$ below. Technically, we require that there is some "semantic relation" $S_\pi \subseteq \mathbb{A}\times\mathbb{V}$ such that for all $x^{\mathbb{V}} \in \mathbb{V}$:

$$\pi(x^{\mathbb{V}}) \;:\; (S_\pi\langle\!\langle\{x^{\mathbb{V}}\}\rangle\!\rangle \to \mathcal{S}) \to \mathcal{S}.$$

This means that $\pi(x^{\mathbb{V}})$ can read the $\tau$-value of $y^{\mathbb{A}}$ if and only if $(y^{\mathbb{A}}, x^{\mathbb{V}}) \in S_\pi$. Note that, for each $\pi : \mathbb{V} \to (\mathbb{A} \leadsto \mathcal{S}) \leadsto \mathcal{S}$, at most one such semantic relation exists, namely the one of the following definition.

**Definition 7.1 (Semantic Relation ($S_\pi$))**
The *semantic relation for* $\pi$ is

$$S_\pi \;:=\; \{\ (y^{\mathbb{A}}, x^{\mathbb{V}}) \mid x^{\mathbb{V}} \in \mathbb{V} \wedge y^{\mathbb{A}} \in \mathrm{dom}(\bigcup(\mathrm{dom}(\pi(x^{\mathbb{V}}))))\ \}.$$

Let us explain our intention with the operator sequence of the set from which $y^{\mathbb{A}}$ is taken here: For $\pi : \mathbb{V} \to (\mathbb{A} \leadsto \mathcal{S}) \leadsto \mathcal{S}$ and $x^{\mathbb{V}} \in \mathbb{V}$ we first get that $\mathrm{dom}(\pi(x^{\mathbb{V}}))$ is a subset of $\mathbb{A} \leadsto \mathcal{S}$, which again is a subset of the power-set of $\mathbb{A} \times \bigcup_\alpha \mathcal{S}(\forall)_\alpha$; therefore we get $\bigcup \mathrm{dom}(\pi(x^{\mathbb{V}})) \subseteq \mathbb{A} \times \bigcup_\alpha \mathcal{S}(\forall)_\alpha$ and $\mathrm{dom}(\bigcup \mathrm{dom}(\pi(x^{\mathbb{V}}))) \subseteq \mathbb{A}$. This means that $S_\pi\langle\!\langle\{x^{\mathbb{V}}\}\rangle\!\rangle$ collects all atoms $y^{\mathbb{A}}$ which may ever occur in the domain of some $\tau : \mathbb{A} \leadsto \mathcal{S}$ on which $\pi(x^{\mathbb{V}})$ is defined. Yet, we still have to specify that these $\tau$ have all the same domain:

**Definition 7.2 ($\mathcal{S}$-Raising-Valuation)**
Let $\mathcal{S}$ be a $\Sigma$-structure. $\pi$ is an *$\mathcal{S}$-raising-valuation* if

$$\pi : \mathbb{V} \to (\mathbb{A} \leadsto \mathcal{S}) \leadsto \mathcal{S}$$

and, for all $x^{\mathbb{V}} \in \mathrm{dom}(\pi)$:

$$\pi(x^{\mathbb{V}}) : (S_\pi\langle\!\langle\{x^{\mathbb{V}}\}\rangle\!\rangle \to \mathcal{S}) \to \mathcal{S}.$$

Finally, we need the technical means to turn an $\mathcal{S}$-raising-valuation $\pi$ together with an $\mathcal{S}$-valuation $\tau$ of the atoms into an $\mathcal{S}$-valuation $\mathsf{e}(\pi)(\tau)$ of the variables:

**Definition 7.3 ($\mathsf{e}$)**
We define the function $\mathsf{e} : \quad (\mathbb{V} \leadsto (\mathbb{A} \leadsto \mathcal{S}) \leadsto \mathcal{S}) \;\to\; (\mathbb{A} \to \mathcal{S}) \;\to\; \mathbb{V} \;\leadsto\; \mathcal{S}$

for $\qquad \pi : \mathbb{V} \leadsto (\mathbb{A} \leadsto \mathcal{S}) \leadsto \mathcal{S}, \qquad \tau : \mathbb{A} \to \mathcal{S}, \;\; x^{\mathbb{V}} \in \mathbb{V}$

by $\qquad \mathsf{e}(\pi)(\tau)(x^{\mathbb{V}}) := \pi(x^{\mathbb{V}})({}_{S_\pi\langle\!\langle\{x^{\mathbb{V}}\}\rangle\!\rangle}\!\upharpoonright\!\tau).$

The "$\mathsf{e}$" stands for "<u>e</u>valuation" and replaces an "$\epsilon$" used in previous publications, which was too easily confused with the symbol for HILBERT's $\varepsilon$.

## 7.3 Compatibility of Choice-Conditions

**Definition 7.4 (Compatibility)**
Let $C$ be a $(P,N)$-choice-condition. Let $\mathcal{S}$ be a $\Sigma$-structure.
$\pi$ is *$\mathcal{S}$-compatible with* $(C,(P,N))$ if the following items hold:

1. $\pi$ is an $\mathcal{S}$-raising-valuation and $(P \cup S_\pi, N)$ is a consistent positive/negative variable-condition, cf. Definitions 7.2, 7.1, and 5.5.
2. For every $y^{\mathbb{V}} \in \mathrm{dom}(C)$, with $C(y^{\mathbb{V}}) = \lambda v_0^{\mathbb{B}},\ldots,v_{l-1}^{\mathbb{B}}.\ \varepsilon v_l^{\mathbb{B}}.\ B$ and $\mathbb{B}(B) \subseteq \{v_0^{\mathbb{B}},\ldots,v_l^{\mathbb{B}}\}$ for some formula $B$, and for every $\tau : \mathbb{A} \to \mathcal{S}$, and for every $\chi : \{v_0^{\mathbb{B}},\ldots,v_l^{\mathbb{B}}\} \to \mathcal{S}$:
   If $B$ is true in $\mathcal{S} \uplus \mathsf{e}(\pi)(\tau) \uplus \tau \uplus \chi$,
   then $B\{v_l^{\mathbb{B}} \mapsto y^{\mathbb{V}}(v_0^{\mathbb{B}},\ldots,v_{l-1}^{\mathbb{B}})\}$ is true in $\mathcal{S} \uplus \mathsf{e}(\pi)(\tau) \uplus \tau \uplus \chi$ as well.

To understand item 2 of Definition 7.4, let us consider a $(P,N)$-choice-condition

$$C := \{(y^{\mathbb{V}},\ \lambda v_0^{\mathbb{B}},\ldots,v_{l-1}^{\mathbb{B}}.\ \varepsilon v_l^{\mathbb{B}}.\ B)\},$$

which restricts the value of $y^{\mathbb{V}}$ according to the term $\lambda v_0^{\mathbb{B}},\ldots,v_{l-1}^{\mathbb{B}}.\ \varepsilon v_l^{\mathbb{B}}.\ B$. Then, roughly speaking, this choice-condition $C$ requires that whenever there is a $\chi$-value of $v_l^{\mathbb{B}}$ such that $B$ is true in $\mathcal{S} \uplus \mathsf{e}(\pi)(\tau) \uplus \tau \uplus \chi$, the $\pi$-value of $y^{\mathbb{V}}$ is chosen in such a way that

$$B\{v_l^{\mathbb{B}} \mapsto y^{\mathbb{V}}(v_0^{\mathbb{B}},\ldots,v_{l-1}^{\mathbb{B}})\}$$

becomes true in $\mathcal{S} \uplus \mathsf{e}(\pi)(\tau) \uplus \tau \uplus \chi$ as well. Note that the variables of the formula $B\{v_l^{\mathbb{B}} \mapsto y^{\mathbb{V}}(v_0^{\mathbb{B}},\ldots,v_{l-1}^{\mathbb{B}})\}$ cannot read the $\chi$-value of any of the bound atoms $v_0^{\mathbb{B}},\ldots,v_l^{\mathbb{B}}$, because variables can never depend on (the values of) any bound atoms.

Moreover, item 2 of Definition 7.4 is closely related to HILBERT's $\varepsilon$-operator in the sense that — roughly speaking — $y^{\mathbb{V}}$ must be given one of the values admissible for

$$\lambda v_0^{\mathbb{B}},\ldots,v_{l-1}^{\mathbb{B}}.\ \varepsilon v_l^{\mathbb{B}}.\ B.$$

As the choice for $y^{\mathbb{V}}$ depends on the symbols that have a free occurrence in that term, we included these dependencies into the positive relation $P$ of the consistent positive/negative variable-condition $(P,N)$ in item 3 of Definition 5.12 (Choice-Condition). By this inclusion, conflicts like the one shown in Example 5.13(c) are obviated.

## 7.4 Existence of Compatible Raising-Valuations

Let $(P,N)$ be a consistent positive/negative variable-condition. Then the empty function $\emptyset$ is a $(P,N)$-choice-condition. Moreover, each $\pi : \mathbb{V} \to \{\emptyset\} \to \mathcal{S}$ is $\mathcal{S}$-compatible with $(\emptyset,(P,N))$ because of $S_\pi = \emptyset$.

Furthermore, assuming an adequate Axiom of Choice on the meta level, a compatible $\pi$ always exists according to the following Theorem 7.5. This existence mainly relies on item 3 of Definition 5.12 and on the well-foundedness of $P$. This theorem is only technically complicated and, roughly speaking, just says that we have formalized our $(P,N)$-choice-conditions well, in the sense that, for any $(P,N)$-choice-condition $C$, we can always define an $\mathcal{S}$-compatible $\mathcal{S}$-raising-valuation $\pi$ by recursion over the well-founded ordering $P^+$.

[4](Which directions of the equivalences $\forall$-, $\exists$-, $\Rightarrow$-, and $\neg$-lemmas are needed where?)
Lemma 7.10 depends on the backward directions of the $\forall$-LEMMA and the $\Rightarrow$-LEMMA, and on the forward direction of the $\exists$-LEMMA. Lemma 7.11 and Theorem 8.5(6) depend on the forward directions of the $\forall$-LEMMA and the $\Rightarrow$-LEMMA, and on the backward direction of the $\exists$-LEMMA. Theorem 8.3 depends on both directions of the $\forall$-LEMMA, of the $\exists$-LEMMA, and of the $\neg$-LEMMA.

We can choose between two variants of $\pi$. For the one reading only the free atoms it must read, we set $R := {}_{\mathbb{A}}\!\upharpoonright(P^{+})$; for the one reading all free atoms it may read without violating the compatibility requirements, we set $R := (\mathbb{A}\times\mathbb{V}) \setminus (P^{*}\circ N)^{-1}$.

Moreover, $\pi$ can be defined even if its values on $\mathbb{V} \setminus \mathrm{dom}(C)$ are already fixed, say by an $\mathcal{S}$-raising-valuation $\rho$ with $S_\rho \subseteq R$.

**Theorem 7.5 (Existence of Two $\mathcal{S}$-Compatible $\mathcal{S}$-Raising-Valuations)**

*Let $C$ be a $(P, N)$-choice-condition. Let $\mathcal{S}$ be a $\Sigma$-structure. Assume that, for every choice type $\alpha$ of $C$, there is a generalized choice function on the power-set of $\mathcal{S}(\forall)_\alpha$.*

*Let the binary relation $R$ be defined according to one of the two alternatives mentioned above, i.e. either by $R := {}_{\mathbb{A}}\!\upharpoonright(P^{+})$, or else by $R := (\mathbb{A}\times\mathbb{V}) \setminus (P^{*}\circ N)^{-1}$.*

[*Let $\rho$ be an $\mathcal{S}$-raising-valuation with $S_\rho \subseteq R$.*]

*Then there is an $\mathcal{S}$-raising-valuation $\pi$ such that the following hold:*

- *$\pi$ is $\mathcal{S}$-compatible with $(C, (P, N))$.*
- *$S_\pi = R$.*

[• *For all $\tau : \mathbb{A} \to \mathcal{S}$ and all $y^{\mathbb{V}} \in \mathbb{V} \setminus \mathrm{dom}(C)$:* $\mathsf{e}(\pi)(\tau)(y^{\mathbb{V}}) = \mathsf{e}(\rho)(\tau)(y^{\mathbb{V}})$. ]

In § 5.5 we were able to state in Corollary 5.9 that consistency of an extension implies the consistency of the original positive/negative variable-condition. The analogous property for an extended extension, however, could not be stated in § 5.7 because it is a semantic property that requires Definition 7.4 (Compatibility) to be already given. So here comes this simple soundness property for extended extensions:

**Lemma 7.6 (Extended Extension)**
*Let $(C', (P', N'))$ be an extended extension of $(C, (P, N))$.*
*If $\pi$ is $\mathcal{S}$-compatible with $(C', (P', N'))$, then $\pi$ is $\mathcal{S}$-compatible with $(C, (P, N))$ as well.*

## 7.5 $(C, (P, N))$-Validity

**Definition 7.7** (**$(C, (P, N))$-Validity,** K)
Let $C$ be a $(P, N)$-choice-condition. Let $G$ be a set of sequents.
Let $\mathcal{S}$ be a $\Sigma$-structure. Let $\delta : \mathbb{VA} \rightsquigarrow \mathcal{S}$ be an $\mathcal{S}$-valuation.

$G$ is *$(C, (P, N))$-valid in $\mathcal{S}$* if
$G$ is $(\pi, \mathcal{S})$-valid for some $\pi$ that is $\mathcal{S}$-compatible with $(C, (P, N))$.

$G$ is *$(\pi, \mathcal{S})$-valid* if $G$ is true in $\mathcal{S} \uplus \mathsf{e}(\pi)(\tau) \uplus \tau$ for every $\tau : \mathbb{A} \to \mathcal{S}$.

$G$ is *true in $\mathcal{S}\uplus\delta$* if $\Gamma$ is true in $\mathcal{S}\uplus\delta$ for all $\Gamma \in G$.

A sequent $\Gamma$ is *true in $\mathcal{S}\uplus\delta$* if there is some formula listed in $\Gamma$ that is true in $\mathcal{S}\uplus\delta$.

Validity in a class of $\Sigma$-structures is understood as validity in each of the $\Sigma$-structures of that class. If we omit the reference to a special $\Sigma$-structure we mean validity in some fixed class K of $\Sigma$-structures, such as the class of all $\Sigma$-structures or the class of Herbrand $\Sigma$-structures.

Note that the quantification over $\pi$ in Definition 7.7 is an existential one. Such a definition of validity makes sense only if such a $\pi$ is always known to exist, simply because otherwise even the standard tautologies would not be valid. In our framework this existence is guaranteed by Theorem 7.5. The price we have to pay for this theorem is that — according to the definition of choice-conditions (Definition 5.12) — for any $(P, N)$-choice-condition $C$, for any $z^{\mathbb{V}} \in \mathrm{dom}(C)$ and $y^{\mathbb{VA}} \in \mathbb{VA}(C(z^{\mathbb{V}}))$, we have $y^{\mathbb{VA}} P^{+} z^{\mathbb{V}}$ in the well-founded ordering $P^{+}$, cf. Definition 5.12. Thus, as a corollary of Theorem 7.5 we get:

**Corollary 7.8 (Validity of Tautologies)**
*The set of the tautologies described in § 3.1 is $(C, (P, N))$-valid in $\mathcal{S}$ for any $(P, N)$-choice-condition $C$ and any $\Sigma$-structure $\mathcal{S}$, provided that for every choice type $\alpha$ of $C$, there is a generalized choice function on the power-set of $\mathcal{S}(\forall)_{\alpha}$.*

Under this assumption on the existence of generalized choice functions, we can now consider some very elementary examples on the basic ideas behind our notion of validity.

**Example 7.9 ($(\emptyset, (P, N))$-Validity)**

For $x^{\mathbb{V}} \in \mathbb{V}$, $y^{\mathbb{A}} \in \mathbb{A}$, the single-formula sequent $x^{\mathbb{V}} = y^{\mathbb{A}}$ is $(\emptyset, (\emptyset, \emptyset))$-valid in any $\Sigma$-structure $\mathcal{S}$ because we can choose $S_{\pi} := \mathbb{A} \times \mathbb{V}$ and $\pi(x^{\mathbb{V}})(\tau) := \tau(y^{\mathbb{A}})$ for $\tau : \mathbb{A} \to \mathcal{S}$, resulting in

$$\mathsf{e}(\pi)(\tau)(x^{\mathbb{V}}) = \pi(x^{\mathbb{V}})({}_{S_{\pi}\langle\{x^{\mathbb{V}}\}\rangle}{\upharpoonright}\tau) = \pi(x^{\mathbb{V}})({}_{\mathbb{A}}{\upharpoonright}\tau) = \pi(x^{\mathbb{V}})(\tau) = \tau(y^{\mathbb{A}}).$$

This means that $(\emptyset, (\emptyset, \emptyset))$-validity of $x^{\mathbb{V}} = y^{\mathbb{A}}$ is equivalent to validity of

$$\forall y_0^{\mathbb{B}}.\ \exists x_0^{\mathbb{B}}.\ (x_0^{\mathbb{B}} = y_0^{\mathbb{B}}). \tag{1}$$

Moreover, note that $\mathsf{e}(\pi)(\tau)$ has access to the $\tau$-value of $y^{\mathbb{A}}$ just as a raising function $x_1^{\mathbb{B}}$ for $x_0^{\mathbb{B}}$ has access to $y_0^{\mathbb{B}}$ in the raised (i.e. dually SKOLEMized) form $\exists x_1^{\mathbb{B}}.\ \forall y_0^{\mathbb{B}}.\ (x_1^{\mathbb{B}}(y_0^{\mathbb{B}}) = y_0^{\mathbb{B}})$ of (1).

Contrary to this, for $P := \emptyset$ and $N := \mathbb{V} \times \mathbb{A}$, the same single-formula sequent $x^{\mathbb{V}} = y^{\mathbb{A}}$ is not $(\emptyset, (P, N))$-valid in general, because then the required consistency of $(P \cup S_{\pi}, N)$ implies $S_{\pi} = \emptyset$; otherwise $P \cup S_{\pi} \cup N$ has a cycle of length 2 with exactly one edge from $N$. Thus, the value of $x^{\mathbb{V}}$ cannot depend on $\tau(y^{\mathbb{A}})$ anymore:

$$\pi(x^{\mathbb{V}})({}_{S_{\pi}\langle\{x^{\mathbb{V}}\}\rangle}{\upharpoonright}\tau) = \pi(x^{\mathbb{V}})({}_{\emptyset}{\upharpoonright}\tau) = \pi(x^{\mathbb{V}})(\emptyset).$$

This means that $(\emptyset, (\emptyset, \mathbb{V} \times \mathbb{A}))$-validity of $x^{\mathbb{V}} = y^{\mathbb{A}}$ is equivalent to validity of

$$\exists x_0^{\mathbb{B}}.\ \forall y_0^{\mathbb{B}}.\ (x_0^{\mathbb{B}} = y_0^{\mathbb{B}}). \tag{2}$$

Moreover, note that $\mathsf{e}(\pi)(\tau)$ has no access to the $\tau$-value of $y^{\mathbb{A}}$ just as a raising function $x_1^{\mathbb{B}}$ for $x_0^{\mathbb{B}}$ has no access to $y_0^{\mathbb{B}}$ in the raised form $\exists x_1^{\mathbb{B}}.\ \forall y_0^{\mathbb{B}}.\ (x_1^{\mathbb{B}}() = y_0^{\mathbb{B}})$ of (2).

For a more general example let $G = \{\ A_{i,0} \ldots A_{i,n_i - 1} \mid i \in I\ \}$, where, for $i \in I$ and $j \prec n_i$, the $A_{i,j}$ are formulas with variables from $\boldsymbol{v}$ and atoms from $\boldsymbol{a}$.
Then $(\emptyset, (\emptyset, \mathbb{V} \times \mathbb{A}))$-validity of $G$ means validity of $\exists \boldsymbol{v}.\ \forall \boldsymbol{a}.\ \forall i \in I.\ \exists j \prec n_i.\ A_{i,j}$
whereas $(\emptyset, (\emptyset, \emptyset))$-validity of $G$ means validity of $\forall \boldsymbol{a}.\ \exists \boldsymbol{v}.\ \forall i \in I.\ \exists j \prec n_i.\ A_{i,j}$

For a further example on validity, see Example 6.1, which treats HENKIN quantification and IF-logic quantifiers.

Ignoring the question of $\gamma$-multiplicity, also any other sequence of universal and existential quantifiers can be represented by a consistent positive/negative variable-condition, simply by starting from the consistent positive/negative variable-condition $(\emptyset,\emptyset)$ and applying the $\gamma$- and $\delta$-rules from §§ 3.4, 3.5, and 3.6. A reverse translation of a positive/negative variable-condition $(P,N)$ into a sequence of quantifiers, however, may require a strengthening of dependencies, in the sense that a subsequent backward translation would result in a *more restrictive* consistent positive/negative variable-condition $(P',N')$ with $P\subseteq P'$ and $N\subseteq N'$. This means that our framework can express quantificational dependencies more fine-grained than standard quantifiers; cf. Example 6.1.

## 7.6 Validity of Our Version $Q_C$ of Hilbert–Bernays' Axiom $(\varepsilon_0)$

As already explained in § 4.14, the single-formula sequent $Q_C(y^{\mathbb{V}})$ of Definition 4.12 is the formulation of axiom $(\varepsilon_0)$ of § 4.6 in our framework. In § 4.14, we already stated its validity, which we have formalized just now in Definition 7.7. It is now high time to show this validity:

**Lemma 7.10 ($(C,(P,N))$-Validity of $Q_C(y^{\mathbb{V}})$)**
*Let $C$ be a $(P,N)$-choice-condition. Let $y^{\mathbb{V}}\in\mathrm{dom}(C)$. Let $\mathcal{S}$ be a $\Sigma$-structure.*

1. *$Q_C(y^{\mathbb{V}})$ is $(\pi,\mathcal{S})$-valid for every $\pi$ that is $\mathcal{S}$-compatible with $(C,(P,N))$.*
2. *$Q_C(y^{\mathbb{V}})$ is $(C,(P,N))$-valid in $\mathcal{S}$; provided that for every choice type $\alpha$ of $C$ (cf. Definition 5.12), there is a generalized choice function on the power-set of $\mathcal{S}(\forall)_\alpha$.*

## 7.7 The Main Lemma on Substitution of Variables

Suppose we have a $(P,N)$-choice-condition $C$ and formula $B$ as our current state of the proof attempt. Let us apply a $(P,N)$-substitution $\sigma$ to this whole state. Then our new choice-condition $(C',(P',N'))$ will be the extended $\sigma$-update of $(C,(P,N))$. Now assume that, for some $\Sigma$-structure $\mathcal{S}$, we have found some $\mathcal{S}$-raising-valuation $\pi'$ which is $\mathcal{S}$-compatible with $(C',(P',N'))$ and for which the formula $B\sigma$ is $(\pi',\mathcal{S})$-valid. To show that the original proof attempt reduces to the new one, we have to construct an $\mathcal{S}$-raising-valuation $\pi$ which is $\mathcal{S}$-compatible with $(C,(P,N))$ and for which $B$ is $(\pi,\mathcal{S})$-valid. Critical for this construction is a set $O\subseteq\mathrm{dom}(\sigma)\cap\mathrm{dom}(C)$ for which our way of expressing the axiom $(\varepsilon_0)$, i.e. the set of single-formula sequents $(\langle O\rangle Q_C)\sigma$, has to be added to the set with the single-formula sequent $B\sigma$ of our new state. Luckily, the reverse of $O\subseteq\mathrm{dom}(\sigma)\cap\mathrm{dom}(C)$ is not required, provided that we find some set $O'\subseteq\mathrm{dom}(C)\setminus(O\cup\mathbb{V}(B))$ (i.e. of variables irrelevant for $B$) such that $\mathrm{dom}(\sigma)\cap\mathrm{dom}(C)\subseteq O'\uplus O$.

For easy look-up we represent the situation described here not with one of the popular VENN diagrams, but with a much clearer LAMBERT diagram [LAMBERT, 1764, Dianoiologie, §§ 173–194]:

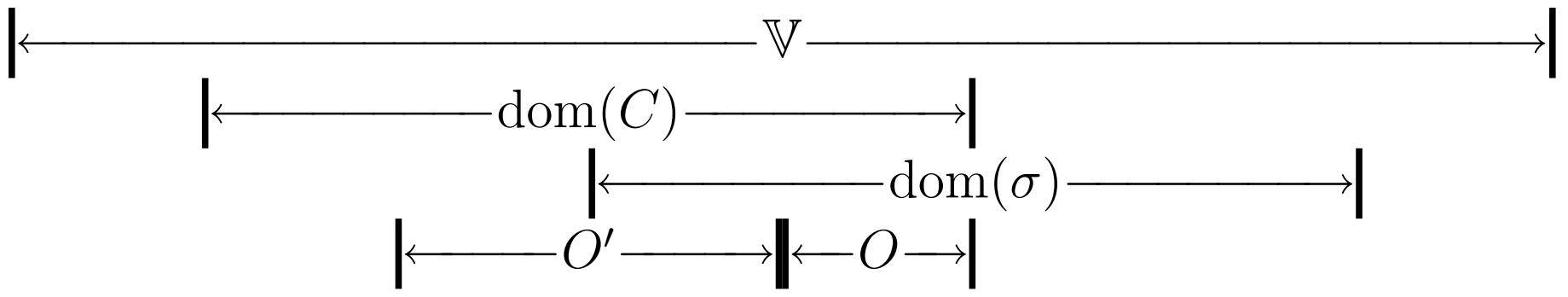

In general, a LAMBERT diagram expresses nothing but the following: If — in vertical projection — each point of the overlap of the lines for classes $A_1, \ldots, A_m$ on different levels is covered by a line for the classes $B_1, \ldots, B_n$ then $A_1 \cap \cdots \cap A_m \subseteq B_1 \cup \cdots \cup B_n$; moreover, on each level, the points not covered by a line for $A$ are considered to be covered by a line for the complement $\overline{A}$.

**Lemma 7.11 ($(P,N)$-Substitutions and $(C,(P,N))$-Validity)**
*Let $C$ be a $(P,N)$-choice-condition.*
*Let $\sigma$ be a $(P,N)$-substitution. Let $(C',(P',N'))$ be the extended $\sigma$-update of $(C,(P,N))$.*
*Let $\mathcal{S}$ be a $\Sigma$-structure. Let $\pi'$ be $\mathcal{S}$-compatible with $(C',(P',N'))$.*
*Let $O$ and $O'$ be two disjoint sets with $O \subseteq \mathrm{dom}(\sigma) \cap \mathrm{dom}(C)$ and $O' \subseteq \mathrm{dom}(C) \setminus O$.*
*Moreover, assume that $\sigma$ respects $C$ on $O$ in the given semantic context in the sense that $(\langle O\rangle Q_C)\sigma$ is $(\pi',\mathcal{S})$-valid. Furthermore, regarding the set $O'$ (where $\sigma$ may disrespect $C$), assume the following items to hold:*

- *$O'$ covers the variables in $\mathrm{dom}(\sigma) \cap \mathrm{dom}(C)$ besides $O$ in the sense of*
$$\mathrm{dom}(\sigma) \cap \mathrm{dom}(C) \;\subseteq\; O' \uplus O.$$
- *$O'$ satisfies the closure condition $\langle O'\rangle P^{+} \cap \mathrm{dom}(C) \subseteq O'$.*
- *For every $y^{\mathbb{V}} \in O'$, for $\alpha$ being the choice type of $C(y^{\mathbb{V}})$ (cf. Definition 5.12), there is a generalized choice function on the power-set of $\mathcal{S}(\forall)_{\alpha}$.*

*Then there is an $\mathcal{S}$-raising-valuation $\pi$ that is $\mathcal{S}$-compatible with $(C,(P,N))$ and satisfies the following:*

1. *For every term or formula $B$ with $O' \cap \mathbb{V}(B) = \emptyset$ and possibly with some unbound occurrences of bound atoms from a set $W \subseteq \mathbb{B}$, and for every $\tau : \mathbb{A} \rightarrow \mathcal{S}$ and every $\chi : W \rightarrow \mathcal{S}$:* $\quad \mathrm{eval}(\mathcal{S} \uplus \mathsf{e}(\pi')(\tau) \uplus \tau \uplus \chi)(B\sigma) \;=\; \mathrm{eval}(\mathcal{S} \uplus \mathsf{e}(\pi)(\tau) \uplus \tau \uplus \chi)(B)$.
2. *For every set of sequents $G$ with $O' \cap \mathbb{V}(G) = \emptyset$ we have:*
$$G\sigma \text{ is } (\pi',\mathcal{S})\text{-valid iff } G \text{ is } (\pi,\mathcal{S})\text{-valid.}$$

Note that Lemma 7.11 gets a lot simpler when we require the entire $(P,N)$-substitution $\sigma$ to respect the $(P,N)$-choice-condition $C$ by setting $O := \mathrm{dom}(\sigma) \cap \mathrm{dom}(C)$ and $O' := \emptyset$; in particular all requirements on $O'$ are trivially satisfied then. Moreover, note that the (still quite long) proof of Lemma 7.11 is more than a factor of 2 shorter than the proof of the analogous Lemma B.5 in [WIRTH, 2004] (together with Lemma B.1, its additionally required sub-lemma).

# 8 Reduction

Reduction is the reverse of consequence. It is the backbone of logical reasoning, especially of abduction and goal-directed deduction. In our case, a reduction step does not only reduce a set of problems describing a state to another set of problems, but also guarantees that the solutions of the latter also solve the former; here "solutions" means those $\mathcal{S}$-raising-valuations of the variables from $\mathbb{V}$ which are $\mathcal{S}$-compatible with $(C,(P,N))$ for the positive/negative variable-condition $(P,N)$ and the $(P,N)$-choice-condition $C$ given as the context of the states.

**Definition 8.1 (Reduction)**
Let $(P,N)$ be a positive/negative variable-condition. Let $C$ be a $(P,N)$-choice-condition. Let $G_0$ and $G_1$ be sets of sequents. Let $\mathcal{S}$ be a $\Sigma$-structure.
$G_0$ *$(C,(P,N))$-reduces to $G_1$ in $\mathcal{S}$* if for every $\pi$ that is $\mathcal{S}$-compatible with $(C,(P,N))$:

$$\text{If } G_1 \text{ is } (\pi,\mathcal{S})\text{-valid, then } G_0 \text{ is } (\pi,\mathcal{S})\text{-valid as well.}$$

The most obvious requirements on logical problem reduction are the following:

1. Validity: If the reduct of a goal is valid, so is the goal.
2. Reflexivity: Any superset of the problems of a goal is a reduct of this goal. This includes reflexivity of reduction by the case of a non-proper superset.
3. Transitivity: The reduct of the reduct of a goal is a reduct of the goal.
4. Additivity: The union of the reducts of two goals is a reduct of the union of these goals.

These requirements are satisfied by our notion of reduction, of course:

**Corollary 8.2 (Reduction)**
*Let $(P,N)$ be a positive/negative variable-condition. Let $C$ be a $(P,N)$-choice-condition. Let $G_0$, $G_1$, $G_2$, and $G_3$ be sets of sequents. Let $\mathcal{S}$ be a $\Sigma$-structure.*

**1. (Validity)** *If $G_0$ $(C,(P,N))$-reduces to $G_1$ in $\mathcal{S}$ and $G_1$ is $(C,(P,N))$-valid in $\mathcal{S}$, then $G_0$ is $(C,(P,N))$-valid in $\mathcal{S}$, too.*

**2. (Reflexivity)** *In case of $G_0 \subseteq G_1$: $G_0$ $(C,(P,N))$-reduces to $G_1$ in $\mathcal{S}$.*

**3. (Transitivity)** *If $G_0$ $(C,(P,N))$-reduces to $G_1$ in $\mathcal{S}$ and $G_1$ $(C,(P,N))$-reduces to $G_2$ in $\mathcal{S}$, then $G_0$ $(C,(P,N))$-reduces to $G_2$ in $\mathcal{S}$.*

**4. (Additivity)** *If $G_0$ $(C,(P,N))$-reduces to $G_2$ in $\mathcal{S}$ and $G_1$ $(C,(P,N))$-reduces to $G_3$ in $\mathcal{S}$, then $G_0 \cup G_1$ $(C,(P,N))$-reduces to $G_2 \cup G_3$ in $\mathcal{S}$.*

## 8.1 Mutual Reduction of $\alpha$-, $\beta$-, $\gamma$-, $\delta$-Rules

*Soundness* of inference rules has the global effect that if we reduce a set of sequents to an empty set, then we know that the original set is valid. Soundness is an essential property of inference rules.

*Safeness* of inference rules has the global effect that if we reduce a set of sequents to an invalid set, then we know that already the original set was invalid. Safeness is helpful in rejecting false assumptions and in patching failed proof attempts.

As explained before, for a reduction step in our framework, we are not contend with soundness: We want *solution-preservation* in the sense that an $\mathcal{S}$-raising-valuation $\pi$ that makes the set of sequents of the reduced proof state $(\pi, \mathcal{S})$-valid is guaranteed to do the same for the original input proposition, provided that $\pi$ is $\mathcal{S}$-compatible with $(C, (P, N))$ for the positive/negative variable-condition $(P, N)$ and the $(P, N)$-choice-condition $C$ given as the context of the states.

All our inference rules of § 3 have all of these three properties. This is obvious for the trivial $\alpha$- and $\beta$-rules. For the inference rules where this is not obvious, i.e. our $\gamma$- and $\delta^{-}$- and $\delta^{+}$-rules of §§ 3.4, 3.5, and 3.6, we state these properties in the following theorem.

**Theorem 8.3 (All $\gamma$- and $\delta$-rules are sound and safe (besides $\alpha$- and $\beta$-rules))**
*Let $(P, N)$ be a positive/negative variable-condition. Let $C$ be a $(P, N)$-choice-condition.*
*Let us consider any of the $\gamma$-, $\delta^{-}$-, and $\delta^{+}$-rules of §§ 3.4, 3.5, and 3.6.*
*Let $G_0$ and $G_1$ be the sets of the sequent above and of the sequents below the bar of that rule, respectively.*
*Let $C''$ be the set of the pair indicated to the upper right of the bar if there is any (which is the case only for the $\delta^{+}$-rules) or the empty set otherwise.*
*Let $V$ be the relation indicated to the lower right of the bar if there is any (which is the case only for the $\delta^{-}$- and $\delta^{+}$-rules) or the empty set otherwise.*
*Let us weaken the informal requirement "Let $x^{\mathbb{A}}$ be a fresh free atom" of the $\delta^{-}$-rules to its technical essence "$x^{\mathbb{A}} \in \mathbb{A} \setminus \big(\ \mathrm{dom}(P) \cup \mathbb{A}(\Gamma, A, \Pi)\ \big)$".*
*Let us weaken the informal statement "Let $x^{\mathbb{V}}$ be a fresh variable" of the $\delta^{+}$-rules to its technical essence "$x^{\mathbb{V}} \in \mathbb{V} \setminus \big(\ \mathrm{dom}(C \cup P \cup N) \cup \mathbb{V}(A)\ \big)$".*
*Let us set $C' := C \cup C''$, $P' := P \cup V{\restriction}_{\mathbb{V}}$, $N' := N \cup V{\restriction}_{\mathbb{A}}$.*
*Then $(C', (P', N'))$ is an extended extension of $(C, (P, N))$ (cf. Definition 5.14).*
*Moreover, for the considered inference rule, in every $\Sigma$-structure $\mathcal{S}$, $\mathcal{S}$-validity below and above the bar mutually imply each other (i.e. the rule is sound and save), even in the stronger form of solution-preservation in the sense that $G_0$ and $G_1$ mutually $(C', (P', N'))$-reduce to each other.*

## 8.2 From Safe Proof Steps to Lemma Application and Instantiation of Variables

We may have come along with the idea that deduction is drawing formulas out of a heap of valid formulas to sew them together into a new valid formula, and that reduction is taking a formula to pieces with the guarantee that an invalid formula produces at least one invalid formula again.

After Theorem 8.3 of § 8.1 — implying that all our $\alpha$-, $\beta$-, $\gamma$-, and $\delta$-rules describe logical equivalence transformations — it may now seem, however, that deduction and reduction do not differ for our rules as they are all both sound and safe. After all, these few rules, all by themselves, joined with the structural tautologies of § 3.1, are complete for classical first-order logic [with equality].

So how can *unsafe* steps occur in our proof trees? Well, to end up with the obligation to prove an invalid set of goals to which we have reduced an initially valid goal, there are the following two ways.

### 8.2.1 Application of Induction Hypotheses and Lemmas

The first is the application of induction hypotheses or lemmas. In such proof steps we may apply the root sequent of a proof tree to a leaf of some proof tree as an induction hypothesis or as a lemma. For soundness, the lemma-application relation among proof trees must be acyclic. In particular, we must not apply a lemma to its own proof tree. With induction hypotheses there is no such acyclicity requirement and an induction hypothesis will typically be applied in its own proof tree; soundness must be guaranteed for induction hypotheses, however, by a weight term with which each sequent is augmented and which generates an additional goal at the leaf to guarantee that the weight of the induction hypothesis is smaller in a well-founded ordering than the weight of the goal to which it is applied.

Technically, the standard way of applying the sequent $A_0 \ \ldots \ A_l$ as a lemma instantiated via a substitution $\nu$ on $\mathbb{A}$ — with $\mathbb{V}(A_0, \ldots, A_l) \times \mathrm{dom}(\nu) \subseteq N$ for our positive/negative variable-condition $(P, N)$ — to a leaf with the sequent $\Pi$ is to add to this leaf the child nodes with the sequents

$$\begin{array}{lllll} \Pi & \neg A_o\nu, & & & \\ \Pi & \neg A_1\nu & A_0\nu, & & \\ \vdots & \vdots & \vdots & \ddots & \\ \Pi & \neg A_l\nu & A_{l-1}\nu & \ldots & A_0\nu. \end{array}$$

As the former leaf sequent $\Pi$ is a sub-sequent of all the sequents of its child nodes, such a lemma application is safe. If it is unsound, then $\Pi$ must be false and all the child sequents true for some valuation and some structure, and then the sequent $A_0\nu \ \ldots \ A_l\nu$ must be false. Then, by Theorem 8.5(3) in the following § 8.3, the lemma $A_0 \ \ldots \ A_l$ must be false itself. Therefore, when the lemma has been shown, lemma application is sound.

The generation of lemmas is an essential element for structuring bigger proofs. If a lemma is generated to close an open goal, it will typically be stronger than what is required for actually closing that goal. This means that the open goal is generalized in the generation of the lemma, so that it will be applicable to many other open goals in the future as well. Moreover, lemmas should always capture the semantic knowledge about a reasoning domain in the strongest possible form, because then they inform the prover and its human user optimally about this domain. While generalization is a good practice in the formation of lemmas, it is a must for the generalization of induction hypotheses, because a stronger induction hypothesis provides a stronger tool in its own proof and because this additional strength is typically required for the induction proof to succeed.

An induction-hypothesis or lemma application step may be unsound if and only if it applies an invalid conjecture, typically resulting from "over-generalization", which is the main source for the obligation to prove an invalid conjecture for showing an initially valid goal. Along the above argumentation, however, induction-hypothesis and lemma application steps are always safe within their proof tree and never turn safe proof steps into unsafe ones.

### 8.2.2 Instantiation of Variables

The second possible source for the obligation to prove an invalid goal for showing an initially valid one is the invalidating instantiation of variables via a $(P,N)$-substitution $\sigma$ in the context of a $(P,N)$-choice-condition $C$. Let us set $M := \mathrm{dom}(\sigma)\cap\mathrm{dom}(C)$. Thus, $M$ is the set of those variables that are instantiated by $\sigma$ although the choice of their values is restricted by $C$.

In most practical cases, the possibility in the Theorem 8.5(2) of the following § 8.3 to reduce the set $M$ to a set $O$ is not required because we can cut down the domain of $\sigma$ to the actually occurring variables. Thus, let us take $V := \mathbb{V}$. Then we get $O = M$ and $O' = \emptyset$, and Theorem 8.5(2) simplifies to its following corollary.

**Corollary 8.4 (Instantiation of Variables)**
*Let $(P,N)$ be a positive/negative variable-condition. Let $C$ be a $(P,N)$-choice-condition. Let $G_0$ and $G_1$ be sets of sequents. Let $\mathcal{S}$ be a $\Sigma$-structure. Let $\sigma$ be a $(P,N)$-substitution. Set $M := \mathrm{dom}(\sigma)\cap\mathrm{dom}(C)$. Let $(C',(P',N'))$ be the extended $\sigma$-update of $(C,(P,N))$.*

*(a) If $G_0\sigma \cup (\langle M\rangle Q_C)\sigma$ is $(C',(P',N'))$-valid in $\mathcal{S}$, then $G_0$ is $(C,(P,N))$-valid in $\mathcal{S}$.*

*(b) If $G_0$ $(C,(P,N))$-reduces to $G_1$ in $\mathcal{S}$,*
*then $G_0\sigma$ $(C',(P',N'))$-reduces to $G_1\sigma \cup (\langle M\rangle Q_C)\sigma$ in $\mathcal{S}$.*

In case of $M = \emptyset$, item (b) of Corollary 8.4 further simplifies to

*If $G_0$ $(C,(P,N))$-reduces to $G_1$ in $\mathcal{S}$,*
*then $G_0\sigma$ $(C',(P',N'))$-reduces to $G_1\sigma$ in $\mathcal{S}$.*

Thus, in case of $M = \emptyset$, global application of $\sigma$ neither can turn any safe proof step in a proof tree into an unsafe one, nor a sound proof step into an unsound one. Therefore, if a sequent of a leaf with a safe branch up to the root is invalidated, so are all sequents of this branch, including the original input sequent at the root; thus, at least one of the variables occurring in the input sequent must be instantiated via $\sigma$, and thereby invalidate the valid original input sequent by providing a wrong witness for an existential property.

Now, let $(C',(P',N'))$ be the extended $\sigma$-update of $(C,(P,N))$. Moreover, be reminded that, for a variable $z^{\mathbb{V}} \in \mathrm{dom}(C)$, the single-formula sequent $Q_C(z^{\mathbb{V}})$ is our formulation of Hilbert–Bernays' axiom $(\varepsilon_0)$ in our framework, cf. Definition 4.12 in § 4.14. Moreover, up to some assumption related to the Axiom of Choice, by Lemma 5.17 in § 5.7 and Lemma 7.10 in § 7.5, $Q_C(z^{\mathbb{V}})$ is both $(C,(P,N))$-valid and $(C',(P',N'))$-valid in $\mathcal{S}$.

In the remaining case of $M \neq \emptyset$, the following terrifying situation may occur for $z^{\mathbb{V}} \in M$: $(Q_C(z^{\mathbb{V}}))\sigma$ is not $(C',(P',N'))$-valid in $\mathcal{S}$ anymore.

This means that $\sigma$ instantiated the variable $z^{\mathbb{V}}$ globally with a term that does not satisfy our choice-condition $C$ on which the safeness and soundness of all our proof steps up to now rely. Thus, by global application of $\sigma$, every proof tree in which $z^{\mathbb{V}}$ occurs might lose its soundness and safeness. For instance, in Example 4.13, the $\delta^+$-step becomes an unsound step after global application of $\sigma$.

Therefore, it is strictly necessary have $(Q_C(z^{\mathbb{V}}))\sigma$ as a root of a proof tree and register the application of this proof tree in our lemma-application relation as being applied to all proof trees in which the variable $z^{\mathbb{V}}$ occurs; in case that there is no proof tree with this root sequence yet, we will have to introduce a new tree for the open lemma $(Q_C(z^{\mathbb{V}}))\sigma$.

Finally, even if $(Q_C(z^{\mathbb{V}}))\sigma$ is valid, this does not mean that the chosen term $\sigma(z^{\mathbb{V}})$ results in a valid set of goals; instead the situation is just like in the case of $M{=}\emptyset$ above and an invalidated goal in a previously safe branch results again in an invalidation of the original input theorem. The substitution $\sigma'$ in Example 4.15 is an example for this case.

## 8.3 Monotonicity, Instantiation of Variables and Free Atoms

The following Theorem 8.5 will formalize that all the typical non-trivial properties of logical problem reduction (in addition to the trivial ones of Corollary 8.2) are given for our notion of reduction as well:

1. Monotonicity: Reduction is monotonic under extended extensions of choice-conditions.
2. Instantiation of variables: For any substitution $\sigma$ on $\mathbb{V}$, the $\sigma$-instantiation of the reduct of a goal united with the $Q_C$-formulas for satisfying the choice-condition $C$ on $\mathrm{dom}(\sigma)\cap\mathrm{dom}(C)$ is a reduct of the $\sigma$-instantiation of this goal, provided that we switch to the extended $\sigma$-update.
3. Instantiation of free atoms: If we instantiate the atoms of a lemma, then this instance reduces to the original lemma, provided that the variables of the lemma cannot depend on the instantiated atoms due to the negative second component of the current variable-condition.

**Theorem 8.5 (Reduction)**

*Let $(P,N)$ be a positive/negative variable-condition. Let $C$ be a $(P,N)$-choice-condition. Let $G_0$, $G_1$, $G_2$, and $G_3$ be sets of sequents. Let $\mathcal{S}$ be a $\Sigma$-structure.*

**1. (Monotonicity)** *For $(C',(P',N'))$ being an extended extension of $(C,(P,N))$:*

*(a) If $G_0$ is $(C',(P',N'))$-valid in $\mathcal{S}$, then $G_0$ is also $(C,(P,N))$-valid in $\mathcal{S}$.*

*(b) If $G_0$ $(C,(P,N))$-reduces to $G_1$ in $\mathcal{S}$, then $G_0$ also $(C',(P',N'))$-reduces to $G_1$ in $\mathcal{S}$.*

**2. (Instantiation of Variables)** *Let $\sigma$ be a $(P,N)$-substitution.*
*Let $(C',(P',N'))$ be the extended $\sigma$-update of $(C,(P,N))$.*
*Set $M := \mathrm{dom}(\sigma)\cap\mathrm{dom}(C)$. Choose some $V\subseteq\mathbb{V}$ with $\mathbb{V}(G_0,G_1)\subseteq V$.*
*Set $O := M\cap P^*\langle V\rangle$. Set $O' := \mathrm{dom}(C)\cap\langle M\backslash O\rangle P^*$.*
*Assume that for every $y^{\mathbb{V}}\in O'$, for $\alpha$ being the choice type of $C(y^{\mathbb{V}})$ (cf. Definition 5.12), there is a generalized choice function on the power-set of $\mathcal{S}(\forall)_\alpha$.*

*(a) If $G_0\sigma\cup(\langle O\rangle Q_C)\sigma$ is $(C',(P',N'))$-valid in $\mathcal{S}$, then $G_0$ is $(C,(P,N))$-valid in $\mathcal{S}$.*

*(b) If $G_0$ $(C,(P,N))$-reduces to $G_1$ in $\mathcal{S}$,*
*then $G_0\sigma$ $(C',(P',N'))$-reduces to $G_1\sigma\cup(\langle O\rangle Q_C)\sigma$ in $\mathcal{S}$.*

**3. (Instantiation of Free Atoms)** *Let $\nu$ be a substitution on $\mathbb{A}$.*
*If $\mathbb{V}(G_0)\times\mathrm{dom}(\nu)\subseteq N$, then $G_0\nu$ $(C,(P,N))$-reduces to $G_0$ in $\mathcal{S}$.*

## 8.4 Leisenring's Axiom (E2) becomes Valid

We mentioned the controversial *extensionality axiom* (E2) of AL(BERT) C. LEISENRING already in § 4.11 in connection with substitutability problems, which disappeared when we replaced all $\varepsilon$-terms with variables in § 4.14, and we will have to come back to this axiom again in § B.1.1, where we will discuss its problems and historical controversies.

In this § 8.4, however, we will argue that this axiom is actually a very straightforward and unproblematic one, which just states that the usual substitutability in case of logical equivalence also holds under the $\varepsilon$-binder. In fact, this axiom is problematic only in connection with LEISENRING's universal treatment of HILBERT's $\varepsilon$ on the one hand and with the misguiding historical idea to escape from misperceived problems with HILBERT's $\varepsilon$ by giving the mere syntax of the scopes of the $\varepsilon$-binders new semantics in logic on the other hand.

Evidence for the straightforwardness of this axiom is given already by the fact that it occurred long before [LEISENRING, 1969], though not under LEISENRING's new label (E2), but as (S7) in [BOURBAKI, 1939ff.] (where $\tau$ is written for the $\varepsilon$, which must not be confused with HILBERT's $\tau$-operator, cf. Note 2) and as (II,4) already in 1937 in [ACKERMANN, 1938]. Repeated reinvention of a labeled axiom strongly indicates that it is straightforward indeed.

We will now show that the axiom becomes valid in our framework. This is an even stronger indication for the straightforwardness of this axiom. The crucial difference of our framework is that we treat the semantics of HILBERT's $\varepsilon$ existentially; whereas LEISENRING treated it universally, taking validity as truth for all generalized choice functions on the universe.

To put

$$\forall x^{\mathbb{B}}.\ (A_0 \Leftrightarrow A_1) \qquad \Rightarrow \qquad \varepsilon x^{\mathbb{B}}.\ A_0 \ = \ \varepsilon x^{\mathbb{B}}.\ A_1 \tag{E2}$$

into our framework, we first have to put the $\varepsilon$-binders into our choice-condition. Thus, suppose we have some $(P, N)$-choice-condition $C$ and two $\varepsilon$-free formulas $A_0, A_1$ with $\mathbb{B}(A_0, A_1) \subseteq \{x^{\mathbb{B}}\}$. Then we get two different fresh variables $x_0^{\mathbb{V}}, x_1^{\mathbb{V}} \in \mathbb{V} \setminus \mathbb{V}(A_0, A_1, C, P, N)$ of the same type as $x^{\mathbb{B}}$. Then we set $P' := P \cup \bigcup_{i=0}^{1} (\mathbb{V}\mathbb{A}(A_i) \times \{x_i^{\mathbb{V}}\})$, $N' := N$, and $C' := C \cup \bigcup_{i=0}^{1} \{(x_i^{\mathbb{V}}, \varepsilon x^{\mathbb{B}}.\ A_i)\}$.

As the variables $x_0^{\mathbb{V}}$, $x_1^{\mathbb{V}}$ are globally fresh and not identical, the new pairs in the positive/negative variable-condition $(P', N')$ cannot be part of any cycle and $C'$ is a partial function on $\mathbb{V}$, and therefore $(P', N')$ is a consistent extension of the positive/negative variable-condition $(P, N)$ and $C'$ is a $(P', N')$-choice-condition and $(C', (P', N'))$ is an extended extension of $(C, (P, N))$.

These details defined, we can now state our version of LEISENRING's (E2) as the following $\varepsilon$-free formula:

$$\forall x^{\mathbb{B}}.\ (A_0 \Leftrightarrow A_1) \qquad \Rightarrow \qquad x_0^{\mathbb{V}} \ = \ x_1^{\mathbb{V}} \tag{E2$'$}$$

**Theorem 8.6 (Validity of (E2′))**
*Under the above settings, the (set of the single-formula sequent given by) the formula* (E2′) *is* $(C', (P', N'))$*-valid in all* $\Sigma$*-structures* $\mathcal{S}$*; provided that for every choice type* $\alpha$ *of* $C$*, there is a generalized choice function on the power-set of* $\mathcal{S}(\forall)_\alpha$.

The following proof is not put it into the appendix, because it introduces to high-level proof techniques with our theorems on reduction — based on the elegant and powerful introduction of new choice-conditions.

**High-Level Proof of Theorem 8.6**

An elegant way to prove Theorem 8.6 seems to be the application of Corollary 8.4(a) to the $(P', N')$-substitution $\{x^{\mathbb{V}}_1 \mapsto x^{\mathbb{V}}_0\}$, because then the conclusion of the instantiated formula $(\mathrm{E2}')\{x^{\mathbb{V}}_1 \mapsto x^{\mathbb{V}}_0\}$ is the tautology $x^{\mathbb{V}}_0 = x^{\mathbb{V}}_0$.

As required in Corollary 8.4(a), we can set $(C'', (P'', N''))$ to the extended $\{x^{\mathbb{V}}_1 \mapsto x^{\mathbb{V}}_0\}$-update of $(C', (P', N'))$ and $M := \mathrm{dom}(\{x^{\mathbb{V}}_1 \mapsto x^{\mathbb{V}}_0\}) \cap \mathrm{dom}(C') = \{x^{\mathbb{V}}_1\}$. Now all we have left to show is the $(C'', (P'', N''))$-validity of $(Q_{C'}(x^{\mathbb{V}}_1))\{x^{\mathbb{V}}_1 \mapsto x^{\mathbb{V}}_0\}$, i.e., by Definition 4.12, of $(\exists x^{\mathbb{B}}.\, A_1 \Rightarrow A_1\{x^{\mathbb{B}} \mapsto x^{\mathbb{V}}_1\})\{x^{\mathbb{V}}_1 \mapsto x^{\mathbb{V}}_0\}$, i.e. of $\exists x^{\mathbb{B}}.\, A_1 \Rightarrow A_1\{x^{\mathbb{B}} \mapsto x^{\mathbb{V}}_0\}$. But we can show this only if the condition of (E2′) holds, which we cannot even assume to be valid.

As there is nothing like a conditional application of the substitution $\{x^{\mathbb{V}}_1 \mapsto x^{\mathbb{V}}_0\}$, all we can do to patch this failed proof attempt is to use a substitution $\{x^{\mathbb{V}}_1 \mapsto y^{\mathbb{V}}\}$ and a conditional choice-condition for $y^{\mathbb{V}}$. Thus, for another fresh variable $y^{\mathbb{V}} \in \mathbb{V} \setminus \mathbb{V}(A_0, A_1, C', P', N', x^{\mathbb{V}}_0, x^{\mathbb{V}}_1)$ of the same type as $x^{\mathbb{B}}$, let us first redefine $P'' := P' \cup \mathbb{VA}(A_0, A_1, x^{\mathbb{V}}_0) \times \{y^{\mathbb{V}}\}$, $N'' := N$, and

$$C'' := C' \cup \left\{ \left( y^{\mathbb{V}},\ \varepsilon y^{\mathbb{B}}.\left( \begin{array}{ll} & (\ \ \forall x^{\mathbb{B}}.\ (A_0 \Leftrightarrow A_1) \ \Rightarrow\ x^{\mathbb{V}}_0 = y^{\mathbb{B}}\ ) \\ \wedge & (\ \neg\forall x^{\mathbb{B}}.\ (A_0 \Leftrightarrow A_1) \ \Rightarrow\ A_1\{x^{\mathbb{B}} \mapsto y^{\mathbb{B}}\}\ ) \end{array} \right) \right) \right\},$$

Now $(P'', N'')$ is a consistent extension of $(P', N')$, $\{x^{\mathbb{V}}_1 \mapsto y^{\mathbb{V}}\}$ is a $(P'', N'')$-substitution, $C''$ is a $(P'', N'')$-choice-condition and $(C'', (P'', N''))$ is an extended extension of $(C', (P', N'))$. By Theorem 8.5(1a) it suffices to show that (E2′) is $(C'', (P'', N''))$-valid in all $\Sigma$-structures $\mathcal{S}$.

As required in Corollary 8.4(a), we can now set $(C''', (P''', N'''))$ to the $\{x^{\mathbb{V}}_1 \mapsto y^{\mathbb{V}}\}$-update of $(C'', (P'', N''))$ and $M := \mathrm{dom}(\{x^{\mathbb{V}}_1 \mapsto y^{\mathbb{V}}\}) \cap \mathrm{dom}(C'') = \{x^{\mathbb{V}}_1\}$. Now again, by Corollary 8.4(a), it suffices to show the $(C''', (P''', N'''))$-validity of two formulas:

(1) $(\mathrm{E2}')\{x^{\mathbb{V}}_1 \mapsto y^{\mathbb{V}}\}$, i.e. $\forall x^{\mathbb{B}}.\ (A_0 \Leftrightarrow A_1) \Rightarrow x^{\mathbb{V}}_0 = y^{\mathbb{V}}$.

(2) $(Q_{C''}(x^{\mathbb{V}}_1))\{x^{\mathbb{V}}_1 \mapsto y^{\mathbb{V}}\}$, i.e. of $(\exists x^{\mathbb{B}}.\, A_1 \Rightarrow A_1\{x^{\mathbb{B}} \mapsto x^{\mathbb{V}}_1\})\{x^{\mathbb{V}}_1 \mapsto y^{\mathbb{V}}\}$, i.e. of $\exists x^{\mathbb{B}}.\, A_1 \Rightarrow A_1\{x^{\mathbb{B}} \mapsto y^{\mathbb{V}}\}$.

Now a simple case analysis on $\forall x^{\mathbb{B}}.\ (A_0 \Leftrightarrow A_1)$ shows that the set of these two single-formula sequents $(C''', (P''', N'''))$-reduces to the set of the following two single-formula sequents:

$$\exists y^{\mathbb{B}}.\left( \begin{array}{ll} & (\ \ \forall x^{\mathbb{B}}.(A_0 \Leftrightarrow A_1) \Rightarrow x^{\mathbb{V}}_0 = y^{\mathbb{B}}) \\ \wedge & (\neg\forall x^{\mathbb{B}}.(A_0 \Leftrightarrow A_1) \Rightarrow A_1\{x^{\mathbb{B}} \mapsto y^{\mathbb{B}}\}) \end{array} \right) \Rightarrow \left( \begin{array}{ll} & (\ \ \forall x^{\mathbb{B}}.(A_0 \Leftrightarrow A_1) \Rightarrow x^{\mathbb{V}}_0 = y^{\mathbb{V}}) \\ \wedge & (\neg\forall x^{\mathbb{B}}.(A_0 \Leftrightarrow A_1) \Rightarrow A_1\{x^{\mathbb{B}} \mapsto y^{\mathbb{V}}\}) \end{array} \right)$$

and $\exists x^{\mathbb{B}}.\ A_0 \ \Rightarrow\ A_0\{x^{\mathbb{B}} \mapsto x^{\mathbb{V}}_0\}$.

Indeed, in case of $\forall x^{\mathbb{B}}.\ (A_0 \Leftrightarrow A_1)$, because $\exists y^{\mathbb{B}}.\ (x^{\mathbb{V}}_0 = y^{\mathbb{B}})$ is true, the former of the two latter formulas as well as formula (1) simplify to $x^{\mathbb{V}}_0 = y^{\mathbb{V}}$; and given this equation and the case assumption, also the latter becomes logically equivalent to formula (2), because they can be rewritten into each other. In the complementary case, however, the formula (1) becomes true; moreover, the former of the two latter formulas simplifies to $\exists y^{\mathbb{B}}.\, A_1\{x^{\mathbb{B}} \mapsto y^{\mathbb{B}}\} \Rightarrow A_1\{x^{\mathbb{B}} \mapsto y^{\mathbb{V}}\}$, which is formula (2) with $x^{\mathbb{B}}$ under $\exists$ renamed to $y^{\mathbb{B}}$.

Notice that this was the case where the first proof attempt failed, because we had to show $\exists x^{\mathbb{B}}.\, A_1 \Rightarrow A_1\{x^{\mathbb{B}} \mapsto x^{\mathbb{V}}_0\}$ , although we can neither rewrite $A_1$ to $A_0$ nor $x^{\mathbb{V}}_0$ to $x^{\mathbb{V}}_1$ in this case. $y^{\mathbb{V}}$ instead of $x^{\mathbb{V}}_0$ in this formula, however, just follows the choice-condition of $x^{\mathbb{V}}_1$.

The latter two formulas, however, are nothing but the formulas $Q_{C'''}(y^{\mathbb{V}})$ and $Q_{C'''}(x^{\mathbb{V}}_0)$, respectively, which are $(C''', (P''', N'''))$-valid by Lemma 7.10; provided that for every choice type $\alpha$ of $C$ (cf. Definition 5.12), there is a generalized choice function on the power-set of $\mathcal{S}(\forall)_\alpha$. Thus, under the same provision of Theorem 8.6, formulas (1) and (2) are $(C''', (P''', N'''))$-valid by Corollary 8.2(1) as well. **Q.e.d. (Theorem 8.6)**

# 9 Summary and Discussion

## 9.1 Positive/Negative Variable-Conditions

We take a *sequent* to be a list of formulas which denotes the disjunction of these formulas. We admit explicit quantification to bind only *bound atoms* (written $x^{\mathbb{B}}$). In addition to standard frameworks of two-valued logics, our formulas may contain variables and free atoms, which are *implicitly quantified* according to a context-independent semantics: Our *variables* (written $x^{\mathbb{V}}$) are quantified *existentially*, our *free atoms* (written $x^{\mathbb{A}}$) *universally*. The structure of this implicit form of quantification without quantifiers and without binders is represented globally in a *positive/negative variable-condition* $(P, N)$, which can be seen as a directed graph on variables and free atoms whose edges are elements of either $P$ or $N$.

Without loss of generality in practice, let us assume that $P$ is finite. Then, a positive/negative variable-condition $(P, N)$ is *consistent* if each cycle of its directed graph has more than one edge from $N$.

Roughly speaking, on the one hand, a *variable* $y^{\mathbb{V}}$ is put into the scope of another variable or free atom $x^{\mathbb{VA}}$ by an edge $(x^{\mathbb{VA}}, y^{\mathbb{V}})$ in $P$; and, on the other hand, a *free atom* $x^{\mathbb{A}}$ is put into the scope of a variable $y^{\mathbb{V}}$ by an edge $(y^{\mathbb{V}}, x^{\mathbb{A}})$ in $N$.

On the one hand, an edge $(x^{\mathbb{VA}}, y^{\mathbb{V}})$ *must* be put into $P$

- if $y^{\mathbb{V}}$ is introduced in a $\delta^{+}$-step where $x^{\mathbb{VA}}$ occurs in the principal formula (cf. § 3), and also
- if $y^{\mathbb{V}}$ is globally replaced with a term in which $x^{\mathbb{VA}}$ occurs.

On the other hand, an edge $(y^{\mathbb{V}}, x^{\mathbb{A}})$ *must* be put into $N$ if $x^{\mathbb{A}}$ is introduced in a $\delta^{-}$-step where $y^{\mathbb{V}}$ occurs in the sequent (i.e. in the principal or parametric formulas, cf. § 3).

Furthermore, such edges *may* always be added to the positive/negative variable-condition, as long as it remains consistent. Such an unforced addition of edges might be appropriate especially in the formulation of a new proposition:

- partly, because we may need this for modeling the intended semantics by representing the intended quantificational structure for the variables and free atoms of the new proposition;
- partly, because we may need this for enabling induction in the form of Fermat's *descente infinie* on the free atoms of the proposition; cf. [Wirth, 2004, §§ 2.5.2 and 3.3]. (This is closely related to the satisfaction of the condition on $N$ in Theorem 8.5(3).)

## 9.2 Semantics of Positive/Negative Variable-Conditions

The value assigned to a variable $y^{\mathbb{V}}$ by an *$\mathcal{S}$-raising-valuation* $\pi$ may depend on the value assigned to an atom $x^{\mathbb{A}}$ by an *$\mathcal{S}$-valuation*. In that case, the *semantic relation* $S_{\pi}$ contains an edge $(x^{\mathbb{A}}, y^{\mathbb{V}})$. Moreover, $\pi$ is enforced to obey the quantificational structure by the requirement that $(P \cup S_{\pi}, N)$ must be consistent; cf. Definitions 7.1 and 7.4.

## 9.3 Replacing $\varepsilon$-Terms with Variables

Suppose that an $\varepsilon$-term $\varepsilon z^{\mathbb{B}}.\ B$ has free occurrences of exactly the bound atoms $v_0^{\mathbb{B}}, \ldots, v_{l-1}^{\mathbb{B}}$ which are not free atoms of our framework, but are actually bound in the syntactic context in which this $\varepsilon$-term occurs. Then we can replace it in this context with the application term $z^{\mathbb{V}}(v_0^{\mathbb{B}}, \ldots, v_{l-1}^{\mathbb{B}})$ for a fresh variable $z^{\mathbb{V}}$ and set the value of a global function $C$ (called the *choice-condition*) at $z^{\mathbb{V}}$ according to

$$C(z^{\mathbb{V}}) \quad := \quad \lambda v_0^{\mathbb{B}}, \ldots, v_{l-1}^{\mathbb{B}}.\ \varepsilon z^{\mathbb{B}}.\ B,$$

and augment $P$ with an edge $(y^{\mathbb{VA}}, z^{\mathbb{V}})$ for each variable or free atom $y^{\mathbb{VA}}$ occurring in $B$.

## 9.4 Semantics of Choice-Conditions

A variable $z^{\mathbb{V}}$ in the domain of the global choice-condition $C$ must take a value that makes $C(z^{\mathbb{V}})$ true — if such a choice is possible. This can be formalized as follows. Let "eval" be the standard evaluation function. Let $\mathcal{S}$ be any of the semantic structures (or models) under consideration. Let $\delta$ be a valuation of the variables and free atoms (resulting from an $\mathcal{S}$-raising-valuation of the variables and an $\mathcal{S}$-valuation of the atoms). Let $\chi$ be an arbitrary $\mathcal{S}$-valuation of the bound atoms $v_0^{\mathbb{B}}, \ldots, v_{l-1}^{\mathbb{B}}, z^{\mathbb{B}}$. Then $\delta(z^{\mathbb{V}})$ must be a function that chooses a value that makes $B$ true whenever possible, in the sense that $\mathrm{eval}(\mathcal{S}{\uplus}\delta{\uplus}\chi)(B) = \mathsf{TRUE}$ implies $\mathrm{eval}(\mathcal{S}{\uplus}\delta{\uplus}\chi)(B\mu) = \mathsf{TRUE}$ for

$$\mu := \{z^{\mathbb{B}} \mapsto z^{\mathbb{V}}(v_0^{\mathbb{B}}, \ldots, v_{l-1}^{\mathbb{B}})\}.$$

## 9.5 Substitution of Variables ("$\varepsilon$-Substitution")

The kind of logical inference we essentially need is (problem-) *reduction,* the backbone of abduction and goal-directed deduction; cf. § 8. In a tree of reduction steps our variables and free atoms show the following behavior with respect to their instantiation:

Atoms behave as constant parameters. A variable $y^{\mathbb{V}}$, however, may be globally instantiated with any term by application of a substitution $\sigma$; unless, of course, in case $y^{\mathbb{V}}$ is in the domain of the global choice-condition $C$, in which case $\sigma$ must additionally satisfy $C(y^{\mathbb{V}})$, in a sense to be explained below.

In addition, the applied substitution $\sigma$ must always be an $(P, N)$-*substitution.* This means that the current positive/negative variable-condition $(P, N)$ remains consistent when we extend it to its so-called $\sigma$-*update*, which augments $P$ with the edges from the variables and free atoms in $\sigma(z^{\mathbb{V}})$ to $z^{\mathbb{V}}$, for each variable $z^{\mathbb{V}}$ in the domain $\mathrm{dom}(\sigma)$.

Moreover, the global choice-condition $C$ must be updated by removing $z^{\mathbb{V}}$ from its domain $\mathrm{dom}(C)$ and by applying $\sigma$ to the $C$-values of the variables remaining in $\mathrm{dom}(C)$.

Now, in case of a variable $z^{\mathbb{V}} \in \mathrm{dom}(\sigma) \cap \mathrm{dom}(C)$, $\sigma$ satisfies the current choice-condition $C$ if $(Q_C(z^{\mathbb{V}}))\sigma$ is valid in the context of the updated variable-condition and choice-condition. Here, for a choice-condition $C(z^{\mathbb{V}})$ and substitution $\mu$ given as above, $Q_C(z^{\mathbb{V}})$ denotes the formula

$$\forall v_0^{\mathbb{B}}.\ \ldots \forall v_{l-1}^{\mathbb{B}}.\ \big(\ \exists z^{\mathbb{B}}.\ B \ \Rightarrow\ B\mu\ \big),$$

which is nothing but our version of HILBERT's axiom $(\varepsilon_0)$; cf. Definition 4.12. Under these conditions, the invariance of reduction under substitution is stated in Corollary 8.4(b).

Finally, note that $Q_C(z^{\mathbb{V}})$ itself is always valid in our framework; cf. Lemma 7.10.

## 9.6 Where Have All the $\varepsilon$-Terms Gone?

After the replacement described in § 9.3 and, in more detail, in § 4.11, the $\varepsilon$-symbol occurs neither in our terms, nor in our formulas, but only in the range of the current choice-condition, where its occurrences are inessential, as explained at the end of § 4.11.

As a consequence of this removal, our formulas are much more readable than in the standard approach of in-line presentation of $\varepsilon$-terms, which always was nothing but a *theoretical* presentation because in practical proofs the $\varepsilon$-terms would have grown so large that the mere size of them made them inaccessible to human inspection. To see this, compare our presentation in Example 4.11 to the one in Example 4.6, and note that the latter is still hard to read although we have invested some efforts in finding a readable form of presentation.

From a mathematical point of view, however, the original $\varepsilon$-terms are still present in our approach; up to isomorphism and with the exception of some irrelevant term sharing. To make these $\varepsilon$-terms explicit in a formula $A$ for a given $(P, N)$-choice-condition $C$, do:

**Step 1:** Let us consider the relation $C$ not as a function, but as a ground term rewriting system: This means that we read $\left( z^{\mathbb{V}},\ \ \lambda v_0^{\mathbb{B}}, \ldots, v_{l-1}^{\mathbb{B}}.\ \varepsilon z^{\mathbb{B}}.\ B \right) \in C$ as a rewrite rule saying that we may replace the variable $z^{\mathbb{V}}$ (the left-hand side of the rule, which is not a variable but a constant w.r.t. the rewriting system) with the right-hand side $\lambda v_0^{\mathbb{B}}, \ldots, v_{l-1}^{\mathbb{B}}.\ \varepsilon z^{\mathbb{B}}.\ B$ in any given context.

By Definition 5.12(3), we know that all variables in $B$ are smaller than $z^{\mathbb{V}}$ in $P^+$. By the consistency of our positive/negative variable-condition $(P, N)$ (according to Definition 5.12), we know that $P^+$ is a well-founded ordering. Thus its multi-set extension is a well-founded ordering as well. Moreover, the multi-set of the variable $z^{\mathbb{V}}$ of the left-hand side is bigger than the multi-set of the occurrences of variables in the right-hand side in the well-founded multi-set extension of $P^+$. Thus, if we rewrite a formula, the multi-set of the occurrences of variables in the rewritten formula is smaller than the multi-set of the occurrences of variables in the original formula. Therefore, normalization of any formula $A$ with these rewrite rules terminates with a formula $A'$.

**Step 2:** As typed $\lambda\alpha\beta$-reduction is also terminating, we can apply it to remove the $\lambda$-terms introduced to $A'$ by the rewriting of Step 1, resulting in a formula $A''$.

Then — with the proper semantics for the $\varepsilon$-binder — the formulas $A'$ and $A''$ are equivalent to $A$, but do not contain any variables that are in the domain of $C$. This means that $A''$ is equivalent to $A$, but does not contain $\varepsilon$-constrained variables anymore. Moreover, if the variables in $A$ resulted from the elimination of $\varepsilon$-terms as described in §§ 4.11 and 9.3, then all $\lambda$-terms that were not already present in $A$ are provided with arguments and are removed by the rewriting of Step 2. Therefore, no $\lambda$-symbol occurs in the formula $A''$ if the formula $A$ resulted from a first-order formula: If we normalize $\mathsf{P}(w_a^{\mathbb{V}}, x_b^{\mathbb{V}}, y_d^{\mathbb{V}}, z_h^{\mathbb{V}})$ with respect to the rewriting system given by the $(P, N)$-choice-condition $C$ of of Example 4.11, and then by $\lambda\alpha\beta$-reduction, we end up in a normal form which is the first-order formula $\mathsf{P}(w_a, x_b, y_d, z_h)$ of Example 4.6, with the exception of the renaming of some bound atoms that are bound by $\varepsilon$. If each element $z^{\mathbb{V}}$ in the domain of $C$ binds a unique bound atom $z^{\mathbb{B}}$ by the $\varepsilon$ in the $\varepsilon$-term $C(z^{\mathbb{V}})$, then the normal form $A''$ can even preserve our information on committed choice when we consider any $\varepsilon$-term binding an occurrence of a bound atom of the same name to be committed to the same choice. In this sense, the representation given by the normal form is equivalent to our original one given by $\mathsf{P}(w_a^{\mathbb{V}}, x_b^{\mathbb{V}}, y_d^{\mathbb{V}}, z_h^{\mathbb{V}})$ and $C$.

## 9.7 Are We Breaking with Traditional Treatment of Hilbert's $\varepsilon$?

Our new semantic free-variable framework was originally developed to meet the requirements analysis for the combination of mathematical induction in the liberal style of FERMAT's *descente infinie* with state-of-the-art logical deduction. The framework provides a formal system in which working mathematicians can straightforwardly develop their proofs supported by powerful automation; cf. [WIRTH, 2004].

If traditionalism meant restriction to the expressional means of the past — say the first half of the 20th century with its foundational crisis and specific emphasis on constructivism, intuitionism, and finitism — then our approach would not classify as traditional and GRIGORI MINTS would have been right to blame our framework as anti-traditional. Although we offer the extras of non-committed choice and a model-theoretic notion of validity, we nevertheless see our framework based on $Q_C$ as a form of $(\varepsilon_0)$ (cf. § 4.14) as a backward compatible extension of HILBERT–BERNAYS' original framework with $(\varepsilon_0)$ as the only axiom for the $\varepsilon$. And with its equivalents for the traditional $\varepsilon$-terms (cf. § 9.6) and with its support for the global proof transformation given by the $\varepsilon$-substitution methods (cf. §§ 4.14, 8, and 9.5), our framework is indeed deeply rooted in the HILBERT–BERNAYS tradition.

Note that the fear of inconsistency should have been soothed anyway in the meantime by WITTGENSTEIN, cf. e.g. [DIAMOND, 1976]. The main disadvantage of an exclusively axiomatic framework as compared to one that also offers a model-theoretic semantics is the following: Constructive proofs of practically relevant theorems easily become too huge and too tedious, whereas semantic proofs are smaller and easier to handle. More important is the possibility to invent *new and more suitable logics for new applications* with semantic means, whereas proof transformations can refer only to already existing logics (cf. § 4.8).

We intend to pass the heritage of HILBERT's $\varepsilon$ on to new generations interested in computational linguistics, automated theorem proving, and mathematics assistance systems; fields in which — with very few exceptions — the overall common opinion still is (the wrong one) that the $\varepsilon$ hardly can be of any practical benefit.

The differences, however, between our free-variable framework for the $\varepsilon$ and HILBERT's original underspecified $\varepsilon$-operator, in the order of increasing importance, are the following:

1. The term-sharing of $\varepsilon$-terms with the help of variables improves the readability of our formulas considerably.
2. We do not have the requirement of globally committed choice for any $\varepsilon$-term: Different variables with the same choice-condition may take different values. Nevertheless, *$\varepsilon$-substitution* works at least as well as in the original framework.
3. Opposed to all other classical validities for the $\varepsilon$ (including the semantics of [ASSER, 1957], [HERMES, 1965], and [LEISENRING, 1969]), the implicit quantification over the choice of our variables is existential instead of universal. This change simplifies formal reasoning in all relevant contexts, because we have to consider only an arbitrary single solution (or choice, substitution) instead of checking all of them.

# 10 Conclusion

Our more flexible semantics for HILBERT's $\varepsilon$ and our choice-conditions presented in this paper were originally developed to combine mathematical induction in the liberal style of FERMAT's *descente infinie* with state-of-the-art deduction, where the well-foundedness required for the soundness of *descente infinie* enforced a notion of reduction with preservation of solutions, which the liberalized $\delta$-rules as found in [FITTING, 1996] cannot satisfy without something like our variables with choice-conditions, cf. [WIRTH, 2004].

Thus, by providing soundness to the first formal combination of goal-directed deduction and *descente infinie*, our choice-conditions had passed an evaluation of their usefulness even before they were recognized as a candidate for the semantics that HILBERT's school in logic may have had in mind for their $\varepsilon$. While this will remain speculation, the semantic framework for HILBERT's $\varepsilon$ proposed in this paper definitely has the following advantages:

**Indication of Commitment:** The requirement of a commitment to a choice is expressed syntactically and most clearly by the sharing of a variable; cf. § 4.11.

**Semantics:** The semantics of the $\varepsilon$ is simple and straightforward in the sense that the $\varepsilon$-operator becomes similar to the referential use of indefinite articles and determiners in natural languages, cf. [WIRTH, 2006b].

Our semantics for the $\varepsilon$ is based on an abstract formal approach extending a semantics for closed formulas (satisfying only very weak requirements, cf. § 7.1) to a semantics with existentially quantified *variables* and universally quantified *free atoms* replacing the three kinds of free variables of [WIRTH, 2004; 2006a; 2008; 2012b; 2006b], i.e. existential (free $\gamma$-variables), universal (free $\delta^-$-variables), and $\varepsilon$-constrained (free $\delta^+$-variables). The simplification achieved by the reduction from three to two kinds of free variables results in a remarkable reduction of the complexity of our framework and will make its adaptation to applications much easier.

In spite of this simplification, we have enhanced the expressiveness of our framework by replacing the variable-conditions of [WIRTH, 2002; 2004; 2006a; 2008; 2012b; 2006b] with our *positive/negative* variable-conditions here, such that our framework now admits us to represent HENKIN quantification directly; cf. Example 6.1. From a philosophical point of view, this clearer differentiation also provides a deep insight into the true nature and the relation of the $\delta^-$- and the $\delta^+$-rules.

**Reasoning:** Our representation of an $\varepsilon$-term $\varepsilon x^{\mathbb{B}}.\, A$ can be replaced with *any* term $t$ that satisfies $\exists x^{\mathbb{B}}.\, A \;\Rightarrow\; A\{x^{\mathbb{B}}{\mapsto}t\}$, cf. § 4.14. Thus, the correctness of such a replacement is likely to be expressible and verifiable in the original calculus.

Our framework for the $\varepsilon$ is especially convenient for developing proofs in the style of a working mathematician (cf. [WIRTH, 2004; 2006a; 2012b]) and makes proof work most simple because we do not have to consider all proper choices $t$ for $x$ (as in all other model-theoretic approaches), but only a single arbitrary one, fixed in a global proof transformation step of *$\varepsilon$-substitution*.

**Adaptability:** We hope that our new semantic framework will help to solve further practical and theoretical problems with the $\varepsilon$ and improve the applicability of the $\varepsilon$ as a logic tool for description and reasoning.

And already without the $\varepsilon$ (i.e. for the case that the choice-condition is empty, cf. e.g. [Wirth, 2012a; 2014]), our framework with its variables and free atoms of very high quantificational expressiveness (even without any quantifiers!) should find a multitude of applications in all areas of computer-supported reasoning.

Finally, a tailoring of operators similar to our $\varepsilon$ — to meet the special demands of specification and computation in various areas — is promising, in particular for describing semantics of discourses in natural language, cf. [Wirth, 2006b, §§ 5.8 and 6].

**Acknowledgments:** I would like to thank Murdoch J. Gabbay for inspiring and encouraging me to write this paper, for reviewing it, and for much more.

# A Are Liberalized $\delta$-Rules Always More Liberal?

We could object with the following two points to the classification in § 3.6, stating that the $\delta^+$-rules are more "liberal" than the $\delta^-$-rules, in the sense that they provide more freedom to the prover or that they admit proofs that are shorter or easier to find:

1. $\mathbb{V}\mathbb{A}(\forall x^{\mathbb{B}}.\ A)$ is not necessarily a subset of $\mathbb{V}(\varGamma\ \ \forall x^{\mathbb{B}}.\ A\ \ \varPi)$, because $\mathbb{V}\mathbb{A}(\forall x^{\mathbb{B}}.\ A)$ may include some additional free atoms.

2. Regarding our positive/negative variable-condition $(P, N)$, the $\delta^+$-rule may contribute a $P$-edge to a cycle with exactly one edge from $N$, whereas the analogous $\delta^-$-rule would contribute an $N$-edge instead. Thus, the analogous cycle would then not count as counterexample to the consistency of the positive/negative variable-condition because it has two edges from $N$.

Be reminded that this is a merely theoretical question, because the $\delta^+$-rules are clearly superior in practice if variables occur in the sequent that is to be reduced. Thus, in practice we apply $\delta^-$-rules only at the beginning of a proof before the first $\gamma$- or $\delta^+$-rule has been applied. And the main motivation for the $\delta^-$-rules is the generation of strong induction hypotheses and lemmas with as many free atoms as possible.

According to our merely practical intentions here (cf. § 3.6), we will not try to present a proof for $\delta^+$ to be always more liberalized than $\delta^-$, but just discuss some proof ideas and explain why it may be hard to find such a proof.

ad 1. First note that $\delta^-$-rules and the free atoms did not occur in inference systems with $\delta^+$-rules before the publication of [Wirth, 2004]; so in the earlier systems with free $\delta^+$-rules only, $\mathbb{V}\mathbb{A}(\forall x^{\mathbb{B}}.\ A)$ was indeed a subset of $\mathbb{V}(\varGamma\ \ \forall x^{\mathbb{B}}.\ A\ \ \varPi)$. Moreover, the additional atoms blocked by the $\delta^+$-rules (as compared to the $\delta^-$-rules) can hardly block any reductive proofs of formulas without free atoms and variables for the following reason:

If a proof uses only $\delta^+$-reductions, then there will be no (free) atoms around and the critical subset relation holds anyway. So a critical variable-condition can only arise if a $\delta^+$-step follows a $\delta^-$-step on the same branch. With a reasonably minimal positive/negative variable-condition $(P, N)$, the only additional cycles that could occur by the $\delta^+$-rule as compared to the alternative application of a $\delta^-$-rules are of the form

$$y^{\mathbb{V}}\ N\ z^{\mathbb{A}}\ P\ x^{\mathbb{V}}\ P^*\ w^{\mathbb{V}}\ P\ y^{\mathbb{V}},$$

resulting from the following scenario: $y^{\mathbb{V}}\ N\ z^{\mathbb{A}}$ results from a $\delta^-$-step, $z^{\mathbb{A}}\ P\ x^{\mathbb{V}}$ results from a subsequent $\delta^+$-step on the same branch, $x^{\mathbb{V}}\ P^*\ w^{\mathbb{V}}$ results from possible further $\delta^+$-steps ($\delta^-$-steps cannot produce a relevant cycle!) and instantiations of variables, and $w^{\mathbb{V}}\ P\ y^{\mathbb{V}}$ finally results from an instantiation of $y^{\mathbb{V}}$.
Let us now see what happens if we replace the $\delta^+$-step with a $\delta^-$-step with $x^{\mathbb{A}}$ replacing $x^{\mathbb{V}}$, *ceteris paribus*. Note that this is only possible if $x^{\mathbb{V}}$ was never instantiated, which again explains why there must be at least one step of $P$ between $x^{\mathbb{V}}$ and $y^{\mathbb{V}}$. If the variable $y^{\mathbb{V}}$ occurs in the upper sequent of this changed step, then new proof immediately fails due to the new cycle

$$y^{\mathbb{V}}\ N\ x^{\mathbb{A}}\ P^*\ w^{\mathbb{V}}\ P\ y^{\mathbb{V}}.$$

Otherwise, $y^{\mathbb{V}}$ was lost on this branch; but then we must ask ourselves why we instantiated it with a term containing $w^{\mathbb{V}}$. If $w^{\mathbb{V}}$ is essentially shared with another branch, on which $y^{\mathbb{V}}$ has survived, then it must occur in the sequent before the original $\delta^{+}$-step, and so we get the cycle

$$w^{\mathbb{V}}\ N\ x^{\mathbb{A}}\ P^{*}\ w^{\mathbb{V}}.$$

Otherwise, if $w^{\mathbb{V}}$ is not shared with another branch, we do not see any reason to instantiate $y^{\mathbb{V}}$ with a term containing $w^{\mathbb{V}}$. Indeed, if $w^{\mathbb{V}}$ is only on this branch, then there is no reason; if $w^{\mathbb{V}}$ occurs only on another branch, then a good reason for $x^{\mathbb{V}}\ P^{*}\ w^{\mathbb{V}}$ can be rejected just as for $y^{\mathbb{V}}$ before.

If we start with an input theorem in which variables and free atoms occur with non-empty variable-condition, it may become very hard to survey the situation and a proof will tend to be faulty.

ad 2. Also in this case we conjecture that, under side-conditions that can be easily met in practice, $\delta^{-}$-rules do not admit any successful proofs that are not possible with the analogous $\delta^{+}$-rules without extra complexity.

A proof of this conjecture, however, cannot be easy:

First, it is a global property which requires us to consider the entire inference system.

Second, $\delta^{-}$-rules actually do admit some extra $(P, N)$-substitutions, which have to be shown not to generate essential additional proofs. E.g., if we want to prove

$$\forall y^{\mathbb{B}}.\ \mathsf{Q}(a^{\mathbb{V}}, y^{\mathbb{B}}) \quad \wedge \quad \forall x^{\mathbb{B}}.\ \mathsf{Q}(x^{\mathbb{B}}, b^{\mathbb{V}}),$$

which is true for a reflexive ordering $\mathsf{Q}$ with a minimal and a maximal element, $\beta$- and $\delta^{-}$-rules reduce this to the two goals $\mathsf{Q}(a^{\mathbb{V}}, y^{\mathbb{A}})$ and $\mathsf{Q}(x^{\mathbb{A}}, b^{\mathbb{V}})$, extending the variable-condition with the positive/negative variable-condition $(P, N)$ given by $P = \emptyset$ and $N = \{(a^{\mathbb{V}}, y^{\mathbb{A}}),\ (b^{\mathbb{V}}, x^{\mathbb{A}})\}$. Then $\sigma_{\mathbb{A}} := \{a^{\mathbb{V}}{\mapsto}x^{\mathbb{A}},\ b^{\mathbb{V}}{\mapsto}y^{\mathbb{A}}\}$ is a $(P, N)$-substitution because the cycle

$$y^{\mathbb{A}}\ D_{\sigma_{\mathbb{A}}}\ b^{\mathbb{V}}\ N\ x^{\mathbb{A}}\ D_{\sigma_{\mathbb{A}}}\ a^{\mathbb{V}}\ N\ y^{\mathbb{A}}$$

needs more than one edge from $N$.

The analogous $\delta^{+}$-rules would have resulted in an extension with the positive/negative variable-condition $(P', N')$ given by $P' = \{(a^{\mathbb{V}}, y^{\mathbb{V}}),\ (b^{\mathbb{V}}, x^{\mathbb{V}})\}$ and $N' = \emptyset$. But then $\sigma_{\mathbb{V}} := \{a^{\mathbb{V}}{\mapsto}x^{\mathbb{V}},\ b^{\mathbb{V}}{\mapsto}y^{\mathbb{V}}\}$ is not a $(P', N')$-substitution due to the circle

$$y^{\mathbb{V}}\ D_{\sigma_{\mathbb{V}}}\ b^{\mathbb{V}}\ P'\ x^{\mathbb{V}}\ D_{\sigma_{\mathbb{V}}}\ a^{\mathbb{V}}\ P'\ y^{\mathbb{V}}.$$

# B Semantics for HILBERT's $\varepsilon$ in the Literature

Here in § B of the appendix, we will review the literature on the $\varepsilon$'s semantics with an emphasis on practical adequacy and the intentions of HILBERT's school in logic.

## B.1 Right-Unique Semantics

In contrast to the indefiniteness we suggested in § 4.9, nearly all semantics for HILBERT's $\varepsilon$ found elsewhere in the literature are functional, i.e. [*right-*] *unique*; cf. e.g. [LEISENRING, 1969] and the references there.

### B.1.1 Extensionality: Ackermann's (II,4) = Bourbaki's (S7) = Leisenring's (E2)

and in [LEISENRING, 1969] under the label (E2), we find the following axiom scheme, which we presented already in § 4.10:

$$\forall x^{\mathbb{B}}.\ (A_0 \Leftrightarrow A_1) \quad\Rightarrow\quad \varepsilon x^{\mathbb{B}}.\ A_0 \;=\; \varepsilon x^{\mathbb{B}}.\ A_1 \tag{E2}$$

This axiom (E2) must not be confused with the similar formula (E2′) from [WIRTH, 2008, Lemma 31, § 5.6] and [WIRTH, 2006b, Lemma 5.18, § 5.6], which reads in our new framework here as follows:

$$\forall x^{\mathbb{B}}.\ (A_0 \Leftrightarrow A_1) \quad\Rightarrow\quad x_0^{\mathbb{V}} \;=\; x_1^{\mathbb{V}} \tag{E2′}$$

For more detail, suppose that we have some $(P, N)$-choice-condition $C$ and two $\varepsilon$-free formulas $A_0, A_1$ with $\mathbb{B}(A_0, A_1) \subseteq \{x^{\mathbb{B}}\}$. Then we get two different fresh variables $x_0^{\mathbb{V}}, x_1^{\mathbb{V}} \in \mathbb{V} \setminus \mathbb{V}(A_0, A_1, C, P, N)$ of the same type as $x^{\mathbb{B}}$. Then we set $P' := P \cup \bigcup_{i=0}^{1} (\mathbb{VA}(A_i) \times \{x_i^{\mathbb{V}}\})$, $N' := N$, and $C' := C \cup \bigcup_{i=0}^{1} \{(x_i^{\mathbb{V}}, \varepsilon x^{\mathbb{B}}.\ A_i)\}$.

These details defined, our (E2′) turns out to be $(C', (P', N'))$-valid in any $\Sigma$-structure $\mathcal{S}$ according to Theorem 8.6, up to some assumption related to the Axiom of Choice.

Contrary to the valid proposition (E2′), however, (E2) is an axiom that imposes a right-unique behavior for the $\varepsilon$ (in the standard framework), depending on the extension of the formula forming the scope of an $\varepsilon$-binder on $x^{\mathbb{B}}$, seen as a predicate on $x^{\mathbb{B}}$. Indeed — from a semantic point of view — the value of $\varepsilon x^{\mathbb{B}}.\ A$ in each $\Sigma$-structure $\mathcal{S}$ is functionally dependent on the extension of the formula $A$, i.e. on

$$\{\ o \mid \operatorname{eval}(\mathcal{S} \uplus \{x^{\mathbb{B}} \mapsto o\})(A) = \mathsf{TRUE}\ \}.$$

Therefore, axiomatizations that have (E2) as an axiom or as a consequence of other axioms are called *extensional.*

Note that (E2) has a disastrous effect in intuitionistic logic: The contrapositive of (E2) — together with $(\varepsilon_0)$ and say "$0 \neq 1$" — turns every classical validity into an intuitionistic one.[5] For the strong consequences of the $\varepsilon$-formula in intuitionistic logic, see also Note 2.

### B.1.2 Weaker than (E2), but still Right-Unique

To overcome this disastrous effect and to get more options for the definition of a semantics of the $\varepsilon$ in general, in [ASSER, 1957], [MEYER-VIOL, 1995], and [GIESE & AHRENDT, 1999] the value of $\varepsilon x^{\mathbb{B}}.\ A$ may additionally depend on the *syntax* besides the semantics of the formula in the scope of the $\varepsilon$. The semantics of the $\varepsilon$ is then given as a function depending on a $\Sigma$-structure and on the syntactic details of the term $\varepsilon x^{\mathbb{B}}.\ A$. In [GIESE & AHRENDT, 1999, p.177] we read: "This definition contains no restriction whatsoever on the valuation of $\varepsilon$-terms." This claim, however, is not justified in its universality, because all considered options do still impose the restriction of a right-unique behavior; thereby the claim denies the possibility of an indefinite behavior as given in §§ 4.10 and 4.11. See also § B.2 for an alternative realization of an indefinite semantics.

### B.1.3 Overspecification even beyond (E2)

In [HERMES, 1965, p.18], the $\varepsilon$ suffers further overspecification in addition to (E2):

$$\varepsilon x.\,\mathsf{false} \;=\; \varepsilon x.\,\mathsf{true} \qquad (\varepsilon_5)$$

Roughly speaking, this axiom sets the value of a generalized choice function on the empty set to its value on the whole universe. For classical logic, we can combine (E2) and ($\varepsilon_5$) into the following axiom of [DEVIDI, 1995] for "<u>v</u>ery <u>ext</u>ensional" semantics:

$$\forall x.\left(\begin{array}{cc} & (\exists y.\,A_0\{x{\mapsto}y\} \;\Rightarrow\; A_0) \\ \Leftrightarrow & (\exists y.\,A_1\{x{\mapsto}y\} \;\Rightarrow\; A_1)\end{array}\right) \;\Rightarrow\; \varepsilon x.\,A_0 = \varepsilon x.\,A_1 \qquad \text{(vext)}$$

---

[5]( $0\neq 1$, $\varepsilon x^{\mathbb{B}}.\ A_0 \neq \varepsilon x^{\mathbb{B}}.\ A_1 \;\Rightarrow\; \neg(\forall x^{\mathbb{B}}.\ A_0 \wedge \forall x^{\mathbb{B}}.\ A_1)$ $\vdash$ $B \vee \neg B$ in Intuitionistic Logic)
For the proof of the slightly weaker result $0\neq 1$, (E2) $\vdash$ $B \vee \neg B$ for any formula $B$, cf. [BELL &AL., 2001, Proof of Theorem 6.4], which already occurs in more detail in [BELL, 1993a, § 3], and sketched in [BELL, 1993b, § 7].
Note that, for any implication $A \Rightarrow B$, its contrapositive $\neg B \Rightarrow \neg A$ is a consequence of it, and — in intuitionistic logic — a *proper* consequence in general.
Let $B$ be an arbitrary formula. By renaming we may w.l.o.g. assume that the free atom $x^{\mathbb{A}}$ of the $\varepsilon$-formula does not occur in $B$. We are going to show that $\vdash B \vee \neg B$ holds in intuitionistic logic under the assumptions of reflexivity, symmetry, and transitivity of "=", the $\varepsilon$-formula (or ($\varepsilon_0$)), and of the formulas $0\neq 1$ and $\varepsilon x^{\mathbb{B}}.\ A_0 \neq \varepsilon x^{\mathbb{B}}.\ A_1 \;\Rightarrow\; \neg(\forall x^{\mathbb{B}}.\ A_0 \wedge \forall x^{\mathbb{B}}.\ A_1)$.
Let $x^{\mathbb{B}}$ be a bound atom not occurring in $B$. Set $A_i := (B \vee x^{\mathbb{B}} = i)$ for $i \in \{0,1\}$.
Now all that we have to show is a trivial consequence of the following Claims 1 and 2,
$\varepsilon x^{\mathbb{B}}.\ A_0 \neq \varepsilon x^{\mathbb{B}}.\ A_1 \;\Rightarrow\; \neg(\forall x^{\mathbb{B}}.\ A_0 \wedge \forall x^{\mathbb{B}}.\ A_1)$, and Claim 3.
<u>Claim 1:</u> $0=0$, $1=1$, ($\varepsilon$-formula)$\{A{\mapsto}A_0\}\{x^{\mathbb{A}}{\mapsto}0\}$, ($\varepsilon$-formula)$\{A{\mapsto}A_1\}\{x^{\mathbb{A}}{\mapsto}1\}$
$\vdash$ $B$ $\vee$ $(\varepsilon x^{\mathbb{B}}.\ A_0 = 0$ $\wedge$ $\varepsilon x^{\mathbb{B}}.\ A_1 = 1)$.
<u>Claim 2:</u> $\varepsilon x^{\mathbb{B}}.\ A_0 = 0$ $\wedge$ $\varepsilon x^{\mathbb{B}}.\ A_1 = 1$, $0\neq 1$, $\forall x^{\mathbb{B}}, y^{\mathbb{B}}, z^{\mathbb{B}}.\ (y^{\mathbb{B}} = x^{\mathbb{B}} \wedge y^{\mathbb{B}} = z^{\mathbb{B}} \Rightarrow x^{\mathbb{B}} = z^{\mathbb{B}})$
$\vdash$ $\varepsilon x^{\mathbb{B}}.\ A_0 \neq \varepsilon x^{\mathbb{B}}.\ A_1$.
<u>Claim 3:</u> $\neg(\forall x^{\mathbb{B}}.\ A_0 \wedge \forall x^{\mathbb{B}}.\ A_1)$ $\vdash$ $\neg B$.
<u>Proof of Claim 1:</u> Because neither $x^{\mathbb{A}}$ nor $x^{\mathbb{B}}$ occur in $B$, and because $x^{\mathbb{A}}$ does not occur in $A_i$, the instances of the $\varepsilon$-formulas read $(B \vee i=i) \Rightarrow (B \vee \varepsilon x^{\mathbb{B}}.\ A_i = i)$. Thus, from $i=i$, we get $B \vee \varepsilon x^{\mathbb{B}}.\ A_i = i$. Thus, we get $(B \vee \varepsilon x^{\mathbb{B}}.\ A_0 = 0) \wedge (B \vee \varepsilon x^{\mathbb{B}}.\ A_1 = 1)$, thus $B$ $\vee$ $(\varepsilon x^{\mathbb{B}}.\ A_0 = 0$ $\wedge$ $\varepsilon x^{\mathbb{B}}.\ A_1 = 1)$ by distributivity. <u>Q.e.d. (Claim 1)</u>
<u>Proof of Claim 2:</u> Trivial. <u>Q.e.d. (Claim 2)</u>
<u>Proof of Claim 3:</u> As $x^{\mathbb{B}}$ does not occur in $B$, we get $B \vdash \forall x^{\mathbb{B}}.\ A_i$. The rest is trivial. <u>Q.e.d. (Claim 3)</u>

Indeed, (vext) implies (E2) and ($\varepsilon_5$). The other direction, however, does not hold for intuitionistic logic, where, roughly speaking, (vext) additionally implies that if the same elements make $A_0$ and $A_1$ as true as possible, then the $\varepsilon$-operator picks the same element of this set, even if the suprema $\exists y.\, A_0\{x\mapsto y\}$ and $\exists y.\, A_1\{x\mapsto y\}$ (in the complete HEYTING algebra) are not equally true.

### B.1.4 Strengthening Semantics to Turn Axiomatizations Complete

Although we have been concerned with soundness and safeness of our inference systems, we always accepted their incompleteness as the natural companion of semantics that are sufficiently weak to be useful in practice. Of course, completeness is the theoreticians' favorite puzzle because — as a global property of inference systems — it may be hard to prove, even for inconsistent systems. The objective of completeness gets particularly detached from practical usefulness, if a useful semantics is strengthened to obtain the completeness of a given inference system. Let us look at two examples for this procedure, resulting in practically useless semantics for the $\varepsilon$.

Different possible choices for the value of the generalized choice function on the empty set are discussed in [LEISENRING, 1969]. As the consequences of any special choice are quite queer, the only solution that is found to be sufficiently adequate in [LEISENRING, 1969] is validity in *all* models given by *all* generalized choice functions on the power-set of the universe. Note, however, that even in this case, in each model, the value of $\varepsilon x.\, A$ is *functionally* dependent on the extension of $A$.

Roughly speaking, in the textbook [LEISENRING, 1969], the axioms ($\varepsilon_1$) and ($\varepsilon_2$) from § 4.6 and (E2) from § 4.10 are shown to be complete w.r.t. this semantics of the $\varepsilon$ in first-order logic.

This completeness makes it unlikely that extensional semantics matches the intentions of HILBERT's school in logic. Indeed, if their intended semantics for the $\varepsilon$ could be completely captured by adding the single and straightforward axiom (E2), this axiom would not have been omitted in [HILBERT & BERNAYS, 1939]; it would at least be possible to derive (E2) from some axiomatization in [HILBERT & BERNAYS, 1939].

What makes LEISENRING's notion of validity problematic for theorem proving is that a proof has to consider all appropriate choice functions and cannot just pick an advantageous single one of them. More specifically, when LEISENRING does the step from satisfiability to validity he does the double duality switch from existence of a model and the existence of a choice function to all models and to all choice functions. Our notion of validity in Definition 7.7 does not switch the second duality, but stays with the *existence* of a choice function. Considering the influence that [LEISENRING, 1969] still has today, our avoidance of the universality requirement for choice functions in the definition of validity may be considered our practically most important conceptual contribution to the $\varepsilon$'s semantics. If we stuck to LEISENRING's definition of validity, then we would either have to give up the hope of finding proofs in practice, or have to avoid considering validity (beyond truth) in connection with HILBERT's $\varepsilon$, which is HARTLEY SLATER' solution, carefully observed in [SLATER, 1994; 2002; 2007b; 2009; 2011].

The misguiding procedure of strengthening semantics to obtain completeness for axiomatizations of the $\varepsilon$ actually originates in [ASSER, 1957]. The main objective of [ASSER, 1957], however, is to find a semantics such that already the basic $\varepsilon$-calculus of [HILBERT & BERNAYS, 1939] — not containing (E2) — is sound and complete for it. This semantics, however, has to depend on the details of the syntactic form of the $\varepsilon$-terms and, moreover, turns out to be necessarily so artificial that ASSER [1957] does not recommend it himself and admits that he thinks that it could not have been intended in [HILBERT & BERNAYS, 1939].

> "Allerdings ist dieser Begriff von Auswahlfunktion so kompliziert, daß sich seine Verwendung in der inhaltlichen Mathematik kaum empfiehlt."
> [ASSER, 1957, p. 59]
>
> "This notion of a choice function, however," (i.e. the type-3 choice function, providing a semantics for the $\varepsilon$-operator) "is so intricate that its application in contentual mathematics is hardly to be recommended."

> "Angesichts der Kompliziertheit des Begriffs der Auswahlfunktion dritter Art ergibt sich die Frage, ob bei HILBERT–BERNAYS (" ... ") wirklich beabsichtigt war, diesen Begriff von Auswahlfunktion axiomatisch zu beschreiben. Aus der Darstellung bei HILBERT–BERNAYS glaube ich entnehmen zu können, daß das nicht der Fall ist," [ASSER, 1957, p. 65]
>
> "The intricacy of the notion of the type-3 choice function puts up the question whether the intention in [HILBERT & BERNAYS, 1939] (" ... ") really was to describe this notion of choice function axiomatically. I believe I can draw from the presentation in [HILBERT & BERNAYS, 1939] that that is not the case,"

### B.1.5 Roots of the Misunderstanding of a Right-Uniqueness Requirement

The described prevalence of the right-uniqueness requirement may have its historical justification in the fact that, if we expand the dots "..." in the quotation preceding Example 4.2 in § 4.6.2, the full quotation on p.12 of [HILBERT & BERNAYS, 1939; 1970] reads:

> "Das $\varepsilon$-Symbol bildet somit eine Art der Verallgemeinerung des $\mu$-Symbols für einen beliebigen Individuenbereich. Der Form nach stellt es eine Funktion eines variablen Prädikates dar, welches außer demjenigen Argument, auf welches sich die zu dem $\varepsilon$-Symbol gehörige gebundene Variable bezieht, noch freie Variable als Argumente ("Parameter") enthalten kann. Der Wert dieser Funktion für ein bestimmtes Prädikat $A$ (bei Festlegung der Parameter) ist ein Ding des Individuenbereichs, und zwar ist dieses Ding gemäß der inhaltlichen Übersetzung der Formel ($\varepsilon_0$) *ein solches, auf das jenes Prädikat $A$ zutrifft, vorausgesetzt, daß es überhaupt auf ein Ding des Individuenbereichs zutrifft.*"
>
> "Thus, the $\varepsilon$-symbol forms a kind of generalization of the $\mu$-symbol for an arbitrary domain of individuals. According to its form, it constitues a function of a variable predicate, which may contain free variables as arguments ("parameters") in addition to the argument to which the bound variable of the $\varepsilon$-symbol refers. The value of this function for a given predicate $A$ (for fixed parameters) is a thing of the domain of individuals for which — according to the contentual translation of the formula ($\varepsilon_0$) — *the predicate $A$ holds, provided that $A$ holds for any thing of the domain of individuals at all.*"

Here the word "function" could be misunderstood in its narrower mathematical sense, namely to denote a (right-) unique relation. It is stated to be a function, however, only "according to its form", which — in the vernacular that becomes obvious from reading [HILBERT & BERNAYS, 2017b] — means nothing but "with respect to the process of the formation of formulas". Thus, HILBERT–BERNAYS' notation of the $\varepsilon$ takes the syntactic form of a function. This syntactic weakness was not bothering the work of HILBERT's school in logic in the field of proof theory. With our more practical intentions, the $\varepsilon$'s form of a function turns out as a problem even regarding syntax alone, cf. §§ 4.10 and 4.11. And we are not the only ones who have seen this applicational problem: For instance, in [HEUSINGER, 1997], an index was introduced to the $\varepsilon$ to overcome right-uniqueness.

If we nevertheless read "function" as a right-unique relation in the above quotation, what kind of function could be meant but a choice function, choosing an element from the set of objects that satisfy $A$, i.e. from its extension $\{\ o \mid \mathrm{eval}(\mathcal{S} \uplus \{x^{\mathbb{B}} \mapsto o\})(A) = \mathsf{TRUE}\ \}$. Accordingly, in the earlier publication [HILBERT, 1928], we read (p. 68):

> "Darüber hinaus hat das $\varepsilon$ die Rolle der Auswahlfunktion, d. h. im Falle, wo $A\,a$ auf mehrere Dinge zutreffen kann, ist $\varepsilon\,A$ *irgendeines* von den Dingen $a$, auf welche $A\,a$ zutrifft."
>
> "Beyond that, the $\varepsilon$ has the rôle of the choice function, i.e., if $A\,a$ may hold for several objects, $\varepsilon\,A$ is *an arbitrary one* of the things $a$ for which $A\,a$ holds."

Regarding the notation in this quotation, the syntax of the $\varepsilon$ is not that of a binder here, but a functional $\varepsilon : (i \to o) \to i$, applied to $A : i \to o$.

The meaning of having "the rôle of the choice function" is defined by the text that follows in the quotation. Thus, it is obvious that HILBERT wants to state the arbitrariness of choice as given by an arbitrary choice function, and that the word "function" does not refer to a requirement of right-uniqueness here.

Moreover, note that the definite article in "the choice function" (instead of the indefinite one) is in conflict with an interpretation as a mathematical function in the narrower sense as well.

Furthermore, DAVID HILBERT was sometimes pretty sloppy with the usage of choice functions in general: For instance, he may well have misinterpreted the consequences of the $\varepsilon$ on the Axiom of Choice (cf. [RUBIN & RUBIN, 1985], [HOWARD & RUBIN, 1998]) in the one but last paragraph of [HILBERT, 1923a]. Let us therefore point out the following: Although the $\varepsilon$ supplies us with a syntactic means for expressing an *indefinite universal (generalized) choice function* (cf. § 5.1), the axioms (E2), ($\varepsilon_0$), ($\varepsilon_1$), and ($\varepsilon_2$) do not imply the Axiom of Choice in set theories, unless the axiom schemes of Replacement (Collection) and Comprehension (Separation, Subset) also range over expressions containing the $\varepsilon$; cf. [LEISENRING, 1969, § IV 4.4].

HILBERT's school in logic may well have wanted to express what we call *"committed choice"* today, but they simply used the word "function" for the following three reasons:

1. They were not too much interested in semantics anyway.
2. The technical term "committed choice" did not exist at their time.
3. Last but not least, right-uniqueness conveniently serves as a global commitment to any choice and thereby avoids the problem illustrated in Example 4.8 of § 4.9.

## B.2 Indefinite Semantics in the Literature

The only occurrence of an indefinite semantics for HILBERT's $\varepsilon$ in the literature seems to be [BLASS & GUREVICH, 2000] (and the references there), unless we count the indexed $\varepsilon$ of [HEUSINGER, 1997] for indefinite indices as such a semantics as well. The right-uniqueness is actually so prevalent in the literature that a "$\delta$" is written instead of an "$\varepsilon$" in [BLASS & GUREVICH, 2000], because there the right-unique behavior is considered to be essential for the $\varepsilon$.

Consider the formula $\varepsilon x.\,(x{=}x) \;=\; \varepsilon x.\,(x{=}x)$ from [BLASS & GUREVICH, 2000] or the even simpler $\varepsilon x.\,\mathsf{true} \;=\; \varepsilon x.\,\mathsf{true}$ (discussed already in § 4.10), which may be valid or not, depending on the question whether the same object is taken on both sides of the equation or not. In natural language this like "Something is equal to something.", whose truth is indefinite. If you do not think so, consider $\varepsilon x.\,\mathsf{true} \;\neq\; \varepsilon x.\,\mathsf{true}$ in addition, i.e. "Something is unequal to something.", and notice that the two sentences seem to be contradictory.

In [BLASS & GUREVICH, 2000], KLEENE's strong three-valued logic is taken as a mathematically elegant means to solve the problems with indefiniteness. In spite of the theoretical significance of this solution, however, KLEENE's strong three-valued logic severely restricts its applicability from a practical point of view: In applications, a logic is not an object of investigation but a meta-logical tool, and logical arguments are never made explicit because the presence of logic is either not realized at all or taken to be trivial, even by academics (unless they are formalists); see, for instance, [PINKAL &AL., 2001, p.14f.] for Wizard of Oz studies with young students.

Therefore, regarding applications, we had better stick to our common meta-logic, which in the western world is a subset of (modal) classical logic: A western court may accept that LEE HARVEY OSWALD killed JOHN F. KENNEDY as well as that he did not — but cannot accept a third possibility, a *tertium,* as required for KLEENE's strong three-valued logic, and especially not the interpretation given in [BLASS & GUREVICH, 2000], namely that he *both* did and did not kill him, which contradicts any common sense.

# C On Formalizing Variable-Conditions

Note that the two relations $P$ and $N$ of a positive/negative variable-condition $(P,N)$ are always disjoint because their ranges must be disjoint according to Definition 5.4.

Thus, from a technical point of view, we could merge $P$ and $N$ into a single relation, but we prefer to have two relations for the two different functions (the positive and the negative one) of the variable-conditions in this paper, instead of the one relation for one function of [Wirth, 2002; 2004; 2006a; 2008; 2012b; 2006b], which realized the negative function only with a significant loss of relevant information.

Our main reason to have two different relations is that it may make sense to relax the restriction on the negative relation in future publications to

$$N \ \subseteq\ \mathbb{V} \times \mathbb{VA},$$

cf. e.g. Example 6.1. We do not know of any of our theorems that its proof would get into serious problems by this relaxation, but we have not meticulously checked this yet.

Moreover, in Definition 5.4, we have excluded the possibility that two free atoms $a^{\mathbb{A}}$, $b^{\mathbb{A}} \in \mathbb{A}$ may be related to each other in any of the two components of a positive/negative variable-condition $(P,N)$:

- $y^{\mathbb{VA}}\ P\ a^{\mathbb{A}}$ is excluded for intentional reasons: An atom $a^{\mathbb{A}}$ cannot depend on any other symbol $y^{\mathbb{VA}}$. In this sense an atom is indeed atomic and can be seen as a black box.

- $b^{\mathbb{A}}\ N\ a^{\mathbb{A}}$, however, is excluded for technical reasons only.
  Two distinct atoms $a^{\mathbb{A}}$, $b^{\mathbb{A}}$ in nominal terms [Urban &al., 2004] are indeed always fresh for each other: $a^{\mathbb{A}} \# b^{\mathbb{A}}$. In our notation, this would read: $b^{\mathbb{A}}\ N\ a^{\mathbb{A}}$.
  The reason why we did not include $(\mathbb{A}{\times}\mathbb{A}) \setminus {}_{\mathbb{A}}\!\rceil\mathrm{id}$ into the negative component $N$ is simply that we want to be close to the data structures of a both efficient and human-oriented graph implementation.
  Furthermore, consistency of a positive/negative variable-condition $(P,N)$ is equivalent to consistency of $\big(\ P,\ \ N \uplus ((\mathbb{A}{\times}\mathbb{A}) \setminus {}_{\mathbb{A}}\!\rceil\mathrm{id})\ \big)$.
  Indeed, if we added $(\mathbb{A}{\times}\mathbb{A}) \setminus {}_{\mathbb{A}}\!\rceil\mathrm{id}$ to $N$, the result of the acyclicity test of Corollary 5.6 would not be changed: If there were a cycle with a single edge from $(\mathbb{A}{\times}\mathbb{A}) \setminus {}_{\mathbb{A}}\!\rceil\mathrm{id}$, then its previous edge would have to be one of the original edges of $N$; and so this cycle would have more than one edge from $N \uplus ((\mathbb{A}{\times}\mathbb{A}) \setminus {}_{\mathbb{A}}\!\rceil\mathrm{id})$, and thus would not count as a counterexample to consistency.

Furthermore, we could remove the set $\mathbb{B}$ of bound atoms from our sets of symbols and consider its elements to be elements of the set $\mathbb{A}$ of atoms. Besides some additional care on free occurrences of atoms in § 5.3, an additional price we would have to pay for this removal is that we would have to include $\mathbb{V}{\times}\mathbb{B}$ as a subset into the negative component $N$ of each of our positive/negative variable-conditions $(P,N)$.

The reason for this inclusion is that we must guarantee that it is not possible that a bound atom $b^{\mathbb{B}}$ can be read by some variable $x^{\mathbb{V}}$, in particular after an elimination of binders. Then, by this inclusion, in case of $b^{\mathbb{B}}\ P^{+}\ x^{\mathbb{V}}$, we would get a cycle $b^{\mathbb{B}}\ P^{+}\ x^{\mathbb{V}}\ N\ b^{\mathbb{B}}$ with only one edge from $N$.

Although, in practical contexts, we can always get along with a finite subset of $\mathbb{V}{\times}\mathbb{B}$, the essential pairs of this subset would still be quite many and would be most confusing already in small examples. For instance, for the choice-condition of Example 4.11, almost four dozen pairs from $\mathbb{V}{\times}\mathbb{B}$ are technically required, compared to only a good dozen pairs that are actually relevant to the problem according to Example 5.13(a).

# Proofs

**Proof of Lemma 5.1**
The backward implications are trivial, the first because $R^+$-minimality in a class $A$ implies $R$-minimality in $A$ due to $R \subseteq R^+$.
For the forward implications, since $R^+$ is clearly transitive and any well-founded relation is irreflexive, it suffices to show that $R^+$ is well-founded.
Thus, assume that $R$ is well-founded and suppose that there is some class $A$ with $\forall a \in A.\ \exists a' \in A.\ a' R^+ a$. It suffices to show that $A$ must be empty. Set $B := \{\ b \mid \exists a \in A.\ a\, R^*\, b\ \}$.

Claim 1: For any $b \in B$, there is some $b' \in B$ with $b'\, R\, b$.
Proof of Claim 1: Assume $b \in B$. By definition of $B$ and our supposition on $A$, there are some $a, a' \in A$ with $a'\, R^+\, a\, R^*\, b$. Thus, $a'\, R^+\, b$. Thus, there is some $b'$ with $a'\, R^*\, b'\, R\, b$. And we also have $b' \in B$ then. Q.e.d. (Claim 1)

By Claim 1 and the assumption that $R$ is well-founded, we get $B = \emptyset$. Then, we also have $A = \emptyset$ due to $A \subseteq B$. **Q.e.d. (Lemma 5.1)**

**Proof of Lemma 5.17**
By assumption, $(C', (P', N'))$ is the extended $\sigma$-update of $(C, (P, N))$. Thus, $(P', N')$ is the $\sigma$-update of $(P, N)$. Thus, because $\sigma$ is a $(P, N)$-substitution, $(P', N')$ is a consistent positive/negative variable-condition by Definition 5.11. Moreover, $C$ is a $(P, N)$-choice-condition. Thus, $C$ is a partial function on $\mathbb{V}$, such that Items 1, 2, and 3 of Definition 5.12 hold. Thus, $C'$ is a partial function on $\mathbb{V}$ satisfying items 1 and 2 of Definition 5.12 as well. For $C'$ to satisfy also item 3 of Definition 5.12, it now suffices to show the following Claim 1.

Claim 1: Let $y^{\mathbb{V}} \in \mathrm{dom}(C')$ and $z^{\mathbb{VA}} \in \mathbb{VA}(C'(y^{\mathbb{V}}))$. Then we have $z^{\mathbb{VA}}\, (P')^+\, y^{\mathbb{V}}$.
Proof of Claim 1: By the definition of $C'$, we have $z^{\mathbb{VA}} \in \mathbb{VA}(C(y^{\mathbb{V}}))$ or else there is some $x^{\mathbb{V}} \in \mathrm{dom}(\sigma) \cap \mathbb{V}(C(y^{\mathbb{V}}))$ with $z^{\mathbb{VA}} \in \mathbb{VA}(\sigma(x^{\mathbb{V}}))$. Thus, as $C$ is a $(P, N)$-choice-condition, we have either $z^{\mathbb{VA}}\, P^+\, y^{\mathbb{V}}$ or else $x^{\mathbb{V}}\, P^+\, y^{\mathbb{V}}$ and $z^{\mathbb{VA}} \in \mathbb{VA}(\sigma(x^{\mathbb{V}}))$. Then, as $(P', N')$ is the $\sigma$-update of $(P, N)$, by Definition 5.10, we have either $z^{\mathbb{VA}}\, (P')^+\, y^{\mathbb{V}}$ or else $x^{\mathbb{V}}\, (P')^+\, y^{\mathbb{V}}$ and $z^{\mathbb{VA}}\, P'\, x^{\mathbb{V}}$. Thus, in any case, $z^{\mathbb{VA}}\, (P')^+\, y^{\mathbb{V}}$. Q.e.d. (Claim 1)

**Q.e.d. (Lemma 5.17)**

**Proof of Theorem 7.5**

Under the given assumptions, set $\lhd := P^+$ and $S_\pi := R$.

Claim A: $\lhd = P^+$ and $(P \cup S_\pi)^+$ are a well-founded orderings.

Claim B: $(P, N)$ and $(P \cup S_\pi, N)$ are consistent positive/negative variable-conditions.

Proof of Claims A and B: $(P, N)$ is a consistent variable-condition because $C$ is a $(P, N)$-choice-condition by assumption of the theorem. Thus, by Definition 5.5, $P$ is well-founded and $P^+{\circ}N$ is irreflexive. Thus, by Lemma 5.1, $\lhd = P^+$ is a well-founded ordering.
$R = {}_{\mathbb{A}}\rceil(P^+)$: In this case, we have $S_\pi \subseteq \lhd$. Thus, $(P \cup S_\pi)^+ = \lhd$ holds and $(P, N)$ is a weak extension of $(P \cup S_\pi, N)$. Thus, by Corollary 5.9, $(P \cup S_\pi, N)$ is a consistent positive/negative variable-condition as well.
$R = (\mathbb{A}{\times}\mathbb{V}) \setminus (P^*{\circ}N)^{-1}$: In this case, we have $S_\pi = (\mathbb{A}{\times}\mathbb{V}) \setminus (P^*{\circ}N)^{-1}$. To show that $P \cup S_\pi$ is well-founded, assume an arbitrary non-empty set $B \subseteq \mathbb{VA}$. As $P$ is well-founded there is a $P$-minimal $b^{\mathbb{VA}} \in B$. If $b^{\mathbb{VA}}$ is $S_\pi$-minimal, then it is also $(P \cup S_\pi)$-minimal as required. Otherwise there is some $a^{\mathbb{VA}} \in B$ with $a^{\mathbb{VA}} S_\pi b^{\mathbb{VA}}$. Then we have $a^{\mathbb{VA}} \in \mathbb{A}$ and thus it must be $(P \cup S_\pi)$-minimal as required, because of $\mathrm{ran}(P \cup S_\pi) \subseteq \mathbb{V}$.
Thus, $P \cup S_\pi$ is well-founded, and, again by Lemma 5.1, $(P \cup S_\pi)^+$ is a well-founded ordering.
To show that $(P \cup S_\pi)^+{\circ}N$ is irreflexive, let us assume the contrary: $x^{\mathbb{A}}(P \cup S_\pi)^+ y^{\mathbb{V}} N x^{\mathbb{A}}$. Because of $N \subseteq \mathbb{V}{\times}\mathbb{A}$, we indeed have $y^{\mathbb{V}} \in \mathbb{V}$ and $x^{\mathbb{A}} \in \mathbb{A}$. If the first step in this assumption is a $P$-step, then, because of $\mathrm{ran}(P) \subseteq \mathbb{V}$ and $\mathrm{dom}(S_\pi) \subseteq \mathbb{A}$, we actually have $x^{\mathbb{A}} P^+ y^{\mathbb{V}} N x^{\mathbb{A}}$, contradicting the irreflexivity of $P^+{\circ}N$. Otherwise, the first step must be an $S_\pi$-step, and then, because of $\mathrm{ran}(P \cup S_\pi) \subseteq \mathbb{V}$ and $\mathrm{dom}(S_\pi) \subseteq \mathbb{A}$, we actually have $x^{\mathbb{A}} S_\pi z^{\mathbb{V}} P^* y^{\mathbb{V}} N x^{\mathbb{A}}$ for some $z^{\mathbb{V}} \in \mathbb{V}$. Then, skipping the first step, we have $x^{\mathbb{A}}(P^*{\circ}N)^{-1} z^{\mathbb{V}}$, contradicting the first step $x^{\mathbb{A}} S_\pi z^{\mathbb{V}}$ by our case for $R$.
Thus, $(P \cup S_\pi, N)$ is a consistent positive/negative variable-condition.

Q.e.d. (Claims A and B)

Claim C: [$S_\rho \subseteq S_\pi$.]

Proof of Claim C: [$S_\rho \subseteq R$ holds by assumption of the theorem. Thus Claim C holds by our definition of $S_\pi$.] Q.e.d. (Claim C)

Claim D: $S_\pi \circ \lhd \;\subseteq\; S_\pi$.

Proof of Claim D:
$R = {}_{\mathbb{A}}\rceil(P^+)$: In this case, we have $S_\pi = {}_{\mathbb{A}}\rceil\lhd$, and thus $S_\pi \circ \lhd \;=\; {}_{\mathbb{A}}\rceil\lhd \circ \lhd \;\subseteq\; {}_{\mathbb{A}}\rceil\lhd \;=\; S_\pi$.
$R = (\mathbb{A}{\times}\mathbb{V}) \setminus (P^*{\circ}N)^{-1}$: In this case, we have $S_\pi = (\mathbb{A}{\times}\mathbb{V}) \setminus (P^*{\circ}N)^{-1}$. Let us assume $x^{\mathbb{A}} S_\pi y^{\mathbb{V}} P^+ z^{\mathbb{V}}$. Because of $S_\pi \subseteq \mathbb{A}{\times}\mathbb{V}$ and $\mathrm{ran}(P) \subseteq \mathbb{V}$, we indeed have $x^{\mathbb{A}} \in \mathbb{A}$ and $y^{\mathbb{V}}, z^{\mathbb{V}} \in \mathbb{V}$. From the first step and our case for $R$, we conclude that $y^{\mathbb{V}} P^*{\circ}N x^{\mathbb{A}}$ cannot hold. Thus, by the second step, neither $z^{\mathbb{V}} P^*{\circ}N x^{\mathbb{A}}$ nor $x^{\mathbb{A}}(P^*{\circ}N)^{-1} z^{\mathbb{V}}$ can be the case. Thus, we have $x^{\mathbb{A}} S_\pi z^{\mathbb{V}}$, as was to be shown. Q.e.d. (Claim D)

By recursion on $y^{\mathbb{V}} \in \mathbb{V}$ in $\lhd$, we can define $\pi(y^{\mathbb{V}}) : (S_\pi\langle\{y^{\mathbb{V}}\}\rangle \to \mathcal{S}) \to \mathcal{S}$ as follows.
Let $\tau' : S_\pi\langle\{y^{\mathbb{V}}\}\rangle \to \mathcal{S}$ be arbitrary.

$y^{\mathbb{V}} \in \mathbb{V}\backslash\mathrm{dom}(C)$: [If an $\mathcal{S}$-raising-valuation $\rho$ is given, then we set

$$\pi(y^{\mathbb{V}})(\tau') := \rho(y^{\mathbb{V}})({}_{S_\rho\langle\{y^{\mathbb{V}}\}\rangle}{\upharpoonright}\tau');$$

which is well-defined according to Claim C.]

Otherwise, we choose an arbitrary value for $\pi(y^{\mathbb{V}})(\tau')$ from the universe of $\mathcal{S}$ (of the appropriate type). Note that $\mathcal{S}$ is assumed to provide some function-choice function $\mathcal{S}(\exists)$ for the universe function $\mathcal{S}(\forall)$ according to § 7.1. Thus, for $y^{\mathbb{V}} : \alpha$, we take:

$$\pi(y^{\mathbb{V}})(\tau') := \mathcal{S}(\exists)_\alpha.$$

$y^{\mathbb{V}} \in \mathrm{dom}(C)$: In this case, we have the following situation: $C(y^{\mathbb{V}}) = \lambda v_0^{\mathbb{B}}, \ldots, v_{l-1}^{\mathbb{B}}.\ \varepsilon v_l^{\mathbb{B}}.\ B$ for some formula $B$ and some $v_0^{\mathbb{B}}, \ldots, v_l^{\mathbb{B}} \in \mathbb{B}$ with $v_0^{\mathbb{B}} : \alpha_0, \ \ldots, \ v_l^{\mathbb{B}} : \alpha_l$, $\mathbb{B}(B) \subseteq \{v_0^{\mathbb{B}}, \ldots, v_l^{\mathbb{B}}\}$, $y^{\mathbb{V}} : \alpha_0, \ldots, \alpha_{l-1} \to \alpha_l$, and $z^{\mathbb{VA}} \lhd y^{\mathbb{V}}$ for all $z^{\mathbb{VA}} \in \mathbb{VA}(B)$, because $C$ is a $(P, N)$-choice-condition, cf. item 3 of Definition 5.12. In particular, by Claim A, $y^{\mathbb{V}} \notin \mathbb{V}(B)$.

Let $\tau'' : (\mathbb{A}\backslash\mathrm{dom}(\tau')) \to \mathcal{S}$ be arbitrary and set $\delta := \mathsf{e}({}_{\lhd\langle\{y^{\mathbb{V}}\}\rangle}{\upharpoonright}\pi)(\tau' \uplus \tau'') \uplus \tau' \uplus \tau''$ .

In this case, with the help of the assumed generalized choice function on the power-set of the universe $\mathcal{S}(\forall)_{\alpha_l}$ for the choice type $\alpha_l$, for an arbitrary $\chi : \{v_0^{\mathbb{B}}, \ldots, v_{l-1}^{\mathbb{B}}\} \to \mathcal{S}$, we choose the object $i_\chi \in \mathcal{S}(\forall)_{\alpha_l}$ from the set

$$\left\{ \ i' \in \mathcal{S}(\forall)_{\alpha_l} \ \middle| \ \mathrm{eval}(\mathcal{S} \uplus \delta \uplus \chi \uplus \{v_l^{\mathbb{B}}{\mapsto}i'\})(B) = \mathsf{TRUE} \ \right\},$$

and then let $\pi(y^{\mathbb{V}})(\tau')$ be the function $f$ given by

$$f(\chi(v_0^{\mathbb{B}}), \ldots, \chi(v_{l-1}^{\mathbb{B}})) = i_\chi.$$

This point-wise definition of $f$ over the single arbitrary point $(\chi(v_0^{\mathbb{B}}), \ldots, \chi(v_{l-1}^{\mathbb{B}}))$ is correct. Indeed, by the Explicitness Lemma and because of $y^{\mathbb{V}} \notin \mathbb{V}(B)$, the choice of the value of $f(\chi(v_0^{\mathbb{B}}), \ldots, \chi(v_{l-1}^{\mathbb{B}}))$ does not depend on the values of $f(\chi''(v_0^{\mathbb{B}}), \ldots, \chi''(v_{l-1}^{\mathbb{B}}))$ for a different $\chi'' : \{v_0^{\mathbb{B}}, \ldots, v_{l-1}^{\mathbb{B}}\} \to \mathcal{S}$. Therefore, the function $f$ is well-defined because it also does not depend on $\tau''$ according to the Explicitness Lemma and Claim 1 below. Finally, $\pi$ is well-defined by induction on $\lhd$ according to Claim 2 below.

Claim 1: For $z^{\mathbb{VA}} \lhd y^{\mathbb{V}}$, the application term $(\delta \uplus \chi)(z^{\mathbb{VA}})$ has the value $\tau'(z^{\mathbb{VA}})$ in case of $z^{\mathbb{VA}} \in \mathbb{A}$, and the value $\pi(z^{\mathbb{VA}})({}_{S_\pi\langle\{z^{\mathbb{VA}}\}\rangle}{\upharpoonright}\tau')$ in case of $z^{\mathbb{VA}} \in \mathbb{V}$.

Claim 2: The definition of $\pi(y^{\mathbb{V}})(\tau')$ depends only on such values of $\pi(v^{\mathbb{V}})$ with $v^{\mathbb{V}} \lhd y^{\mathbb{V}}$, and does not depend on $\tau''$ at all.

Proof of Claim 1: Assume $z^{\mathbb{VA}} \lhd y^{\mathbb{V}}$. For $z^{\mathbb{VA}} \in \mathbb{V}$, we have $S_\pi\langle\{z^{\mathbb{VA}}\}\rangle \subseteq S_\pi\langle\{y^{\mathbb{V}}\}\rangle$ by Claim D, and thus the applicative term has the value $\pi(z^{\mathbb{VA}})({}_{S_\pi\langle\{z^{\mathbb{VA}}\}\rangle}{\upharpoonright}(\tau' \uplus \tau'')) = \pi(z^{\mathbb{VA}})({}_{S_\pi\langle\{z^{\mathbb{VA}}\}\rangle}{\upharpoonright}\tau')$. Moreover, for $z^{\mathbb{VA}} \in \mathbb{A}$, to show that the application term has the value $\tau'(z^{\mathbb{VA}})$, it suffices to show $z^{\mathbb{VA}} \in S_\pi\langle\{y^{\mathbb{V}}\}\rangle$. For the case of $R = {}_{\mathbb{A}}{\upharpoonright}(P^+)$, this is trivial. For the case of $R = (\mathbb{A}{\times}\mathbb{V}) \setminus (P^*{\circ}N)^{-1}$, the contrary assumption $z^{\mathbb{VA}} \notin S_\pi\langle\{y^{\mathbb{V}}\}\rangle$ can be refuted as follows: By definition of $S_\pi$ in this case, we first get $z^{\mathbb{VA}}(P^*{\circ}N)^{-1}y^{\mathbb{V}}$. Then, from this $y^{\mathbb{V}}P^*{\circ}Nz^{\mathbb{VA}}$. Finally, from $z^{\mathbb{VA}}P^+y^{\mathbb{V}}$, we get $z^{\mathbb{VA}}P^+{\circ}Nz^{\mathbb{VA}}$, contradicting the irreflexivity given by Claim B.

Q.e.d. (Claim 1)

Proof of Claim 2: In case of $y^{\mathbb{V}} \notin \mathrm{dom}(C)$, the definition of $\pi(y^{\mathbb{V}})(\tau')$ is immediate and independent. Otherwise, we have $z^{\mathbb{VA}} \lhd y^{\mathbb{V}}$ for all $z^{\mathbb{VA}} \in \mathbb{VA}(C(y^{\mathbb{V}}))$. Thus, Claim 2 follows from the Explicitness Lemma and Claim 1. Q.e.d. (Claim 2)

Moreover, $\pi : \mathbb{V} \to (\mathbb{A} \rightsquigarrow \mathcal{S}) \rightsquigarrow \mathcal{S}$ is obviously an $\mathcal{S}$-raising-valuation. Thus, item 1 of Definition 7.4 is satisfied for $\pi$ by Claim B.

To show that also item 2 of Definition 7.4 is satisfied, let us assume $y^{\mathbb{V}} \in \mathrm{dom}(C)$ and $\tau : \mathbb{A} \to \mathcal{S}$ to be arbitrary with $C(y^{\mathbb{V}}) = \lambda v_0^{\mathbb{B}}, \ldots, v_{l-1}^{\mathbb{B}}.\ \varepsilon v_l^{\mathbb{B}}.\ B$, and let us then assume to the contrary of item 2 that, for some $\chi' : \{v_0^{\mathbb{B}}, \ldots, v_l^{\mathbb{B}}\} \to \mathcal{S}$ and for $\delta' := \mathsf{e}(\pi)(\tau) \uplus \tau \uplus \chi'$ and $\sigma := \{v_l^{\mathbb{B}} \mapsto y^{\mathbb{V}}(v_0^{\mathbb{B}}, \ldots, v_{l-1}^{\mathbb{B}})\}$, we have $\mathrm{eval}(\mathcal{S} \uplus \delta')(B) = \mathsf{TRUE}$, but $\mathrm{eval}(\mathcal{S} \uplus \delta')(B\sigma) = \mathsf{FALSE}$.

Set $\tau' := {}_{S_\pi\langle\{y^{\mathbb{V}}\}\rangle}\!\upharpoonright\tau$, $\delta := {}_{\mathbb{VA}}\!\upharpoonright\delta'$, $\chi := {}_{\{v_0,\ldots,v_{l-1}\}}\!\upharpoonright\chi'$, and $\iota := \{v_l^{\mathbb{B}} \mapsto \pi(y^{\mathbb{V}})(\tau')(\chi(v_0^{\mathbb{B}}), \ldots, \chi(v_{l-1}^{\mathbb{B}}))\}$.

Then $\delta \uplus \chi' = \delta'$, and thus

$$\mathrm{eval}(\mathcal{S} \uplus \delta \uplus \chi \uplus {}_{\{v_l^{\mathbb{B}}\}}\!\upharpoonright\chi')(B) \;=\; \mathrm{eval}(\mathcal{S} \uplus \delta \uplus \chi')(B) \;=\; \mathrm{eval}(\mathcal{S} \uplus \delta')(B) \;=\; \mathsf{TRUE}.$$

This means that $\chi'(v_l^{\mathbb{B}})$ is an element of the set that was defined for the definition of $f$, and therefore $f(\chi(v_0^{\mathbb{B}}), \ldots, \chi(v_{l-1}^{\mathbb{B}}))$ must be from that set as well, i.e.

$$\mathrm{eval}(\mathcal{S} \uplus \delta \uplus \chi \uplus \iota)(B) = \mathsf{TRUE}.$$

By the Valuation-Lemma($l$), we have

$$\begin{array}{rl} & \mathrm{eval}(\mathcal{S} \uplus \delta')(y^{\mathbb{V}}(v_0^{\mathbb{B}}, \ldots, v_{l-1}^{\mathbb{B}})) \\ = & \delta'(y^{\mathbb{V}})(\delta'(v_0^{\mathbb{B}}), \ldots, \delta'(v_{l-1}^{\mathbb{B}})) \\ = & \mathsf{e}(\pi)(\tau)(y^{\mathbb{V}})(\chi'(v_0^{\mathbb{B}}), \ldots, \chi'(v_{l-1}^{\mathbb{B}})) \\ = & \pi(y^{\mathbb{V}})(\tau')(\chi(v_0^{\mathbb{B}}), \ldots, \chi(v_{l-1}^{\mathbb{B}})). \end{array}$$

Then we get

$$\iota = \sigma \circ \mathrm{eval}(\mathcal{S} \uplus \delta').$$

Moreover, $\delta \uplus \chi = {}_{\mathbb{VAB}\setminus\{v_l^{\mathbb{B}}\}}\!\upharpoonright\delta' = {}_{\mathbb{VAB}\setminus\{v_l^{\mathbb{B}}\}}\!\upharpoonright\mathrm{id} \circ \mathrm{eval}(\mathcal{S} \uplus \delta')$ by the Valuation-Lemma(0). Thus $\delta \uplus \chi \uplus \iota = \iota \uplus \delta \uplus \chi = (\sigma \uplus {}_{\mathbb{VAB}\setminus\{v_l^{\mathbb{B}}\}}\!\upharpoonright\mathrm{id}) \circ \mathrm{eval}(\mathcal{S} \uplus \delta')$. Thus, we have:

$$\begin{array}{rcl} \mathrm{eval}(\mathcal{S} \uplus \delta \uplus \chi \uplus \iota)(B) & = & \mathrm{eval}(\mathcal{S} \uplus ((\sigma \uplus {}_{\mathbb{VAB}\setminus\{v_l^{\mathbb{B}}\}}\!\upharpoonright\mathrm{id}) \circ \mathrm{eval}(\mathcal{S} \uplus \delta')))(B) \\ & = & \mathrm{eval}(\mathcal{S} \uplus \delta')(B\sigma) \;=\; \mathsf{FALSE}, \end{array}$$

where the second equation holds by the Substitution [Value] Lemma.

Thus we get the contradiction $\mathsf{TRUE} = \mathsf{FALSE}$. **Q.e.d. (Theorem 7.5)**

**Proof of Lemma 7.6**

Let us assume that $\pi$ is $\mathcal{S}$-compatible with $(C', (P', N'))$.

Then, by item 1 of Definition 7.4, $\pi : \mathbb{V} \to (\mathbb{A} \rightsquigarrow \mathcal{S}) \rightsquigarrow \mathcal{S}$ is an $\mathcal{S}$-raising-valuation and $(P' \cup S_\pi, N')$ is consistent. As $(P', N')$ is an extension of $(P, N)$, we have $P \subseteq P'$ and $N \subseteq N'$. Thus, $(P' \cup S_\pi, N')$ is an extension of $(P \cup S_\pi, N)$. Thus, $(P \cup S_\pi, N)$ is consistent by Corollary 5.9.

For $\pi$ to be $\mathcal{S}$-compatible with $(C, (P, N))$, it now suffices to show item 2 of Definition 7.4. As this property does not depend on the positive/negative variable-conditions anymore, it suffices to note that it actually holds because it holds for $C'$ by assumption and we also have $C \subseteq C'$ by assumption. **Q.e.d. (Lemma 7.6)**

**Proof of Lemma 7.10**

Let $C$ be a $(P,N)$-choice-condition. Let $y^{\mathbb{V}} \in \mathrm{dom}(C)$. Let $\mathcal{S}$ be a $\Sigma$-structure.

Then, by Definition 5.12 (Choice-Condition), for some mutually distinct variables $v_0^{\mathbb{B}}, \ldots, v_l^{\mathbb{B}}$ and some formula $B$, we have

$$\mathbb{B}(B) \subseteq \{v_0^{\mathbb{B}}, \ldots, v_l^{\mathbb{B}}\} \tag{1}$$

and

$$C(y^{\mathbb{V}}) \;=\; \lambda v_0^{\mathbb{B}}, \ldots, v_{l-1}^{\mathbb{B}}.\ \varepsilon v_l^{\mathbb{B}}.\ B.$$

Set $\sigma := \{v_l^{\mathbb{B}} \mapsto y^{\mathbb{V}}(v_0^{\mathbb{B}}, \ldots, v_{l-1}^{\mathbb{B}})\}$. Then, by Definition 4.12, we have

$$Q_C(y^{\mathbb{V}}) \;=\; \forall v_0^{\mathbb{B}}.\ \ldots \forall v_{l-1}^{\mathbb{B}}.\ \big(\ \exists v_l^{\mathbb{B}}.\ B \ \Rightarrow\ B\sigma\ \big).$$

Let $\pi$ be $\mathcal{S}$-compatible with $(C,(P,N))$; namely, in the case of item 1, the arbitrary $\pi$ mentioned in the lemma, or, in the case of item 2, the $\pi$ that exists according to Theorem 7.5. Let $\tau : \mathbb{A} \to \mathcal{S}$ be arbitrary. It now suffices to show

$$\mathrm{eval}(\mathcal{S} \uplus \mathsf{e}(\pi)(\tau) \uplus \tau)(Q_C(y^{\mathbb{V}})) = \mathsf{TRUE}.$$

By the backward direction of the $\forall$-LEMMA, it suffices to show

$$\mathrm{eval}(\mathcal{S} \uplus \delta)(\exists v_l^{\mathbb{B}}.\ B \ \Rightarrow\ B\sigma) = \mathsf{TRUE}$$

for an arbitrary $\chi : \{v_0^{\mathbb{B}}, \ldots, v_{l-1}^{\mathbb{B}}\} \to \mathcal{S}$, setting $\delta := \mathsf{e}(\pi)(\tau) \uplus \tau \uplus \chi$. By the backward direction of the $\Rightarrow$-LEMMA, it suffices to show

$$\mathrm{eval}(\mathcal{S} \uplus \delta)(B\sigma) = \mathsf{TRUE} \tag{2}$$

under the assumption of $\mathrm{eval}(\mathcal{S} \uplus \delta)(\exists v_l^{\mathbb{B}}.\ B) = \mathsf{TRUE}$. By the latter and the forward direction of the $\exists$-LEMMA, there is a $\chi' : \{v_l^{\mathbb{B}}\} \to \mathcal{S}$ such that $\mathrm{eval}(\mathcal{S} \uplus \delta \uplus \chi')(B) = \mathsf{TRUE}$. By the $\mathcal{S}$-compatibility of $\pi$ with $(C,(P,N))$ and by item 2 of Definition 7.4, we get

$$\mathrm{eval}(\mathcal{S} \uplus \delta \uplus \chi')(B\sigma) = \mathsf{TRUE}. \tag{3}$$

Set $\mathrm{X} := \mathbb{VAB}(B\sigma)$. By mutual distinctness of the variables $v_0^{\mathbb{B}}, \ldots, v_l^{\mathbb{B}}$ and by (1), we have

$$\begin{array}{lll}
\mathrm{X} & = & (\mathbb{VAB}(B) \setminus \{v_l^{\mathbb{B}}\}) \cup \mathbb{VAB}(y^{\mathbb{V}}(v_0^{\mathbb{B}}, \ldots, v_{l-1}^{\mathbb{B}})) \\
 & \subseteq & \mathbb{VA} \cup (\mathbb{B}(B) \setminus \{v_l^{\mathbb{B}}\}) \cup \{v_0^{\mathbb{B}}, \ldots, v_{l-1}^{\mathbb{B}}\} \\
 & \subseteq & \mathbb{VA} \cup (\{v_0^{\mathbb{B}}, \ldots, v_l^{\mathbb{B}}\} \setminus \{v_l^{\mathbb{B}}\}) \cup \{v_0^{\mathbb{B}}, \ldots, v_{l-1}^{\mathbb{B}}\} \\
 & = & \mathbb{VA} \cup \{v_0^{\mathbb{B}}, \ldots, v_{l-1}^{\mathbb{B}}\} \\
 & = & \mathrm{dom}(\delta)
\end{array}$$

Now, all we have to do to show (2) is to apply the EXPLICITNESS LEMMA twice, first in forward direction due to $\mathrm{X} \subseteq \mathrm{dom}(\delta)$, then in backward direction due to $\mathrm{X} \subseteq \mathrm{dom}(\delta \uplus \chi')$, and finally apply (3):

$$\begin{array}{lll}
\mathrm{eval}(\mathcal{S} \uplus \delta)(B\sigma) & = & \mathrm{eval}(\mathcal{S} \uplus {}_{\mathrm{X}}|\delta)(B\sigma) \\
 & = & \mathrm{eval}(\mathcal{S} \uplus {}_{\mathrm{X}}|(\delta \uplus \chi'))(B\sigma) \\
 & = & \mathrm{eval}(\mathcal{S} \uplus \delta \uplus \chi')(B\sigma) \\
 & = & \mathsf{TRUE}.
\end{array}$$

**Q.e.d. (Lemma 7.10)**

**Proof of Lemma 7.11**

Let us assume the situation described in the lemma.

We set $A := \mathrm{dom}(\sigma)\backslash(O'\uplus O)$. As $\sigma$ is a substitution on $\mathbb{V}$, we have $\mathrm{dom}(\sigma) \subseteq O'\uplus O\uplus A \subseteq \mathbb{V}$.

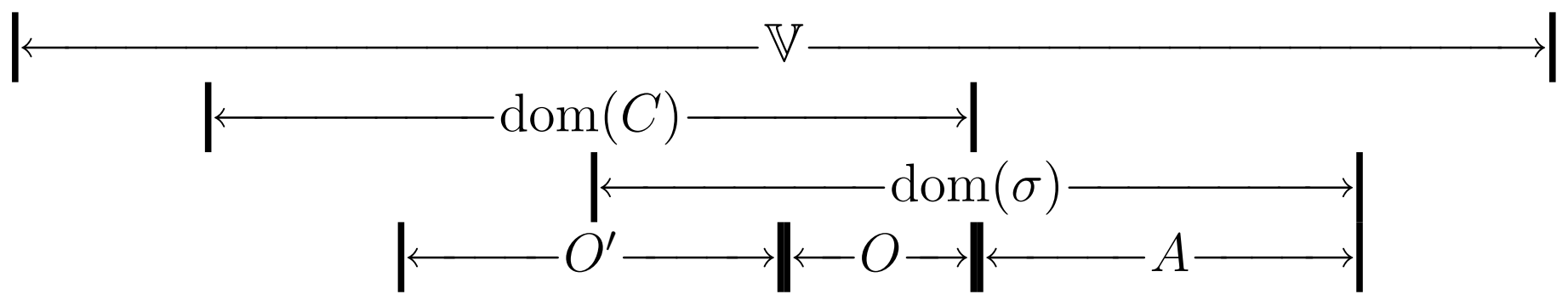

Note that $C'$ is a $(P',N')$-choice-condition by Lemma 5.17.

As $\pi'$ is $\mathcal{S}$-compatible with $(C',(P',N'))$, we know that $(P' \cup S_{\pi'}, N')$ s a consistent positive/negative variable-condition. Thus, $\lhd := (P' \cup S_{\pi'})^+$ is a well-founded ordering.

Let $D$ be the dependence relation of $\sigma$. Set $S_\pi := {}_{\mathbb{A}}{\upharpoonright}{\lhd}$.

<u>Claim 1:</u> We have $P', S_{\pi'}, P, D, S_\pi \subseteq \lhd$ and
$(P' \cup S_{\pi'}, N')$ is a weak extension of $(P \cup S_\pi, N)$ and of $(\lhd, N)$ (cf. Definition 5.7).

<u>Proof of Claim 1:</u> As $(P',N')$ is the $\sigma$-update of $(P,N)$, we have $P' = P \cup D$ and $N' = N$. Thus, $P', S_{\pi'}, P, D, S_\pi \subseteq (P' \cup S_{\pi'})^+ = \lhd$. <u>Q.e.d. (Claim 1)</u>

<u>Claim 2:</u> $(P \cup S_\pi, N)$ and $(\lhd, N)$ are consistent positive/negative variable-conditions.

<u>Proof of Claim 2:</u> This follows from Claim 1 by Corollary 5.9. <u>Q.e.d. (Claim 2)</u>

<u>Claim 3:</u> ${}_{O'}{\upharpoonright}C$ is an $(\lhd, N)$-choice-condition.

<u>Proof of Claim 3:</u> By Claims 1 and 2 and the assumption that $C$ is a $(P,N)$-choice-condition. <u>Q.e.d. (Claim 3)</u>

The plan for defining the $\mathcal{S}$-raising-valuation $\pi$ (which we have to find) is to give $\pi(y^{\mathbb{V}})({}_{S_\pi\langle\{y^{\mathbb{V}}\}\rangle}{\upharpoonright}\tau)$ a value as follows:

($\alpha$) For $y^{\mathbb{V}} \in \mathbb{V}\backslash(O'\uplus O\uplus A)$, we take this value to be
$$\pi'(y^{\mathbb{V}})({}_{S_{\pi'}\langle\{y^{\mathbb{V}}\}\rangle}{\upharpoonright}\tau).$$
This is indeed possible because of $S_{\pi'} \subseteq {}_{\mathbb{A}}{\upharpoonright}{\lhd} = S_\pi$, so ${}_{S_{\pi'}\langle\{y^{\mathbb{V}}\}\rangle}{\upharpoonright}\tau \subseteq {}_{S_\pi\langle\{y^{\mathbb{V}}\}\rangle}{\upharpoonright}\tau$.

($\beta$) For $y^{\mathbb{V}} \in O\uplus A$, we take this value to be
$$\mathrm{eval}(\mathcal{S} \uplus \mathsf{e}(\pi')(\tau) \uplus \tau)(\sigma(y^{\mathbb{V}})).$$
Note that, in case of $y^{\mathbb{V}} \in O$, we know that $(Q_C(y^{\mathbb{V}}))\sigma$ is $(\pi',\mathcal{S})$-valid by assumption of the lemma. Moreover, the case of $y^{\mathbb{V}} \in A$ is unproblematic because of $y^{\mathbb{V}} \notin \mathrm{dom}(C)$. Again, $\pi$ is well-defined in this case because the only part of $\tau$ that is accessed by the given value is ${}_{S_\pi\langle\{y^{\mathbb{V}}\}\rangle}{\upharpoonright}\tau$. Indeed, this can be seen as follows: By Claim 1 and the transitivity of $\lhd$, we have: ${}_{\mathbb{A}}{\upharpoonright}D \cup S_{\pi'}\circ D \subseteq {}_{\mathbb{A}}{\upharpoonright}{\lhd} = S_\pi$.

($\gamma$) For $y^{\mathbb{V}} \in O'$, however, we have to take care of $\mathcal{S}$-compatibility with $(C,(P,N))$ explicitly in an $\lhd$-recursive definition on the basis a function $\rho$ implementing ($\alpha$) and ($\beta$). This disturbance does not interfere with the semantic invariance stated in the lemma because occurrences of variables from $O'$ are explicitly excluded in the relevant terms and formulas and, according to the statement of lemma, $O'$ satisfies the appropriate closure condition.

Set $S_\rho := S_\pi$. Let $\rho$ be defined by $(y^{\mathrm{v}} \in \mathbb{V},\ \tau : \mathbb{A} \to \mathcal{S})$

$$\rho(y^{\mathrm{v}})({}_{S_\pi\langle\{y^{\mathrm{v}}\}\rangle}{\upharpoonright}\tau) := \begin{cases} \pi'(y^{\mathrm{v}})({}_{S_{\pi'}\langle\{y^{\mathrm{v}}\}\rangle}{\upharpoonright}\tau) & \text{if } y^{\mathrm{v}} \in \mathbb{V}\backslash(O \uplus A) \\ \mathrm{eval}(\mathcal{S} \uplus \mathsf{e}(\pi')(\tau) \uplus \tau)(\sigma(y^{\mathrm{v}})) & \text{if } y^{\mathrm{v}} \in O \uplus A \end{cases}$$

Let $\pi$ be the $\mathcal{S}$-raising-valuation that exists according to Theorem 7.5 for the $\mathcal{S}$-raising-valuation $\rho$ and the $(\lhd, N)$-choice-condition ${}_{O'}{\upharpoonright}C$ (cf. Claim 3). Note that the assumptions of Theorem 7.5 are indeed satisfied here and that the resulting semantic relation $S_\pi$ of Theorem 7.5 is indeed identical to our pre-defined relation of the same name, thereby justifying our abuse of notation: Indeed, by assumption of Lemma 7.11, for every choice type $\alpha$ of ${}_{O'}{\upharpoonright}C$, there is a generalized choice function on the power-set of the universe of $\mathcal{S}$ for the type $\alpha$; and we have

$$S_\rho \ = \ S_\pi \ = \ {}_{\mathbb{A}}{\upharpoonright}\lhd \ = \ {}_{\mathbb{A}}{\upharpoonright}(\lhd^+).$$

Because of $\mathrm{dom}({}_{O'}{\upharpoonright}C) = O'$, according to Theorem 7.5, we then have

$${}_{\mathbb{V}\backslash O'}{\upharpoonright}\pi \ = \ {}_{\mathbb{V}\backslash O'}{\upharpoonright}\rho$$

and $\pi$ is $\mathcal{S}$-compatible with $({}_{O'}{\upharpoonright}C, (\lhd, N))$.

<u>Claim 4:</u> For all $y^{\mathrm{v}} \in O \uplus A$ and $\tau : \mathbb{A} \to \mathcal{S}$, when we set $\delta' := \mathsf{e}(\pi')(\tau) \uplus \tau$:

$$\mathsf{e}(\pi)(\tau)(y^{\mathrm{v}}) = \mathrm{eval}(\mathcal{S} \uplus \delta')(\sigma(y^{\mathrm{v}})).$$

<u>Proof of Claim 4:</u> We have $O \uplus A \subseteq \mathbb{V}\backslash O'$. Thus, Claim 4 follows immediately from the definition of $\rho$. <u>Q.e.d. (Claim 4)</u>

<u>Claim 5:</u> For all $y^{\mathrm{v}} \in \mathbb{V}\backslash(O' \uplus O \uplus A)$ and $\tau : \mathbb{A} \to \mathcal{S}$: $\mathsf{e}(\pi)(\tau)(y^{\mathrm{v}}) = \mathsf{e}(\pi')(\tau)(y^{\mathrm{v}})$.

<u>Proof of Claim 5:</u> For $y^{\mathrm{v}} \in \mathbb{V}\backslash(O' \uplus O \uplus A)$, we have $y^{\mathrm{v}} \in \mathbb{V}\backslash O'$ and $y^{\mathrm{v}} \in \mathbb{V}\backslash(O \uplus A)$. Thus, $\mathsf{e}(\pi)(\tau)(y^{\mathrm{v}}) = \pi(y^{\mathrm{v}})({}_{S_\pi\langle\{y^{\mathrm{v}}\}\rangle}{\upharpoonright}\tau) = \rho(y^{\mathrm{v}})({}_{S_\pi\langle\{y^{\mathrm{v}}\}\rangle}{\upharpoonright}\tau) = \pi'(y^{\mathrm{v}})({}_{S_{\pi'}\langle\{y^{\mathrm{v}}\}\rangle}{\upharpoonright}\tau) = \mathsf{e}(\pi')(\tau)(y^{\mathrm{v}})$. <u>Q.e.d. (Claim 5)</u>

<u>Claim 6:</u> For any term or formula $B$ (possibly with some unbound occurrences of bound atoms from the set $W \subseteq \mathbb{B}$) with $O' \cap \mathbb{V}(B) = \emptyset$, and for every $\tau : \mathbb{A} \to \mathcal{S}$ and every $\chi : W \to \mathcal{S}$, when we set $\delta := \mathsf{e}(\pi)(\tau) \uplus \tau$ and $\delta' := \mathsf{e}(\pi')(\tau) \uplus \tau$, we have

$$\mathrm{eval}(\mathcal{S} \uplus \delta' \uplus \chi)(B\sigma) = \mathrm{eval}(\mathcal{S} \uplus \delta \uplus \chi)(B).$$

<u>Proof of Claim 6:</u> $\mathrm{eval}(\mathcal{S} \uplus \delta' \uplus \chi)(B\sigma) =$ (by the SUBSTITUTION [VALUE] LEMMA)
$\mathrm{eval}(\mathcal{S} \uplus (\sigma \uplus {}_{\mathbb{VAB}\backslash\mathrm{dom}(\sigma)}{\upharpoonright}\mathrm{id}) \circ \mathrm{eval}(\mathcal{S} \uplus \delta' \uplus \chi))(B) =$
(by the EXPLICITNESS LEMMA and the VALUATION-LEMMA(0))
$\mathrm{eval}(\mathcal{S} \uplus (\sigma \circ \mathrm{eval}(\mathcal{S} \uplus \delta')) \uplus {}_{\mathbb{VA}\backslash\mathrm{dom}(\sigma)}{\upharpoonright}\delta' \uplus \chi)(B) =$
(by $O \uplus A \subseteq \mathrm{dom}(\sigma) \subseteq O' \uplus O \uplus A$, $O' \cap \mathbb{V}(B) = \emptyset$, and the EXPLICITNESS LEMMA)
$\mathrm{eval}(\mathcal{S} \uplus {}_{O \uplus A}{\upharpoonright}\sigma \circ \mathrm{eval}(\mathcal{S} \uplus \delta') \uplus {}_{\mathbb{VA}\backslash(O' \uplus O \uplus A)}{\upharpoonright}\delta' \uplus \chi)(B) =$ (by Claim 4 and Claim 5)
$\mathrm{eval}(\mathcal{S} \uplus {}_{O \uplus A}{\upharpoonright}\delta \uplus {}_{\mathbb{VA}\backslash(O' \uplus O \uplus A)}{\upharpoonright}\delta \uplus \chi)(B) =$
(by $O' \cap \mathbb{V}(B) = \emptyset$ and the EXPLICITNESS LEMMA)
$\mathrm{eval}(\mathcal{S} \uplus \delta \uplus \chi)(B)$. <u>Q.e.d. (Claim 6)</u>

<u>Claim 7:</u> For every set of sequents $G$ (possibly with some unbound occurrences of bound atoms from the set $W \subseteq \mathbb{B}$) with $O' \cap \mathbb{V}(G) = \emptyset$, and for every $\tau : \mathbb{A} \to \mathcal{S}$ and for every $\chi : W \to \mathcal{S}$: Truth of $G$ in $\mathcal{S} \uplus \mathsf{e}(\pi)(\tau) \uplus \tau \uplus \chi$ is equivalent to truth of $G\sigma$ in $\mathcal{S} \uplus \mathsf{e}(\pi')(\tau) \uplus \tau \uplus \chi$.

Proof of Claim 7: This is a trivial consequence of Claim 6. Q.e.d. (Claim 7)

Claim 8: For $y^{\mathbb{V}} \in \mathrm{dom}(C) \setminus O'$, we have $O' \cap \mathbb{V}(C(y^{\mathbb{V}})) = \emptyset$.

Proof of Claim 8: Otherwise there is some $y^{\mathbb{V}} \in \mathrm{dom}(C) \setminus O'$ and some $z^{\mathbb{V}} \in O' \cap \mathbb{V}(C(y^{\mathbb{V}}))$. Then $z^{\mathbb{V}} P^+ y^{\mathbb{V}}$ because $C$ is a $(P, N)$-choice-condition, and then, as $\langle O' \rangle P^+ \cap \mathrm{dom}(C) \subseteq O'$ by assumption of the lemma, we have the contradicting $y^{\mathbb{V}} \in O'$. Q.e.d. (Claim 8)

Claim 9: Let $y^{\mathbb{V}} \in \mathrm{dom}(C)$ and $C(y^{\mathbb{V}}) = \lambda v_0^{\mathbb{B}}, \ldots, v_{l-1}^{\mathbb{B}}.\ \varepsilon v_l^{\mathbb{B}}.\ B$. Let $\tau : \mathbb{A} \to \mathcal{S}$ and $\chi : \{v_0^{\mathbb{B}}, \ldots, v_l^{\mathbb{B}}\} \to \mathcal{S}$. Set $\delta := \mathsf{e}(\pi)(\tau) \uplus \tau \uplus \chi$. Set $\mu := \{v_l^{\mathbb{B}} \mapsto y^{\mathbb{V}}(v_0^{\mathbb{B}}) \cdots (v_{l-1}^{\mathbb{B}})\}$. If $B$ is true in $\mathcal{S} \uplus \delta$, then $B\mu$ is true in $\mathcal{S} \uplus \delta$ as well.

Proof of Claim 9: Set $\delta' := \mathsf{e}(\pi')(\tau) \uplus \tau \uplus \chi$.

$y^{\mathbb{V}} \notin O' \uplus O$: In this case, because of $\mathrm{dom}(\sigma) \cap \mathrm{dom}(C) \subseteq O' \uplus O$, we have $y^{\mathbb{V}} \notin \mathrm{dom}(\sigma)$. Thus, as $(C', (P', N'))$ is the extended $\sigma$-update of $(C, (P, N))$, we have $C'(y^{\mathbb{V}}) = (C(y^{\mathbb{V}}))\sigma$. By Claim 8, we have $O' \cap \mathbb{V}(B) = \emptyset$.
And then, by our case assumption, also $O' \cap \mathbb{V}(B\mu) = \emptyset$.
By assumption of Claim 9, $B$ is true in $\mathcal{S} \uplus \delta$. Thus, by Claim 7, $B\sigma$ is true in $\mathcal{S} \uplus \delta'$. Thus, as $\pi'$ is $\mathcal{S}$-compatible with $(C', (P', N'))$, we know that $(B\sigma)\mu$ is true in $\mathcal{S} \uplus \delta'$. Because of $y^{\mathbb{V}} \notin \mathrm{dom}(\sigma)$, $v_0^{\mathbb{B}}, \ldots, v_l^{\mathbb{B}} \in \mathbb{B}$, and $\mathbb{B}(\mathrm{dom}(\sigma) \cup \mathrm{ran}(\sigma)) = \emptyset$, we have $(B\sigma)\mu = B(\sigma \uplus \mu) = (B\mu)\sigma$. Thus, $(B\mu)\sigma$ is true in $\mathcal{S} \uplus \delta'$ as well. Thus, by Claim 7, $B\mu$ is true in $\mathcal{S} \uplus \delta$.

$y^{\mathbb{V}} \in O$: By Claim 8, we have $O' \cap \mathbb{V}(B) = \emptyset$.
And then, by our case assumption, also $O' \cap \mathbb{V}(B\mu) = \emptyset$.
Moreover, $(Q_C(y^{\mathbb{V}}))\sigma$ is equal to $\forall v_0^{\mathbb{B}}.\ \ldots \forall v_{l-1}^{\mathbb{B}}.\ (\ \exists v_l^{\mathbb{B}}.\ B \ \Rightarrow\ B\mu\ )\sigma$ and $(\pi', \mathcal{S})$-valid by assumption of the lemma. Thus, by the forward direction of the $\forall$-Lemma, $(\ \exists v_l^{\mathbb{B}}.\ B \ \Rightarrow\ B\mu\ )\sigma$ is true in $\mathcal{S} \uplus \delta'$. Thus, by Claim 7, $\exists v_l^{\mathbb{B}}.\ B \ \Rightarrow\ B\mu$ is true in $\mathcal{S} \uplus \delta$. As, by assumption of Claim 9, $B$ is true in $\mathcal{S} \uplus \delta$, by the backward direction of the $\exists$-Lemma, $\exists v_l^{\mathbb{B}}.\ B$ is true in $\mathcal{S} \uplus \delta$ as well. Thus, by the forward direction of the $\Rightarrow$-Lemma, $B\mu$ is true in $\mathcal{S} \uplus \delta$ as well.

$y^{\mathbb{V}} \in O'$: $\pi$ is $\mathcal{S}$-compatible with $({}_{O'}{\upharpoonright}C, (\lhd, N))$ by definition, as explicitly stated before Claim 4. Thus, in this case, Claim 9 is just the respective item 2 of Definition 7.4 (Compatibility). Q.e.d. (Claim 9)

By Claims 2 and 9, $\pi$ is $\mathcal{S}$-compatible with $(C, (P, N))$. And then items 1 and 2 of the lemma are trivial consequences of Claims 6 and 7, respectively.

**Q.e.d. (Lemma 7.11)**

### Proof of Theorem 8.3

To illustrate our techniques, we only treat the first rule of each kind; the other rules can be treated most similarly. In the situation described in the theorem, it suffices to show that $C'$ is a $(P', N')$-choice-condition (because the other properties of an extended extension are trivial), and that, for every $\mathcal{S}$-raising-valuation $\pi$ that is $\mathcal{S}$-compatible with $(C', (P', N'))$, the sets $G_0$ and $G_1$ of the upper and lower sequents of the inference rule are equivalent w.r.t. their $(\pi, \mathcal{S})$-validity.

<u>$\gamma$-rule:</u> In this case we have $(C', (P', N')) = (C, (P, N))$. Thus, $C'$ is a $(P', N')$-choice-condition by assumption of the theorem. Moreover, for every $\mathcal{S}$-valuation $\tau : \mathbb{A} \to \mathcal{S}$, and for $\delta := \mathsf{e}(\pi)(\tau) \uplus \tau$, the truths of

$$\{\mathit{\Gamma} \quad \exists y^{\mathbb{B}}.\ A \quad \mathit{\Pi}\} \quad \text{and} \quad \{A\{y^{\mathbb{B}}{\mapsto}t\} \quad \mathit{\Gamma} \quad \exists y^{\mathbb{B}}.\ A \quad \mathit{\Pi}\}$$

in $\mathcal{S}{\uplus}\delta$ are indeed equivalent. The implication from left to right is trivial because the former sequent is a sub-sequent of the latter.

For the other direction, assume that $A\{y^{\mathbb{B}}{\mapsto}t\}$ is true in $\mathcal{S}{\uplus}\delta$. Thus, by the Substitution [Value] Lemma (second equation) and the Valuation-Lemma(0) (third equation):

$$\begin{aligned}\mathsf{TRUE} &= \mathrm{eval}(\mathcal{S} \uplus \delta)(A\{y^{\mathbb{B}}{\mapsto}t\}) \\ &= \mathrm{eval}(\mathcal{S} \uplus ((\{y^{\mathbb{B}}{\mapsto}t\} \uplus {}_{\mathbb{VAB}\setminus\{y^{\mathbb{B}}\}}\!\upharpoonright\!\mathrm{id}) \circ \mathrm{eval}(\mathcal{S}{\uplus}\delta)))(A) \\ &= \mathrm{eval}(\mathcal{S} \uplus \{y^{\mathbb{B}}{\mapsto}\mathrm{eval}(\mathcal{S}{\uplus}\delta)(t)\} \uplus \delta)(A)\end{aligned}$$

Thus, by the backward direction of the $\exists$-Lemma, $\exists y^{\mathbb{B}}.\ A$ is true in $\mathcal{S}{\uplus}\delta$. Thus, the upper sequent is true $\mathcal{S}{\uplus}\delta$.

<u>$\delta^-$-rule:</u> In this case, we have $x^{\mathbb{A}} \in \mathbb{A} \setminus (\mathrm{dom}(P) \cup \mathbb{A}(\Gamma, A, \Pi))$, $C'' = \emptyset$, and $V = \mathbb{V}(\Gamma\ \ \forall x^{\mathbb{B}}.\ A\ \ \Pi) \times \{x^{\mathbb{A}}\}$. Thus, $C' = C$, $P' = P$, and $N' = N \cup V$.

<u>Claim 1:</u> $C'$ is a $(P', N')$-choice-condition.

<u>Proof of Claim 1:</u> By assumption of the theorem, $C$ is a $(P, N)$-choice-condition. Thus, $(P, N)$ is a consistent positive/negative variable-condition. By Definition 5.5, $P$ is well-founded and $P^+ \circ N$ is irreflexive. Since $x^{\mathbb{A}} \notin \mathrm{dom}(P)$, we have $x^{\mathbb{A}} \notin \mathrm{dom}(P^+)$. Thus, because of $\mathrm{ran}(V) = \{x^{\mathbb{A}}\}$, also $P^+ \circ N'$ is irreflexive. Thus, $(P', N')$ is a consistent positive/negative variable-condition, and $C'$ is a $(P', N')$-choice-condition. <u>Q.e.d. (Claim 1)</u>

Now, for the soundness direction, it suffices to show the contrapositive, namely to assume that there is an $\mathcal{S}$-valuation $\tau : \mathbb{A} \to \mathcal{S}$ such that $\{\Gamma\ \ \forall x^{\mathbb{B}}.\ A\ \ \Pi\}$ is false in $\mathcal{S} \uplus \mathsf{e}(\pi)(\tau) \uplus \tau$, and to find an $\mathcal{S}$-valuation $\tau' : \mathbb{A} \to \mathcal{S}$ such that $\{A\{x^{\mathbb{B}} \mapsto x^{\mathbb{A}}\}\ \Gamma\ \Pi\}$ is false in $\mathcal{S} \uplus \mathsf{e}(\pi)(\tau') \uplus \tau'$. Under this assumption, the sequent $\Gamma\Pi$ is false in $\mathcal{S} \uplus \mathsf{e}(\pi)(\tau) \uplus \tau$.

<u>Claim 2:</u> $\Gamma\Pi$ is false in $\mathcal{S} \uplus \mathsf{e}(\pi)(\tau') \uplus \tau'$ for all $\tau' : \mathbb{A} \to \mathcal{S}$ with ${}_{\mathbb{A}\setminus\{x^{\mathbb{A}}\}}\!\upharpoonright\tau' = {}_{\mathbb{A}\setminus\{x^{\mathbb{A}}\}}\!\upharpoonright\tau$.

<u>Proof of Claim 2:</u> Because of $x^{\mathbb{A}} \notin \mathbb{A}(\Gamma\Pi)$, by the EXPLICITNESS LEMMA, if Claim 2 did not hold, there would have to be some $u^{\mathbb{V}} \in \mathbb{V}(\Gamma\Pi)$ with $x^{\mathbb{A}}\ S_\pi\ u^{\mathbb{V}}$. Then we have $u^{\mathbb{V}}\ N'\ x^{\mathbb{A}}$. Thus, we know that $(P' \cup S_\pi)^+ \circ N'$ is not irreflexive, which contradicts $\pi$ being $\mathcal{S}$-compatible with $(C', (P', N'))$. <u>Q.e.d. (Claim 2)</u>

Moreover, under the above assumption, also $\forall x^{\mathbb{B}}.\ A$ is false in $\mathcal{S} \uplus \mathsf{e}(\pi)(\tau) \uplus \tau$. By the backward direction of the $\forall$-LEMMA, this means that there is some object $o$ such that $A$ is false in $\mathcal{S} \uplus \{x^{\mathbb{B}} \mapsto o\} \uplus \mathsf{e}(\pi)(\tau) \uplus \tau$. Set $\tau' := {}_{\mathbb{A}\setminus\{x^{\mathbb{A}}\}}\!\upharpoonright\tau \uplus \{x^{\mathbb{A}} \mapsto o\}$. Then, by the SUBSTITUTION [VALUE] LEMMA (1st equation), by the VALUATION-LEMMA(0) (2nd equation), and by the EXPLICITNESS LEMMA and $x^{\mathbb{A}} \notin \mathbb{A}(A)$ (3rd equation), we have:

$$
\begin{aligned}
\mathrm{eval}(\mathcal{S} \uplus \mathsf{e}(\pi)(\tau) \uplus \tau')(A\{x^{\mathbb{B}} \mapsto x^{\mathbb{A}}\}) &= \\
\mathrm{eval}(\mathcal{S} \uplus ((\{x^{\mathbb{B}} \mapsto x^{\mathbb{A}}\} \uplus {}_{\mathbb{V}\mathbb{A}\mathbb{B}\setminus\{x^{\mathbb{B}}\}}\!\upharpoonright\mathrm{id}) \circ \mathrm{eval}(\mathcal{S} \uplus \mathsf{e}(\pi)(\tau) \uplus \tau')))(A) &= \\
\mathrm{eval}(\mathcal{S} \uplus \{x^{\mathbb{B}} \mapsto o\} \uplus \mathsf{e}(\pi)(\tau) \uplus \tau')(A) &= \\
\mathrm{eval}(\mathcal{S} \uplus \{x^{\mathbb{B}} \mapsto o\} \uplus \mathsf{e}(\pi)(\tau) \uplus \tau)(A) &= \ \mathsf{FALSE}.
\end{aligned}
$$

<u>Claim 4:</u> $A\{x^{\mathbb{B}} \mapsto x^{\mathbb{A}}\}$ is false in $\mathcal{S} \uplus \mathsf{e}(\pi)(\tau') \uplus \tau'$.

<u>Proof of Claim 4:</u> Otherwise, because of the EXPLICITNESS LEMMA and the entire previous equation, there must be some $u^{\mathbb{V}} \in \mathbb{V}(A\{x^{\mathbb{B}} \mapsto x^{\mathbb{A}}\})$ with $x^{\mathbb{A}}\ S_\pi\ u^{\mathbb{V}}$. Then we have $u^{\mathbb{V}}\ N'\ x^{\mathbb{A}}$. Thus, we know that $(P' \cup S_\pi)^+ \circ N'$ is not irreflexive, which contradicts $\pi$ being $\mathcal{S}$-compatible with $(C', (P', N'))$. <u>Q.e.d. (Claim 4)</u>

By the Claims 4 and 2, $\{A\{x^{\mathbb{B}} \mapsto x^{\mathbb{A}}\}\ \ \Gamma\ \ \Pi\}$ is false in $\mathcal{S} \uplus \mathsf{e}(\pi)(\tau') \uplus \tau'$, as was to be show for the soundness direction of the proof.

Finally, for the safeness direction, assume that the sequent $\Gamma\ \ \forall x^{\mathbb{B}}.\ A\ \ \Pi$ is $(\pi, \mathcal{S})$-valid. For arbitrary $\tau : \mathbb{A} \to \mathcal{S}$, we have to show that the lower sequent $A\{x^{\mathbb{B}} \mapsto x^{\mathbb{A}}\}\ \Gamma\ \Pi$ is true in $\mathcal{S} \uplus \delta$ for $\delta := \mathsf{e}(\pi)(\tau) \uplus \tau$. If some formula in $\Gamma\Pi$ is true in $\mathcal{S} \uplus \delta$, then the lower sequent is true in $\mathcal{S} \uplus \delta$ as well. Otherwise, $\forall x^{\mathbb{B}}.\ A$ is true in $\mathcal{S} \uplus \delta$. Then, by the forward direction of the $\forall$-LEMMA, this means that $A$ is true in $\mathcal{S} \uplus \chi \uplus \delta$ for all $\mathcal{S}$-valuations $\chi : \{x^{\mathbb{B}}\} \to \mathcal{S}$. Then, by the SUBSTITUTION [VALUE] LEMMA (1st equation), and by the VALUATION-LEMMA(0) (2nd equation), we have:

$$
\begin{aligned}
\mathrm{eval}(\mathcal{S} \uplus \delta)(A\{x^{\mathbb{B}} \mapsto x^{\mathbb{A}}\}) &= \\
\mathrm{eval}(\mathcal{S} \uplus ((\{x^{\mathbb{B}} \mapsto x^{\mathbb{A}}\} \uplus {}_{\mathbb{V}\mathbb{A}\mathbb{B}\setminus\{x^{\mathbb{B}}\}}\!\upharpoonright\mathrm{id}) \circ \mathrm{eval}(\mathcal{S} \uplus \delta)))(A) &= \\
\mathrm{eval}(\mathcal{S} \uplus \{x^{\mathbb{B}} \mapsto \delta(x^{\mathbb{A}})\} \uplus \delta)(A) &= \ \mathsf{TRUE}.
\end{aligned}
$$

<u>$\delta^+$-rule:</u> In this case, we have $x^{\mathbb{V}} \in \mathbb{V} \setminus (\mathrm{dom}(C \cup P \cup N) \cup \mathbb{V}(A))$,
$C'' = \{(x^{\mathbb{V}}, \varepsilon x^{\mathbb{B}}.\ \neg A)\}$, and $V = \mathbb{VA}(\forall x^{\mathbb{B}}.\ A) \times \{x^{\mathbb{V}}\} = \mathbb{VA}(A) \times \{x^{\mathbb{V}}\}$.
Thus, $C' = C \cup \{(x^{\mathbb{V}}, \varepsilon x^{\mathbb{B}}.\ \neg A)\}$, $P' = P \cup V$, and $N' = N$.
By assumption of the theorem, $C$ is a $(P, N)$-choice-condition. Thus, $(P, N)$ is a consistent positive/negative variable-condition. Thus, by Definition 5.5, $P$ is well-founded and $P^+ \circ N$ is irreflexive.

<u>Claim 5:</u> $P'$ is well-founded.
<u>Proof of Claim 5:</u> Let $B$ be a non-empty class. We have to show that there is a $P'$-minimal element in $B$. Because $P$ is well-founded, there is some $P$-minimal element in $B$. If this element is $V$-minimal in $B$, then it is a $P'$-minimal element in $B$. Otherwise, this element must be $x^{\mathbb{V}}$ and there is an element $n^{\mathbb{VA}} \in B \cap \mathbb{VA}(A)$. Set $B' := \{\ b^{\mathbb{VA}} \in B \mid b^{\mathbb{VA}}\ P^*\ n^{\mathbb{VA}}\ \}$. Because of $n^{\mathbb{VA}} \in B'$, we know that $B'$ is a non-empty subset of $B$. Because $P$ is well-founded, there is some $P$-minimal element $m^{\mathbb{VA}}$ in $B'$. Then $m^{\mathbb{VA}}$ is also a $P$-minimal element in $B$. Because of $x^{\mathbb{V}} \notin \mathbb{VA}(A) \cup \mathrm{dom}(P)$, we know that $x^{\mathbb{V}} \notin B'$. Thus, $m^{\mathbb{VA}} \neq x^{\mathbb{V}}$. Thus, $m^{\mathbb{VA}}$ is also a $V$-minimal element of $B$. Thus, $m^{\mathbb{VA}}$ is also a $P'$-minimal element of $B$.
<u>Q.e.d. (Claim 5)</u>

<u>Claim 6:</u> $(P')^+ \circ N'$ is irreflexive.
<u>Proof of Claim 6:</u> Suppose the contrary. Because $P^+ \circ N$ is irreflexive, $P^* \circ (V \circ P^*)^+ \circ N$ must be reflexive. Because of $\mathrm{ran}(V) = \{x^{\mathbb{V}}\}$ and $\{x^{\mathbb{V}}\} \cap \mathrm{dom}(P \cup N) = \emptyset$, we have $V \circ P = \emptyset$ and $V \circ N = \emptyset$. Thus, $P^* \circ (V \circ P^*)^+ \circ N = P^* \circ V^+ \circ N = \emptyset$. <u>Q.e.d. (Claim 6)</u>

<u>Claim 7:</u> $C'$ is a $(P', N')$-choice-condition.
<u>Proof of Claim 7:</u> By Claims 5 and 6, $(P', N')$ is a consistent positive/negative variable-condition. As $x^{\mathbb{V}} \in \mathbb{V} \setminus \mathrm{dom}(C)$, we know that $C'$ is a partial function on $\mathbb{V}$ just as $C$. Moreover, for $y^{\mathbb{V}} \in \mathrm{dom}(C')$, we either have $y^{\mathbb{V}} \in \mathrm{dom}(C)$ and then
$\mathbb{VA}(C'(y^{\mathbb{V}})) \times \{y^{\mathbb{V}}\} = \mathbb{VA}(C(y^{\mathbb{V}})) \times \{y^{\mathbb{V}}\} \subseteq P^+ \subseteq (P')^+$, or $y^{\mathbb{V}} = x^{\mathbb{V}}$ and then
$\mathbb{VA}(C'(y^{\mathbb{V}})) \times \{y^{\mathbb{V}}\} = \mathbb{VA}(\varepsilon x^{\mathbb{B}}.\ \neg A) \times \{x^{\mathbb{V}}\} = V \subseteq P' \subseteq (P')^+$. <u>Q.e.d. (Claim 7)</u>

Now it suffices to show that, for each $\tau : \mathbb{A} \to \mathcal{S}$, and for $\delta := \mathsf{e}(\pi)(\tau) \uplus \tau$, the truth of $\{\Gamma \quad \forall x^{\mathbb{B}}.\ A \quad \Pi\}$ in $\mathcal{S} \uplus \delta$ is equivalent that of $\{A\{x^{\mathbb{B}} \mapsto x^{\mathbb{V}}\} \quad \Gamma \quad \Pi\}$.

For the soundness direction, it suffices to show that the former sequent is true in $\mathcal{S} \uplus \delta$ under the assumption that the latter is. If some formula in $\Gamma\Pi$ is true in $\mathcal{S} \uplus \delta$, then the former sequent is true in $\mathcal{S} \uplus \delta$ as well. Otherwise, this means that $A\{x^{\mathbb{B}} \mapsto x^{\mathbb{V}}\}$ is true in $\mathcal{S} \uplus \delta$. Then, by the forward direction of the ¬-LEMMA, $\neg A\{x^{\mathbb{B}} \mapsto x^{\mathbb{V}}\}$ is false in $\mathcal{S} \uplus \delta$. By the EXPLICITNESS LEMMA, $\neg A\{x^{\mathbb{B}} \mapsto x^{\mathbb{V}}\}$ is false in $\mathcal{S} \uplus \delta \uplus \chi$ for all $\chi : \{x^{\mathbb{B}}\} \to \mathcal{S}$. Because $\pi$ is $\mathcal{S}$-compatible with $(C', (P', N'))$ and because of $C'(x^{\mathbb{V}}) = \varepsilon x^{\mathbb{B}}.\ \neg A$, by Item 2 of Definition 7.4, $\neg A$ is false in $\mathcal{S} \uplus \delta \uplus \chi$ for all $\chi : \{x^{\mathbb{B}}\} \to \mathcal{S}$. Then, by the backward direction of the ¬-LEMMA, $A$ is true in $\mathcal{S} \uplus \delta \uplus \chi$ for all $\chi : \{x^{\mathbb{B}}\} \to \mathcal{S}$. Then, by the backward direction of the ∀-LEMMA, $\forall x^{\mathbb{B}}.\ A$ is true in $\mathcal{S} \uplus \delta$.

The safeness direction is perfectly analogous to the case of the $\delta^-$-rule.

**Q.e.d. (Theorem 8.3)**

**Proof of Theorem 8.5**

(1a): If $G_0$ is $(C',(P',N'))$-valid in $\mathcal{S}$, then there is some $\pi$ that is $\mathcal{S}$-compatible with $(C',(P',N'))$ such that $G_0$ is $(\pi,\mathcal{S})$-valid. By Lemma 7.6, $\pi$ is also $\mathcal{S}$-compatible with $(C,(P,N))$. Thus, $G_0$ is $(C,(P,N))$-valid, in $\mathcal{S}$.

(1b): Suppose that $\pi$ is $\mathcal{S}$-compatible with $(C',(P',N'))$, and that $G_1$ is $(\pi,\mathcal{S})$-valid. By Lemma 7.6, $\pi$ is also $\mathcal{S}$-compatible with $(C,(P,N))$. Thus, since $G_0$ $(C,(P,N))$-reduces to $G_1$, also $G_0$ is $(\pi,\mathcal{S})$-valid as was to be shown.

(2): Assume the situation described in the lemma.

Claim 1: $O' \subseteq \mathrm{dom}(C) \setminus O$.

Proof of Claim 1: By definition, $O' \subseteq \mathrm{dom}(C)$. It remains to show $O' \cap O = \emptyset$. To the contrary, suppose that there is some $y^{\mathbb{V}} \in O' \cap O$. Then, by the definition of $O'$, there is some $z^{\mathbb{V}} \in M \backslash O$ with $z^{\mathbb{V}}\, P^*\, y^{\mathbb{V}}$. By definition of $O$, however, we have $y^{\mathbb{V}} \in P^*\langle V\rangle$. Thus, $z^{\mathbb{V}} \in P^*\langle V\rangle$. Thus, $z^{\mathbb{V}} \in O$, a contradiction. Q.e.d. (Claim 1)

Claim 2: $\langle O'\rangle P^+ \cap \mathrm{dom}(C) \subseteq O'$.

Proof of Claim 2: Assume $y^{\mathbb{V}} \in O'$ and $z^{\mathbb{V}} \in \mathrm{dom}(C)$ with $y^{\mathbb{V}}\ P^+\ z^{\mathbb{V}}$. It now suffices to show $z^{\mathbb{V}} \in O'$. Because of $y^{\mathbb{V}} \in O'$, there is some $x^{\mathbb{V}} \in M \backslash O$ with $x^{\mathbb{V}}\, P^*\, y^{\mathbb{V}}$. Thus, $x^{\mathbb{V}}\, P^*\, z^{\mathbb{V}}$. Thus, $z^{\mathbb{V}} \in O'$. Q.e.d. (Claim 2)

Claim 3: $\mathrm{dom}(\sigma) \cap \mathrm{dom}(C) \subseteq O' \cup O$.

Proof of Claim 3: $\mathrm{dom}(\sigma) \cap \mathrm{dom}(C) = \mathrm{dom}(C) \cap M \subseteq O \cup (\mathrm{dom}(C) \cap (M \backslash O)) \subseteq O \cup (\mathrm{dom}(C) \cap \langle M \backslash O\rangle P^*) = O \cup O'$. Q.e.d. (Claim 3)

Claim 4: $O' \cap \mathbb{V}(G_0, G_1) = O' \cap V = \emptyset$.

Proof of Claim 4: Because of $\mathbb{V}(G_0, G_1) \subseteq V$, it suffices to show the second equality. To the contrary of the second equality, suppose that there is some $y^{\mathbb{V}} \in O' \cap V$. Then, by the definition of $O'$, there is some $z^{\mathbb{V}} \in M \backslash O$ with $z^{\mathbb{V}}\, P^*\, y^{\mathbb{V}}$. By definition of $O$, however, we have $z^{\mathbb{V}} \in O$, a contradiction. Q.e.d. (Claim 4)

(2a): In case that $G_0\sigma \cup (\langle O\rangle Q_C)\sigma$ is $(C',(P',N'))$-valid in $\mathcal{S}$, there is some $\pi'$ that is $\mathcal{S}$-compatible with $(C',(P',N'))$ such that $G_0\sigma \cup (\langle O\rangle Q_C)\sigma$ is $(\pi',\mathcal{S})$-valid. Then both $G_0\sigma$ and $(\langle O\rangle Q_C)\sigma$ are $(\pi',\mathcal{S})$-valid. By Claims 1, 2, 3, and 4, let $\pi$ be given as in Lemma 7.11. Then $G_0$ is $(\pi,\mathcal{S})$-valid. Moreover, as $\pi$ is $\mathcal{S}$-compatible with $(C,(P,N))$, $G_0$ is $(C,(P,N))$-valid in $\mathcal{S}$.

(2b): Let $\pi'$ be $\mathcal{S}$-compatible with $(C',(P',N'))$, and suppose that $G_1\sigma \cup (\langle O\rangle Q_C)\sigma$ is $(\pi',\mathcal{S})$-valid. Then both $G_1\sigma$ and $(\langle O\rangle Q_C)\sigma$ are $(\pi',\mathcal{S})$-valid. By Claims 1, 2, 3, and 4, let $\pi$ be given as in Lemma 7.11. Then $\pi$ is $\mathcal{S}$-compatible with $(C,(P,N))$, and $G_1$ is $(\pi,\mathcal{S})$-valid. By assumption, $G_0$ $(C,(P,N))$-reduces to $G_1$. Thus, $G_0$ is $(\pi,\mathcal{S})$-valid, too. Thus, by Lemma 7.11, $G_0\sigma$ is $(\pi',\mathcal{S})$-valid as was to be shown.

<u>(3):</u> Let $\pi$ be $\mathcal{S}$-compatible with $(C,(P,N))$, and suppose that $G_0$ is $(\pi,\mathcal{S})$-valid. Let $\tau : \mathbb{A} \to \mathcal{S}$ be an arbitrary $\mathcal{S}$-valuation. Set $\delta := \mathsf{e}(\pi)(\tau) \uplus \tau$. It now suffices to show the equation $\mathrm{eval}(\mathcal{S} \uplus \delta)(G_0\nu) = \mathsf{TRUE}$.

Define $\tau' : \mathbb{A} \to \mathcal{S}$ via $\tau'(y^{\mathbb{A}}) := \left\{\begin{array}{ll} \tau(y^{\mathbb{A}}) & \text{for } y^{\mathbb{A}} \in \mathbb{A}\backslash\mathrm{dom}(\nu) \\ \mathrm{eval}(\mathcal{S} \uplus \delta)(\nu(y^{\mathbb{A}})) & \text{for } y^{\mathbb{A}} \in \mathrm{dom}(\nu) \end{array}\right\}$.

<u>Claim 5:</u> For $v^{\mathbb{V}} \in \mathbb{V}(G_0)$ we have $\mathsf{e}(\pi)(\tau)(v^{\mathbb{V}}) = \mathsf{e}(\pi)(\tau')(v^{\mathbb{V}})$.

<u>Proof of Claim 5:</u> Otherwise there must be some $y^{\mathbb{A}} \in \mathrm{dom}(\nu)$ with $y^{\mathbb{A}}\ S_\pi\ v^{\mathbb{V}}$. Because of $v^{\mathbb{V}} \in \mathbb{V}(G_0)$ and $\mathbb{V}(G_0) \times \mathrm{dom}(\nu) \subseteq N$, we have $v^{\mathbb{V}}\ N\ y^{\mathbb{A}}$. But then $(P \cup S_\pi, N)$ is not consistent, which contradicts $\pi$ being $\mathcal{S}$-compatible with $(C,(P,N))$. <u>Q.e.d. (Claim 5)</u>

Then, all in all, we get the sufficient equation:

$$\begin{array}{rcl}
\mathrm{eval}(\mathcal{S} \uplus \delta)(G_0\nu) & = & \mathrm{eval}\Big(\ \mathcal{S}\ \uplus\ (\ (\ \nu\ \uplus\ {}_{\mathbb{V}\mathbb{A}\backslash\mathrm{dom}(\nu)}\!\upharpoonright\!\mathrm{id}\ )\ \circ\ \mathrm{eval}(\mathcal{S} \uplus \delta)\ )\ \Big)\Big(\ G_0\ \Big) \\
& = & \mathrm{eval}\Big(\ \mathcal{S}\ \uplus\ (\ \nu\ \circ\ \mathrm{eval}(\mathcal{S} \uplus \delta)\ )\ \uplus\ {}_{\mathbb{V}\mathbb{A}\backslash\mathrm{dom}(\nu)}\!\upharpoonright\!\delta\ \Big)\Big(\ G_0\ \Big) \\
& = & \mathrm{eval}\Big(\ \mathcal{S}\ \uplus\ \tau'\ \uplus\ \mathsf{e}(\pi)(\tau)\ \Big)\Big(\ G_0\ \Big) \\
& = & \mathrm{eval}\Big(\ \mathcal{S}\ \uplus\ \tau'\ \uplus\ {}_{\mathbb{V}(G_0)}\!\upharpoonright\!(\mathsf{e}(\pi)(\tau))\ \Big)\Big(\ G_0\ \Big) \\
& = & \mathrm{eval}\Big(\ \mathcal{S}\ \uplus\ \tau'\ \uplus\ {}_{\mathbb{V}(G_0)}\!\upharpoonright\!(\mathsf{e}(\pi)(\tau'))\ \Big)\Big(\ G_0\ \Big) \\
& = & \mathrm{eval}\Big(\ \mathcal{S}\ \uplus\ \tau'\ \uplus\ \mathsf{e}(\pi)(\tau')\ \Big)\Big(\ G_0\ \Big) \\
& = & \mathsf{TRUE},
\end{array}$$

as follows:
1st equation by the forward direction of the SUBSTITUTION [VALUE] LEMMA,
2nd equation by the VALUATION-LEMMA(0),
3rd equation by definition of $\tau'$ and $\delta$,
4th equation by the forward direction of the EXPLICITNESS LEMMA,
5th equation by Claim 5,
6th equation by the backward direction of the EXPLICITNESS LEMMA, and
7th equation by the $(\pi,\mathcal{S})$-validity of $G_0$.

**Q.e.d. (Theorem 8.5)**

# Index